\documentclass{sig-alternate}
\DeclareRobustCommand{\ttfamily}{\fontencoding{T1}\fontfamily{lmtt}\selectfont}

\usepackage{verbatim}
\usepackage[pagebackref=true,breaklinks=true,letterpaper=true,bookmarks=false]{hyperref}

\usepackage{color}
\usepackage[percent]{overpic}

\newcommand{\red}{\color[rgb]{1,0,0}}

\usepackage{transparent}

\usepackage{paralist}
\usepackage{graphicx}
\usepackage{subcaption}
\graphicspath{{eps/}}

\newcommand{\X}{\mathbf{x}}
\newcommand{\Y}{\mathbf{y}}
\newcommand{\M}{\mathbf{M}}

\begin{document}

\setcopyright{acmcopyright}
\conferenceinfo{ICIMCS '15,}{August 19-21, 2015, Zhangjiajie, Hunan, China}
\isbn{978-1-4503-3528-7/15/08}\acmPrice{\$15.00}
\doi{http://dx.doi.org/10.1145/2808492.2808564}

\title{Using User Generated Online Photos to Estimate and Monitor Air Pollution in Major Cities}

%
%
%
%
%

\numberofauthors{3} 
%
\author{
    \alignauthor
    Yuncheng Li\\
    \affaddr{University of Rochester}\\
    \affaddr{617 Computer Studies Bldg}\\
    \affaddr{Rochester, New York 14627}\\
    \email{raingomm@gmail.com}
    \alignauthor
    Jifei Huang \\
    \affaddr{University of Rochester}\\
    \email{jhuang48@ur.rochester.edu}
    \alignauthor
    Jiebo Luo  \\
    \affaddr{University of Rochester}\\
    \affaddr{611 Computer Studies Bldg}\\
    \affaddr{Rochester, New York 14627}\\
    \email{jiebo.luo@gmail.com}
}
\date{30 July 1999}

\maketitle
\begin{abstract}
    With the rapid development of economy in China over the past decade, air pollution has become an increasingly serious problem in major cities and caused grave public health concerns in China. Recently, a number of studies have dealt with air quality and air pollution. Among them, some attempt to predict and monitor the air quality from different sources of information, ranging from deployed physical sensors to social media. These methods are either too expensive or unreliable, prompting us to search for a novel and effective way to sense the air quality. In this study, we propose to employ the state of the art in computer vision techniques to analyze photos that can be easily acquired from online social media. Next, we establish the correlation between the haze level computed directly from photos with the official PM 2.5 record of the taken city at the taken time. Our experiments based on both synthetic and real photos have shown the promise of this image-based approach to estimating and monitoring air pollution.
\end{abstract}




\section{Introduction}
\begin{figure}[t]
    \centering
    \includegraphics[width=.99\columnwidth]{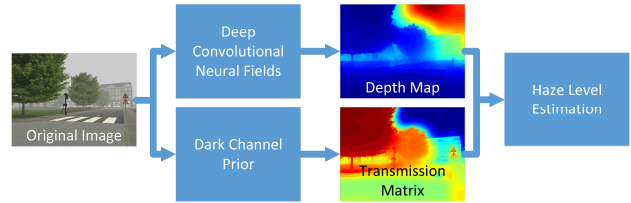}
    \caption{The overview of the proposed framework. Given a photo, we first estimate the transmission matrix using the Dark Channel Prior (DCP) \cite{5567108}. In parallel, we estimate the depth map based on the Deep Convolutional Neural Fields (DCNF) \cite{Depth2015Liu}. By combining the transmission matrix and depth map, we estimate the haze level of the photo.}
    \label{fig:overview}
\end{figure}

Air pollution is one of the major environmental side products caused by moderm industrialization. First step to control air pollution is to monitor the air quality and raise the awareness among people. Airborne Particulate Matter is one kind of air pollutant transmitting hazardous chemicals, which penetrate deeply into human lung and blood, causing many healthy problems \cite{raaschou2013air}. PM2.5/Haze, a finest kind of Airborne Particulate Matter, has recently attracted much attention among people living in large cities in China, such as Beijing, because it has been the major air pollutant since the government began to publish the PM2.5/Haze data in 2012. In this paper, we propose a system to estimate haze level based on single photo.

While an accurate air quality sensor network has been established across the world, there are multiple advantages to use a photo to estimate the haze level: \begin{inparaenum}[1)]
\item Sensors are expensive and therefore the coverage is limited. According to an official real time air quality data platform\footnote{\url{http://113.108.142.147:20035/emcpublish/}}, there are only 12 monitor stations for the giant Beijing city. Also, many cities and rural areas have no monitor stations at all. Therefore, haze monitoring using the ubiquitous online photos can serve as an information source complementary to official data.
\item Haze estimation from a photo will enable mobile phone users to snap a photo and measure air quality. The micro level information, in contrast to the macro level metrics, such as Air Quality Index, is especially valuable for individuals.
\end{inparaenum}

Although it seems simple for bare eyes, estimating haze level automatically using photo is challenging, partly due to the large visual variations of the scenes, different photography skill levels of the mobile users, and even various photo resolutions. Our solution to this problem is illustrated in Fig.~\ref{fig:overview}. We first estimate a transmission matrix generated from a haze removal algorithm, and estimate the depth map for all pixels in the photo. A haze level score is computed by combining the transmission matrix and depth map, and can be calibrated to estimate the PM2.5 level. We consider the transmission matrix as the perceived depth of hazy photos, which is a combination of actual depth and haze effects. Therefore, by ruling out the actual depth factor, we can isolate the haze effects from the transmission matrix, which is used to estimate the haze level.

We make the following contributions in this paper,
\begin{itemize}
    \item We propose an effective method to estimate the haze level from photo.
    \item We augment an existing haze removal benchmark for haze level estimation research.
    \item We collect a large scale dataset with more than 8,000 photos associated with PM2.5 data. Along with a synthetic image dataset, the real world data helps validate the effectiveness of our image-based approach.
\end{itemize}


\section{Related Work}
\label{sec:related}

Share the similar motivation to provide information source complementary to official data, there are several previous methods on using auxiliary data to monitor air quality. For example, \cite{Aoki:2009:VRU:1518701.1518762} proposed to install sensors on city street sweepers to monitor air quality in San Fransisco. \cite{Chen:2014:BSM:2567948.2576941} proposed to integrate social media and official records to monitor air quality and predict health hazardous. \cite{6921638} proposed to identify keywords on Weibo (Chinese version of Twitter) to track city level Air Quality Index. \cite{poduri2010visibility} also proposed to estimate visibility/haze level based on photo. Our approach is different from \cite{poduri2010visibility} in that \begin{inparaenum}[1)]
\item \cite{poduri2010visibility} assumes manually segmented sky regions.
\item \cite{poduri2010visibility} needs camera calibration and other sensors, such as accelerometers and magnetometers, to calibrate the luminance.
\end{inparaenum} However, our method does not have these restrictions. \cite{jcsb.1000161} also proposed to estimate haze level using photo and their method is based on statistics computed directly from image pixels and therefore is most related to our method. We compare with \cite{jcsb.1000161} in our Experiments, which show that our method is superior in the presence of complex scenes and haze conditions.

\section{Proposed Method}
\label{sec:method}
Fig.~\ref{fig:overview} illustrates the framework of the proposed method. There are three major components: transmission matrix estimation, depth map estimation and haze level estimation.

\subsection{Haze Model}
Following \cite{6190796}, the imaging process of a photo taken under haze condition is modeled by the following equation,
\begin{equation}
    \begin{array}[h]{c}
        L(\X) = L_0(\X) t(\X) + L_s(\X) (1 - t(\X)) \\
        t(\X) = e^{-k d(\X)},
    \end{array}
    \label{eqn:haze-model}
\end{equation}
in which $\X$ is the pixel coordinates, $L(\X)$ is the pixel value sensed by the camera, $L_0(\X)$ is the actual luminance of the scene, $t(\X)$ is called transmission matrix, $d(\X)$ is the depth map of the scene, $k$ controls the haze intensity,  and $L_s(\X)$ denotes the lighting condition, e.g., sky luminance. In this paper, we are interested in estimate the haze level, i.e., the $k$ value. As illustrated in Fig.~\ref{fig:frida-k}, a larger $k$ indicates heavier haze.

\subsection{Estimate Transmission}
Based on an effective Dark Channel Prior, \cite{5567108} proposed to estimate the transmission matrix $t(\X)$ using the following equation,
\begin{equation}
    \tilde{t}(\X) = 1 - \omega \min_c \min_{\Y \in \Omega(\X)} \frac{L^c(\Y)}{A^c},
    \label{eqn:raw-t}
\end{equation}
in which $c$ denotes the color channels, e.g., RGB, $\omega$ controls the amount of haze to preserve to make the final dehazed photo look natural and is empirically fixed at $0.95$, $\Omega(\X)$ denotes an image patch centered at $\X$ and the patch size is fixed at $15$, $L^c(\Y)$ is the pixel value of channel $c$ at $\Y$, and $A^c$ denotes the estimated sky luminance. We follow the same algorithm as \cite{5567108} to estimate $A^c$ and fix it for each channel and image. Note that Eqn.~\eqref{eqn:raw-t} can be easily implemented using elementwise operations and an image erosion.

After the rough estimation in Eqn.~\eqref{eqn:raw-t}, a soft matting or more efficiently guided filtering \cite{6319316} is applied to refine the transmission matrix. Given the guided filter is becoming a standard operation, we simply denote the refining process as the following,
\begin{equation}
    t(\X) = \text{GuidedFilter} (\tilde{t}(\X), L(\X), W),
    \label{eqn:refine-t}
\end{equation}
in which $W$, the window size, is a parameter for the guided filter and is empirically fixed at $60$.

\subsection{Depth Estimation}
Given the transmission matrix $t(\X)$, the value $k$ in Eqn.~\eqref{eqn:haze-model} can be computed directly if the depth map $d(\X)$ is known. Therefore, we propose to use a standalone image depth estimator to remove the effect of $d(\X)$ in $t(\X)$. We adopt the Deep Convolutional Neural Fields (DCNF) proposed in \cite{Depth2015Liu} for depth estimation. DCNF estimates depth using image by inference from a learned CRF over superpixels, and objective function of the CRF is a combination of the unary and pairwise potentials as follows,
\begin{equation}
    E(\Y, \X; \theta, \beta) = \sum_{p \in \mathcal{N}} U(y_p, \X; \theta) + \sum_{(p, q) \in \mathcal{S}} V(y_p, y_q, \X; \beta),
    \label{eqn:depth}
\end{equation}
where $\mathcal{N}$ is the set of superpixels, $\mathcal{S}$ is the set of neighborhood superpixel pairs, $U(*)$ is the unary potential parameterized by a multi-layer Convolutional Neural Network over the pixel values and $\theta$ is its network parameters, and $V(*)$ is the pairwise potential parameterized by a single layer Neural Network over a set of similarity measurements, e.g., color histogram and LBP similarity \cite{Depth2015Liu}. The model parameters ($\theta$ and $\beta$) are learned using a standard dataset and we use the model trained from the Make3D dataset \cite{4531745} for our outdoor case.

\subsection{Haze Estimate}
Given the transmission matrix $t(\X)$ and depth map $d(\X)$, it becomes straightforward to estimate the haze level $k$ according to Eqn.~\eqref{eqn:haze-model}. However, given the scaling issues and the fact that while there is only a single haze level $k$ for each image, $t(\X)$ and $d(\X)$ is computed for each pixel, the interactions among these quantities are complicated. We propose to select from a large pool of combinations of transformation and pooling functions, denoted as follows,
\begin{equation}
    \hat{k} = P\{C[T^t(t(\X)), T^d(d(\X))]\},
    \label{eqn:haze}
\end{equation}
where $T^t(*)$ and $T^d(*)$ are the transformation functions, e.g, $log$, over the transmission matrix and depth map, respectively. $C[*]$ is a bivariate function, e.g., \textit{division}, to combine the matrices, and $P\{*\}$ is a pooling function, e.g., \textit{max}, to aggregate the matrix to a single value. We will explain the choices of these functions in the Experiments section.

\newcommand{\egwidth}{.15\textwidth}
\newcommand{\egsubwidth}{1\textwidth}
\begin{figure*}[t]
    \centering
    \begin{subfigure}[b]{\egwidth}
        \includegraphics[width=\egsubwidth]{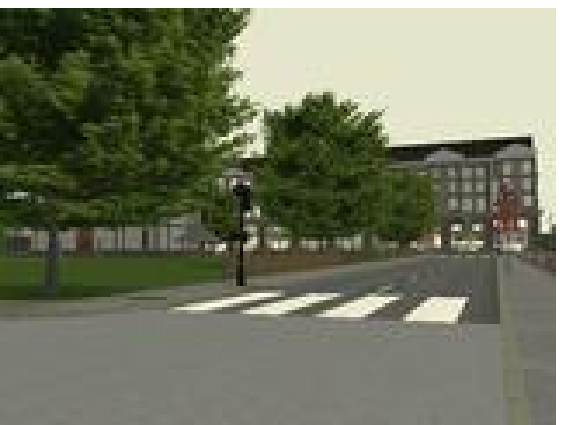}
    \end{subfigure}
    \begin{subfigure}[b]{\egwidth}
        \includegraphics[width=\egsubwidth]{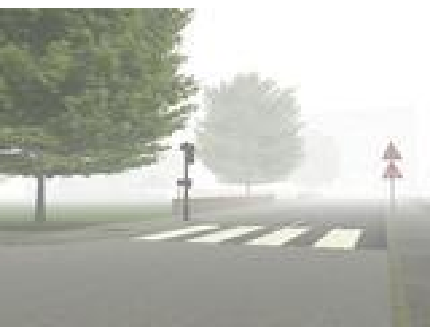}
    \end{subfigure}
    \begin{subfigure}[b]{\egwidth}
        \includegraphics[width=\egsubwidth]{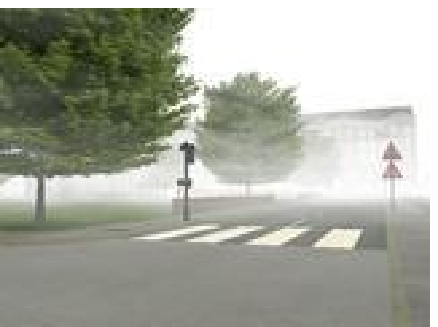}
    \end{subfigure}
    \begin{subfigure}[b]{\egwidth}
        \includegraphics[width=\egsubwidth]{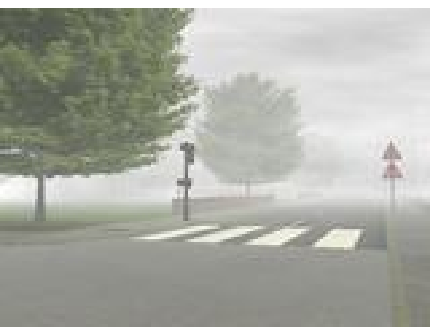}
    \end{subfigure}
    \begin{subfigure}[b]{\egwidth}
        \includegraphics[width=\egsubwidth]{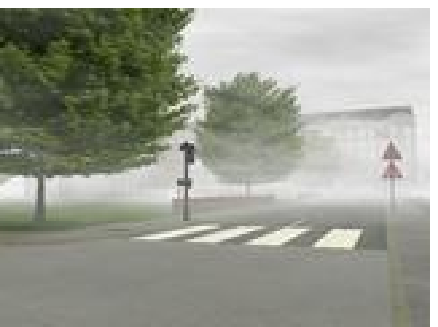}
    \end{subfigure}
    \begin{subfigure}[b]{\egwidth}
        \includegraphics[width=\egsubwidth]{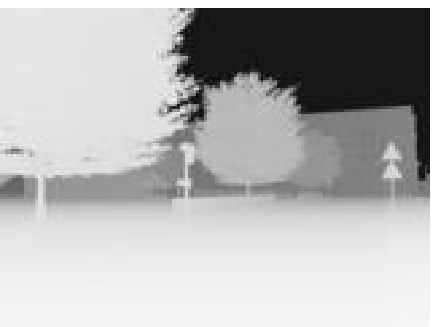}
    \end{subfigure}
    \caption{Example scene from the FRIDA dataset \cite{6190796}. The images are the original image, 4 types of haze conditions and the depth map, respectively.}
    \label{fig:frida}
\end{figure*}

\section{Experiments}
\label{sec:exp}
In this section, we present our experiments to validate the proposed method. We first present the synthetic and real image datasets. Then, we describe the baselines for comparison. Next, we present the comparison results, which demonstrate the effectiveness of the proposed method.

\subsection{Datasets}
We experiment on both synthetic and real images.

\noindent \textbf{FRIDA} is a synthetic haze image dataset serving as a benchmark for haze removal related research. FRIDA1 contains 90 synthetic images of 18 urban road scenes \cite{5548128}. FRIDA2 contains 330 synthetic images of 66 various road scenes \cite{6190796}. Both FRIDA1 and FRIDA2 are generated artificially using the same algorithm. \begin{inparaenum}[1)]
\item A scene, together with its depth map, is generated using a computer software.
\item Given the depth map, 4 types of haze conditions are applied to the generated image (CGI) according to the model in Eqn.~\eqref{eqn:haze-model}.
\end{inparaenum}
An example of image, its depth map and the haze applied images are shown in Fig.~\ref{fig:frida}. The $k$ value in Eqn.~\eqref{eqn:haze-model} is fixed for the released images, which is suitable for haze detection and removal, but not for the haze level estimation. We reproduce the synthetic algorithms using the provided original CGI and depths, but with varying $k$ value to simulate various haze level. Together with 4 types of haze conditions and 9 haze levels, we generate 36 haze images for each scene, so together with the original images, there are 666 images in FRIDA1 and 2437 images in FRIDA2. The effects of larger $k$ is illustrated in Fig.~\ref{fig:frida-k}.

\noindent \textbf{PM25} is the real image dataset we crawled from a tourist website \footnote{http://goo.gl/svzxLm}. The photos in this dataset were taken at various attraction sites in the Beijing city, and the timestamps was recorded. We then associate these photos with the hourly PM2.5 records sensed by the U.S. Embassy in Beijing \footnote{http://goo.gl/0DpK8S}. There are a total of 8,761 photos with associated PM2.5 records in this dataset. We use PM2.5 as a proxy for the haze level $k$ to evaluate the proposed method. Because of measurement errors and other factors involved in PM2.5 records, there are noises in using PM2.5 as a proxy of the haze level. Therefore, we select 46 photos that are manually categorized to \textit{NonHaze}, \textit{LightHaze} and \textit{HeavyHaze}, and we use 0, 1 and 2 as the proxy of the haze level $k$ for each of the categories. There are 22, 14 and 10 photos for each of the categories, respectively. We refer to the full set as \textbf{PM25} and the subset as \textbf{PM25-s}.

\newcommand{\eglwidth}{.19\columnwidth}
\newcommand{\eglsubwidth}{1\textwidth}
\begin{figure}[t]
    \centering
    \begin{subfigure}[b]{\eglwidth}
        \includegraphics[width=\eglsubwidth]{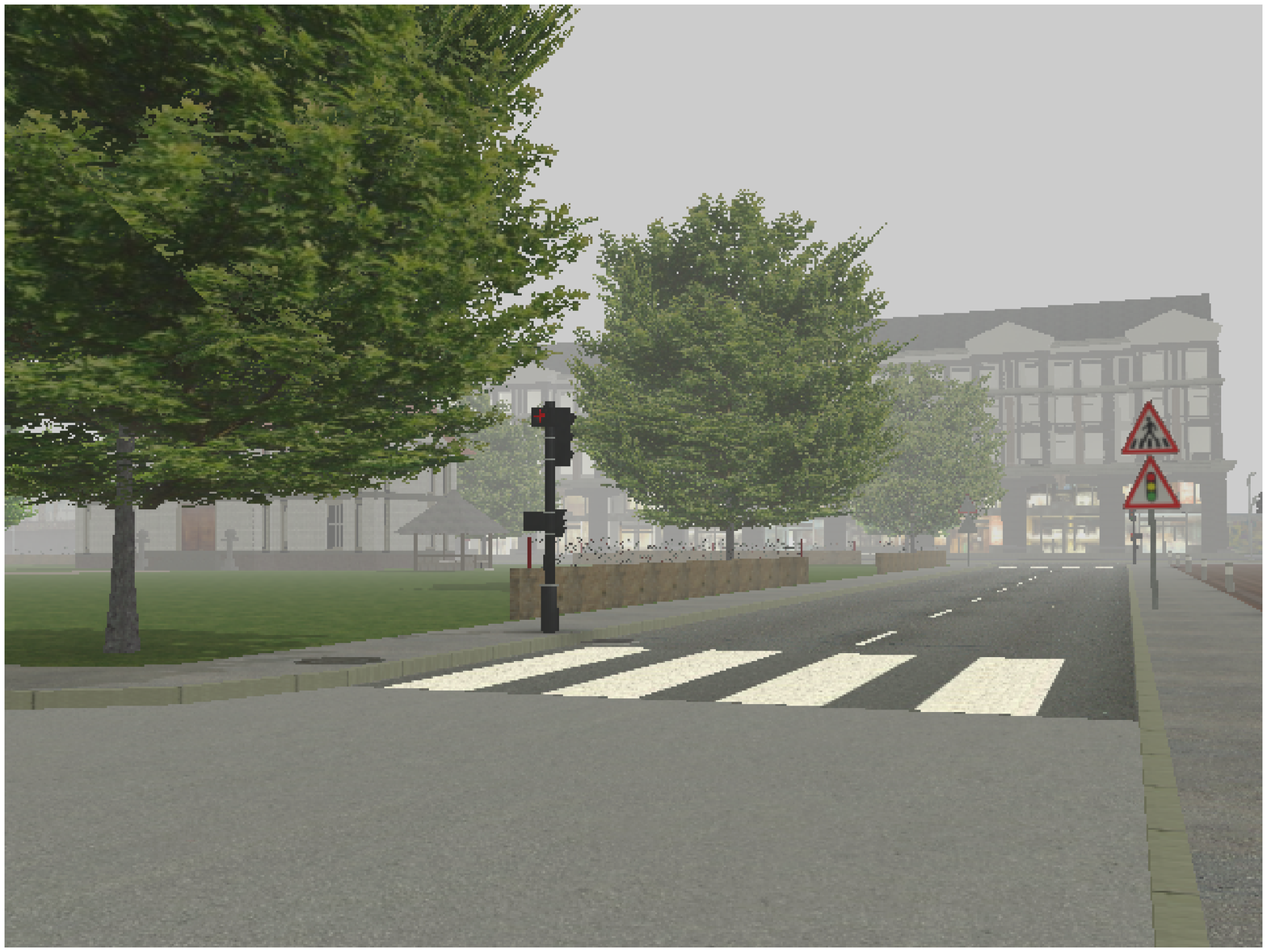}
    \end{subfigure}
    \begin{subfigure}[b]{\eglwidth}
        \includegraphics[width=\eglsubwidth]{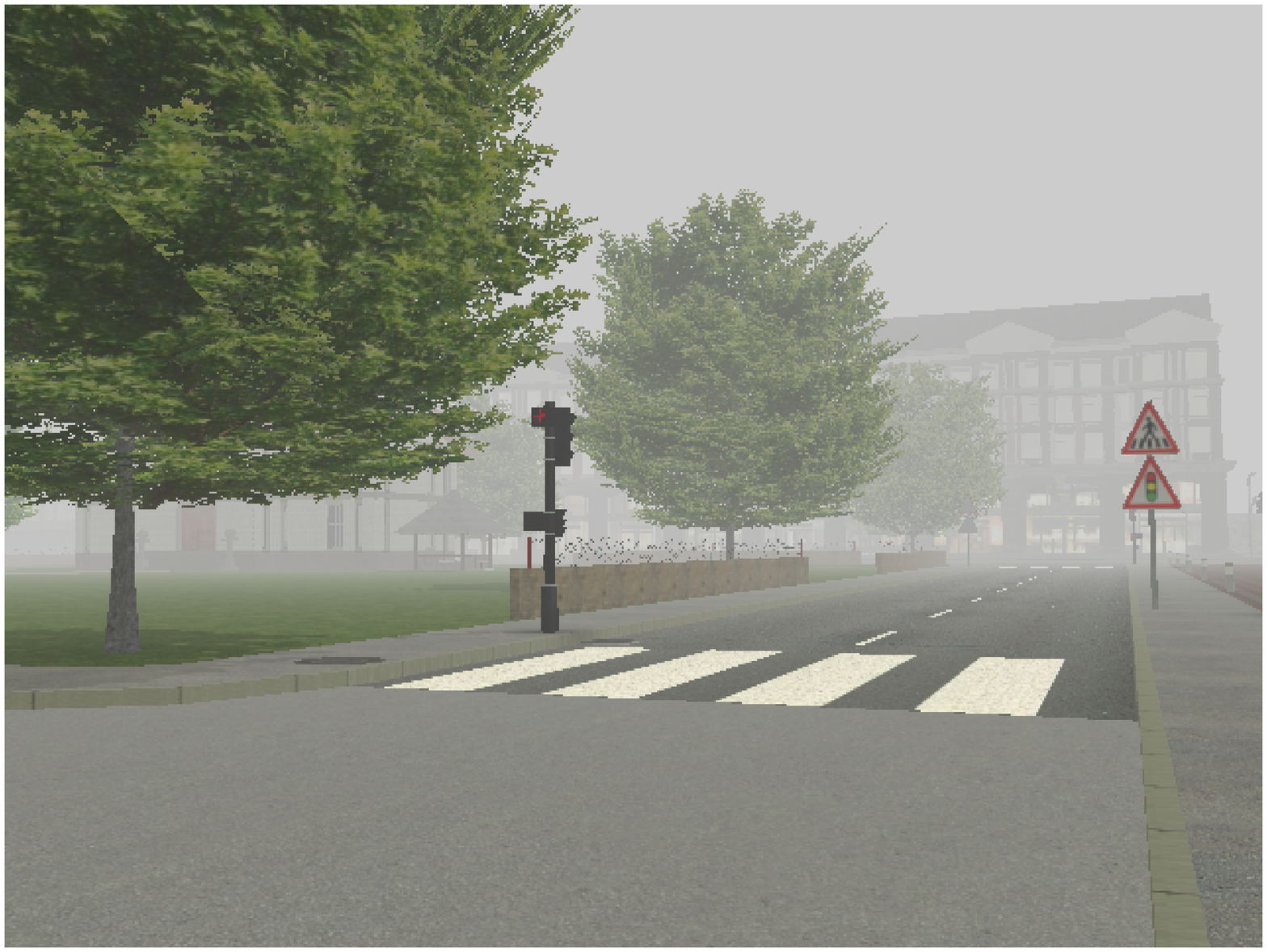}
    \end{subfigure}
    \begin{subfigure}[b]{\eglwidth}
        \includegraphics[width=\eglsubwidth]{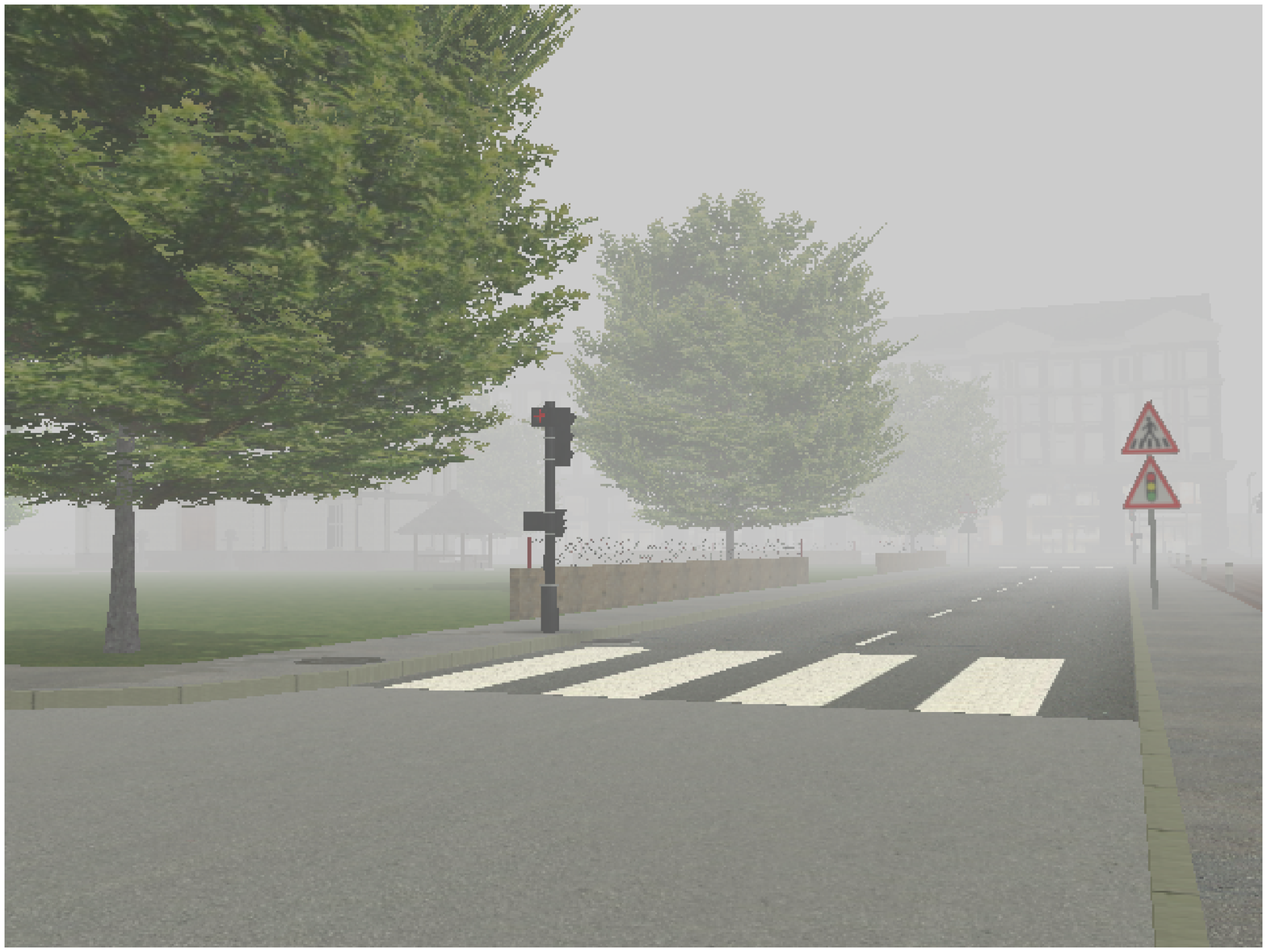}
    \end{subfigure}
    \begin{subfigure}[b]{\eglwidth}
        \includegraphics[width=\eglsubwidth]{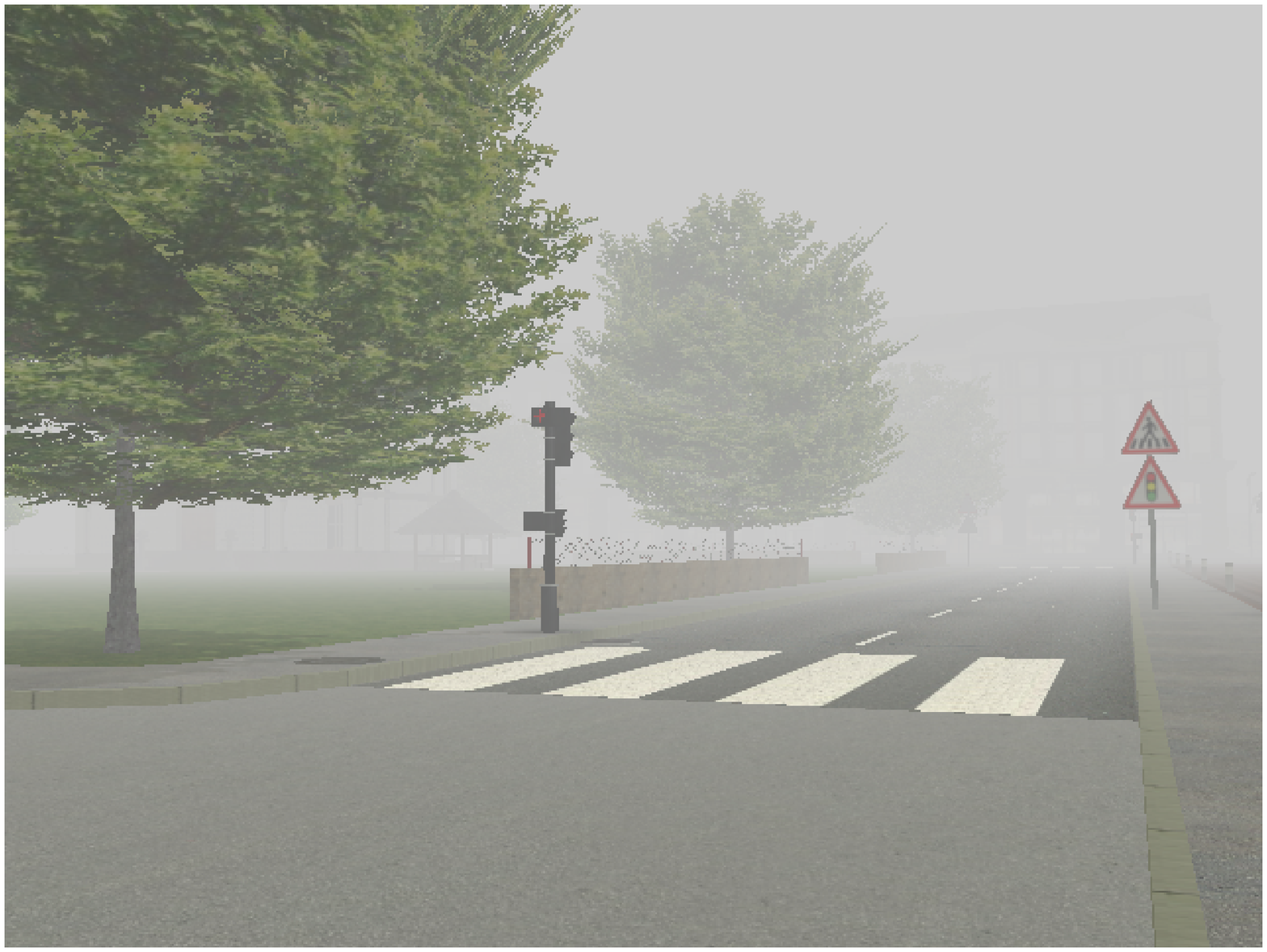}
    \end{subfigure}
    \begin{subfigure}[b]{\eglwidth}
        \includegraphics[width=\eglsubwidth]{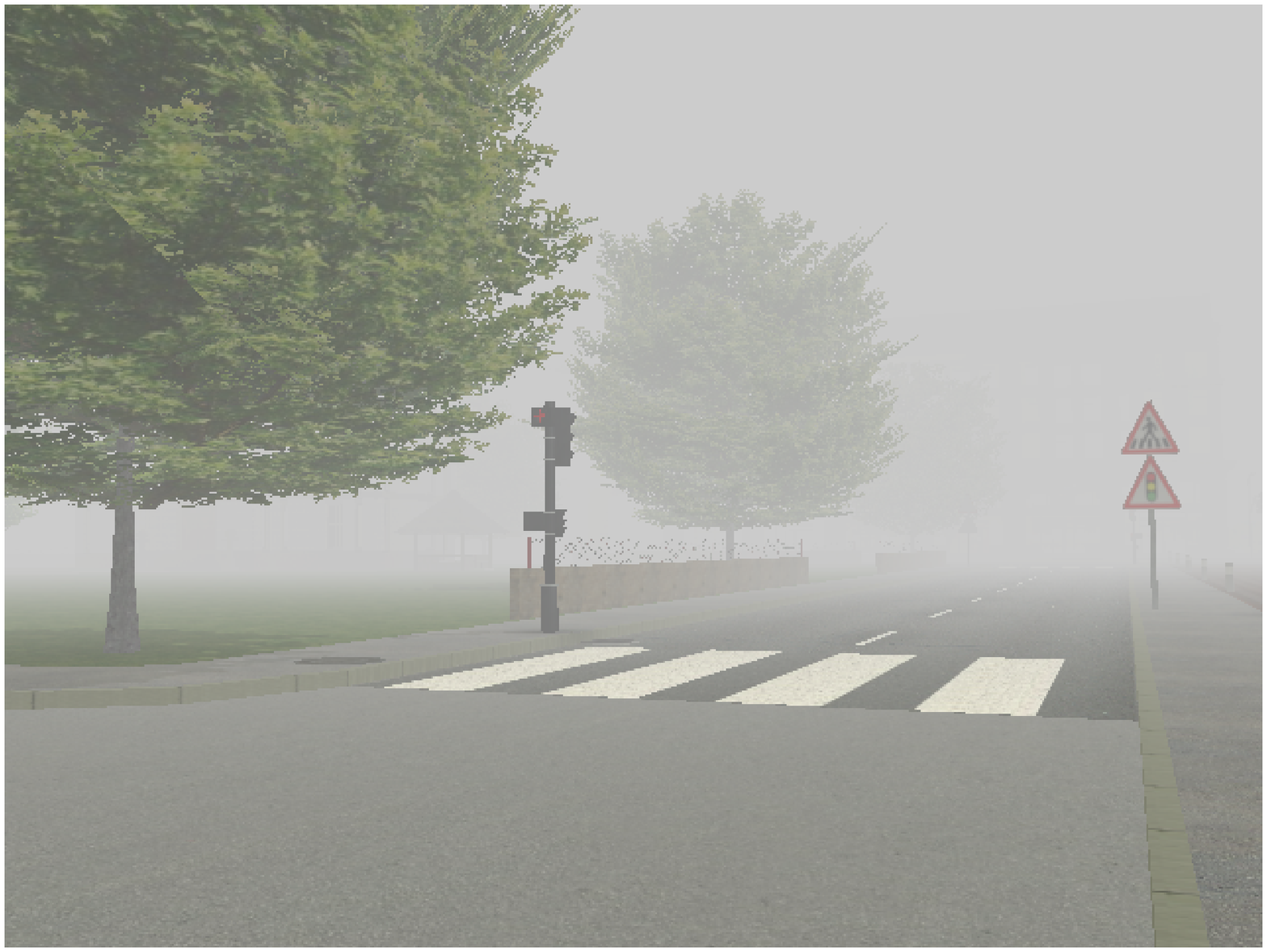}
    \end{subfigure}
    \caption{Varying $k$ in Eqn.~\eqref{eqn:haze-model} for the image in Fig.~\ref{fig:frida}. The haze level increases with increasing $k$ values.}
    \label{fig:frida-k}
\end{figure}

\subsection{Evaluation Protocols}
We compare the proposed method with various baselines and a previous work proposed method in \cite{jcsb.1000161}, in order to show that the combination of transmission matrix and depth map achieves superior performance. The baselines \textsf{trans} and \textsf{depth} are the methods that use only a single factor, i.e., the transmission matrix and the depth map, respectively. \textsf{depth$\otimes$trans} is the proposed method that combines both factors. \textsf{jcsb2014} is the statistical method proposed in \cite{jcsb.1000161}, which is the only previous work we found in the literature that dealt with haze level estimation.

Different choices of the transmission matrix (\textit{raw} and \textit{refined} in Eqn.~\eqref{eqn:raw-t} \eqref{eqn:refine-t}) and the functions in Eqn.~\eqref{eqn:haze} contribute to the pool of all possible variations of the proposed method and the baselines. The choices of functions in Eqn.~\eqref{eqn:haze} are based on our observations on Eqn.~\eqref{eqn:haze-model} and are listed as follows,
\begin{itemize}
    \item Transformation function $T(x)$: $\log(x+1)$, $\log(\log(x+1)+1)$ and the unit transformation $T(x) = x$.
    \item Bivariate function $C[t,d]$: $t * d$, $t/d$ and $d/t$. Also, $C[t,d]$ is $t$ and $d$ in the baseline \textsf{trans} and \textsf{depth}, respectively.
    \item The pooling function $P[\M]$: \textit{mean}, \textit{median}, \textit{max}, 75$^{\text{th}}$ percentile and 90$^{\text{th}}$ percentile.
\end{itemize}

There are 991 variations of the methods, and we report \textit{the best result} for each estimation model. Because of the ordinal natural of the (proxy) haze level, we consider the Spearman correlation coefficients \cite{1975} as the evaluation metric to compare different methods. In addition, the sign of the correlation is irrelevant in the comparison, thus we use the absolute value of the correlation as the final performance metric.

In addition, because all of the methods contain the single feature and no parameter fitting is involved, we do not need to use the standard practice to cross validate the methods.

\subsection{Results}

The evaluation results are shown in Table~\ref{tab:result}, from which we make the following observations:
\begin{itemize}
    \item All methods perform very well on the synthetic image dataset, which means all methods, including the proposed method, baselines and the one proposed in \cite{jcsb.1000161}, are able to capture the haze level to some extent.
    \item The proposed method and baselines perform better than the \textsf{jcsb2014} work. The gain becomes more significant when the scenes and haze conditions are more complicated. See the example photos from different datasets in Fig.~\ref{fig:egs}.
    \item The proposed method, combining depth and transmission, are better than the baselines, using single factors,  especially on the really difficult \textbf{PM25} dataset. This indicates that it is important to consider transmission and depth together. Neither factor alone can correlate well with the haze level.
    \item The proposed method can achieve very high correlation on the manually labeled real images \textbf{PM25-s}, but still not very high on the full \textbf{PM25} dataset, which indicates there are noises using only the PM25 value as a proxy of haze level. In other words, the estimate can only explain 40\% of the variation.
\end{itemize}

In order to further validate the correlations and show the scale of the dataset, the predicated haze level and the ground truth is plotted on Fig.~\ref{fig:corr} for the best \textsf{depth$\otimes$trans} option for each dataset. By simple calibration, we can estimate the haze condition into three levels: \textit{Clear}, \textit{Light} and \textit{Heavy}. In Fig.~\ref{fig:egs}, we show examples of the prediction results on all three datasets. While the results illustrated in Fig.~\ref{fig:egs} are very promising, we can observe following error patterns: \begin{inparaenum}[1)]
\item The uniform sky luminance assumption is violated.
\item Single big object occupy in the photo failing the depth estimator.
\item The ground truth label is wrong.
\end{inparaenum}

\begin{table}[t]
    \centering
    \begin{tabular}{ccccc}
        \% & \textbf{FRIDA1} & \textbf{FRIDA2} & \textbf{PM25} & \textbf{PM25-s} \\
        \textsf{jcsb2014 \cite{jcsb.1000161}} & 77.34 & 77.44 & 3.95 & N/A \\
        \textsf{depth} & 76.74 & 53.47 & 25.32 & 70.14 \\
        \textsf{trans} & 85.56 & 87.38 & 28.10 & 84.32 \\
        \textsf{depth$\otimes$trans} & \textbf{90.60} & \textbf{87.43} & \textbf{40.83} & \textbf{89.05} \\
    \end{tabular}
    \caption{Absolute Spearman correlation coefficients (\%) performance. First row show the datasets, first column show the methods and \textsf{depth$\otimes$trans} is the proposed method. All shown values have a p-value smaller than 0.001. \textsf{jcsb2014} for \textbf{PM25-s} is N/A, because the p-value is 0.3781.}
    \label{tab:result}
\end{table}


\section{Conclusions}
\label{sec:conclude}

We have proposed an effective method to estimate haze level from single images. The input image is first fed into a haze removal algorithm to generate the transmission matrix, and the depth map is also estimated from the pixels. By removing the effects of depth, we estimate the haze level from the transmission matrix. Using a GPU backend of the Deep Convolutional Neural Fields, the whole processing time for one image is less than one second. The superior performance of combining the transmission matrix and depth map is validated by the experiment results of Spearman correlation between the estimated haze level and ground truth on both synthetic and real image datasets. The results on real image dataset need further research to make large scale monitoring based on online user photos more reliable, e.g, defining a better proxy for the ground truth haze level. In order to encourage future research, we will release datasets online \footnote{\url{https://goo.gl/kmdd2M}}.

\newcommand{\cowidth}{.24\columnwidth}
\newcommand{\cofwidth}{.24\columnwidth}
\newcommand{\cosubwidth}{1\textwidth}
\begin{figure}[t]
    \centering
    \begin{subfigure}[b]{\cofwidth}
        \includegraphics[width=\cosubwidth]{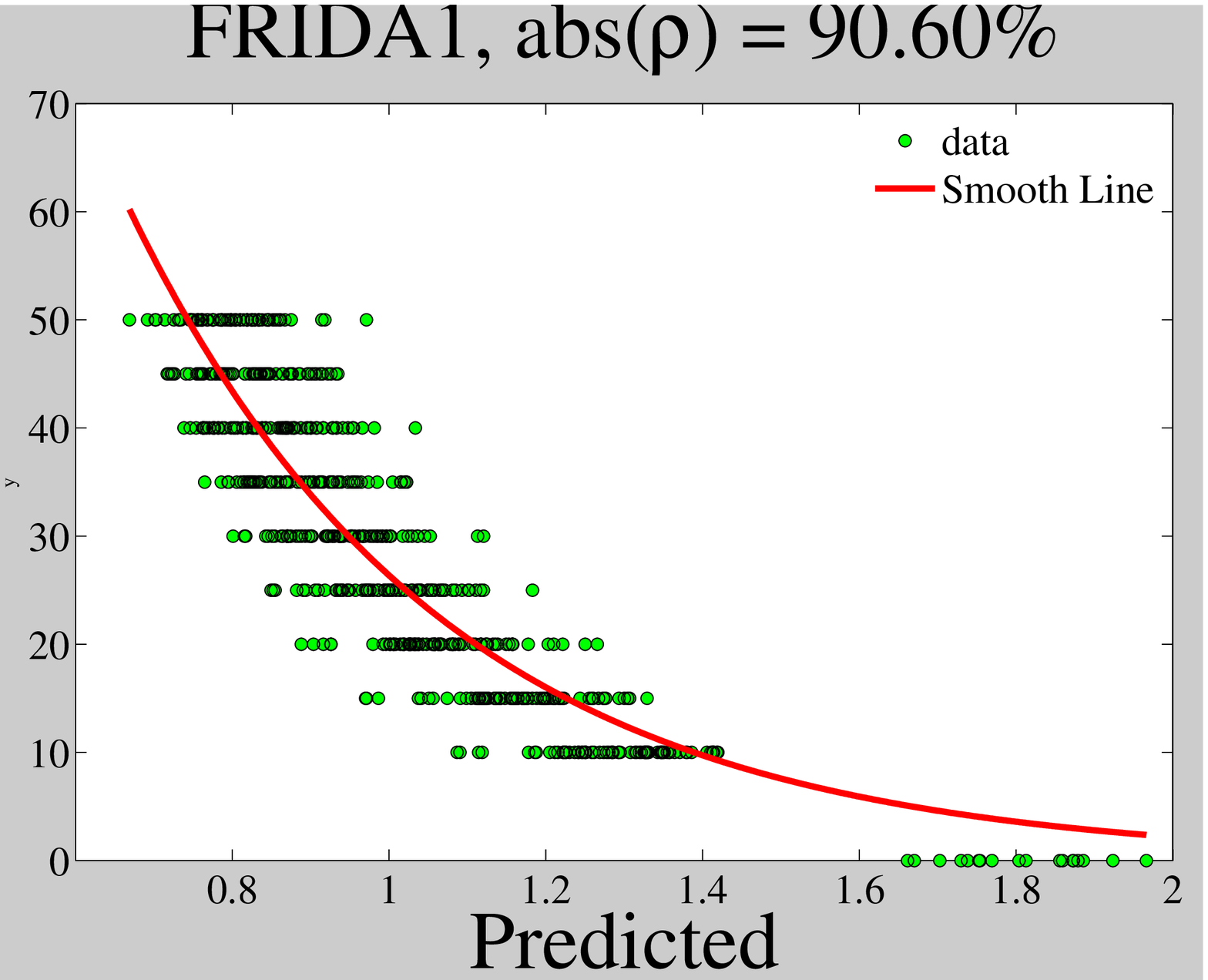}
    \end{subfigure}
    \begin{subfigure}[b]{\cowidth}
        \includegraphics[width=\cosubwidth]{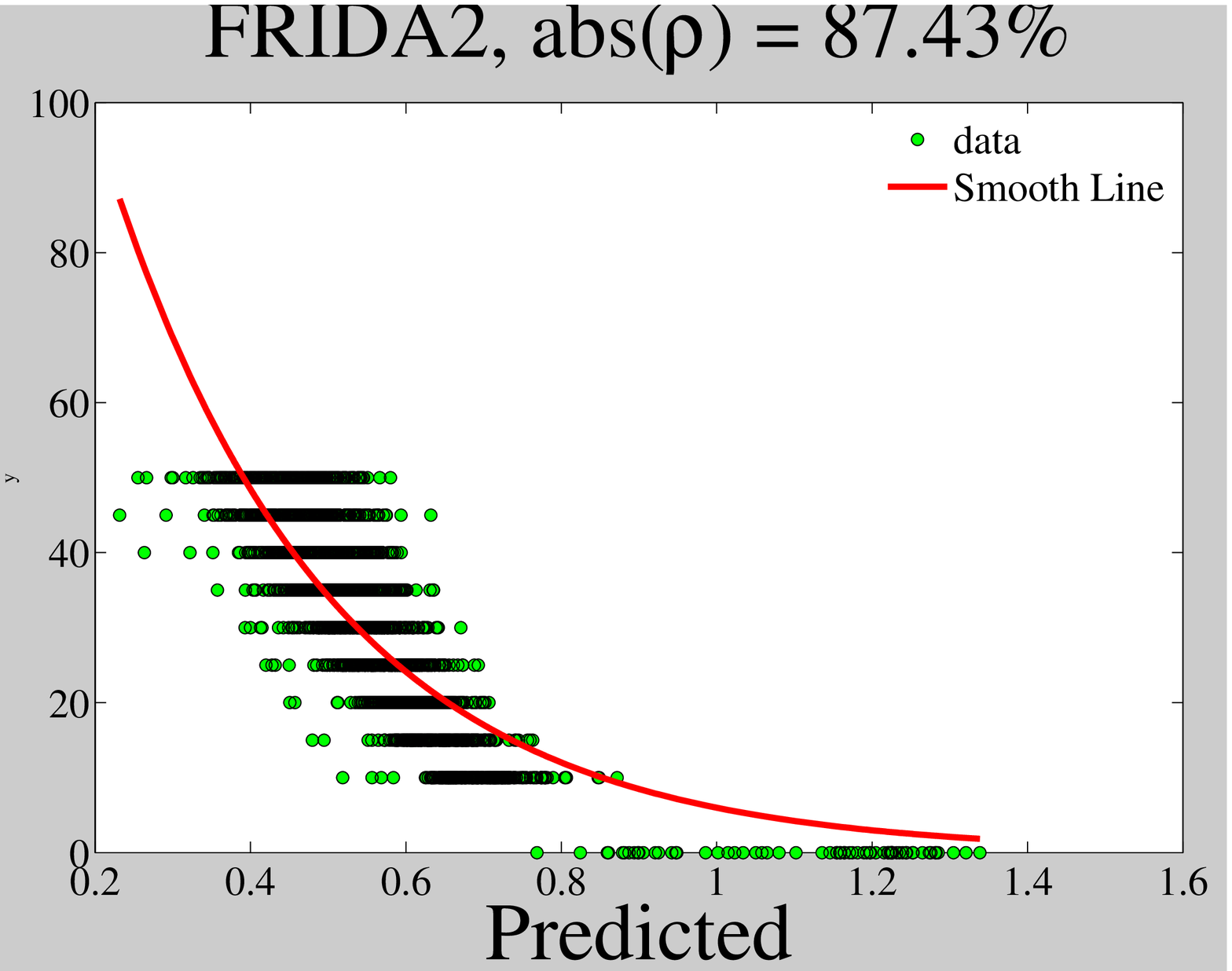}
    \end{subfigure}
    \begin{subfigure}[b]{\cowidth}
        \includegraphics[width=\cosubwidth]{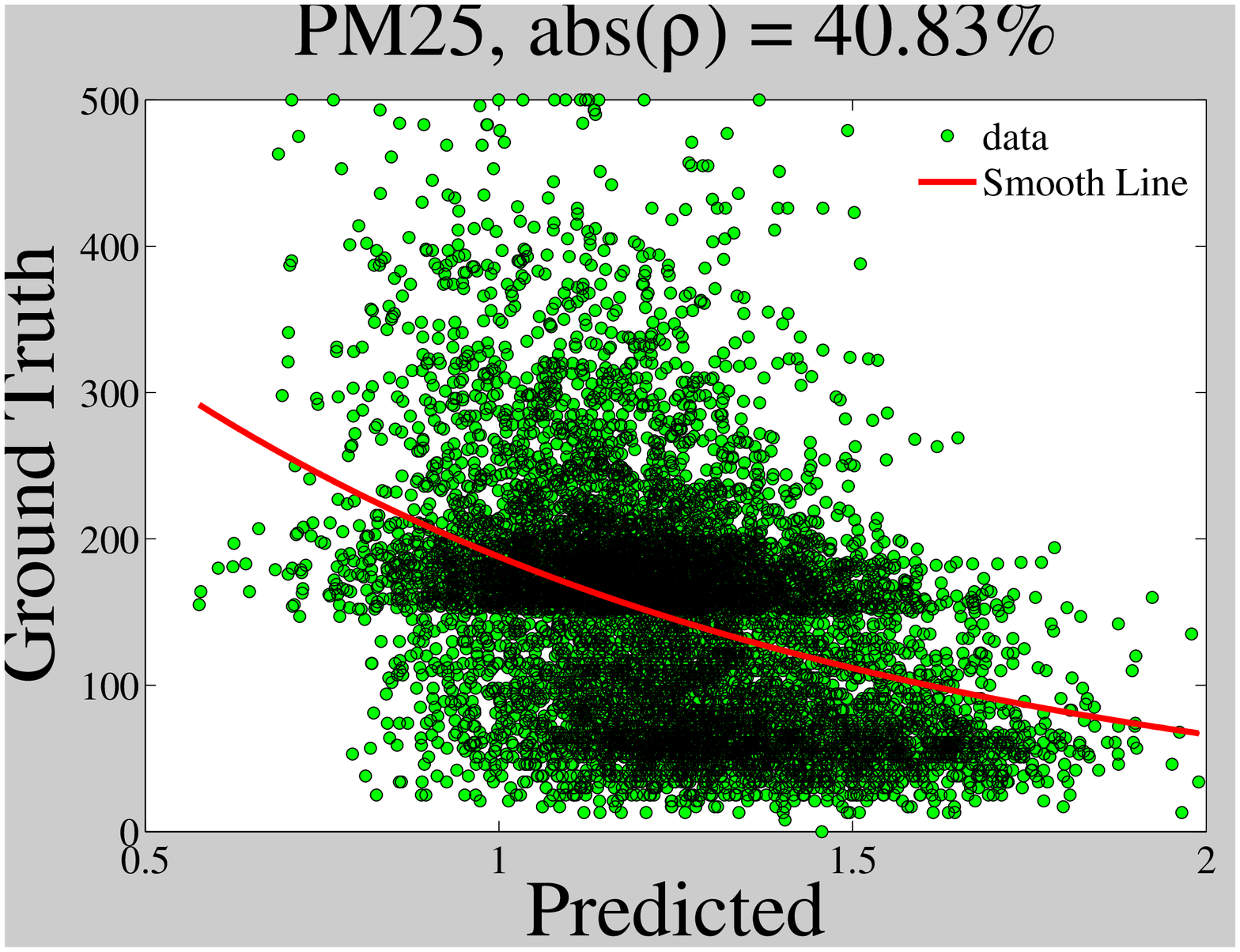}
    \end{subfigure}
    \begin{subfigure}[b]{\cowidth}
        \includegraphics[width=\cosubwidth]{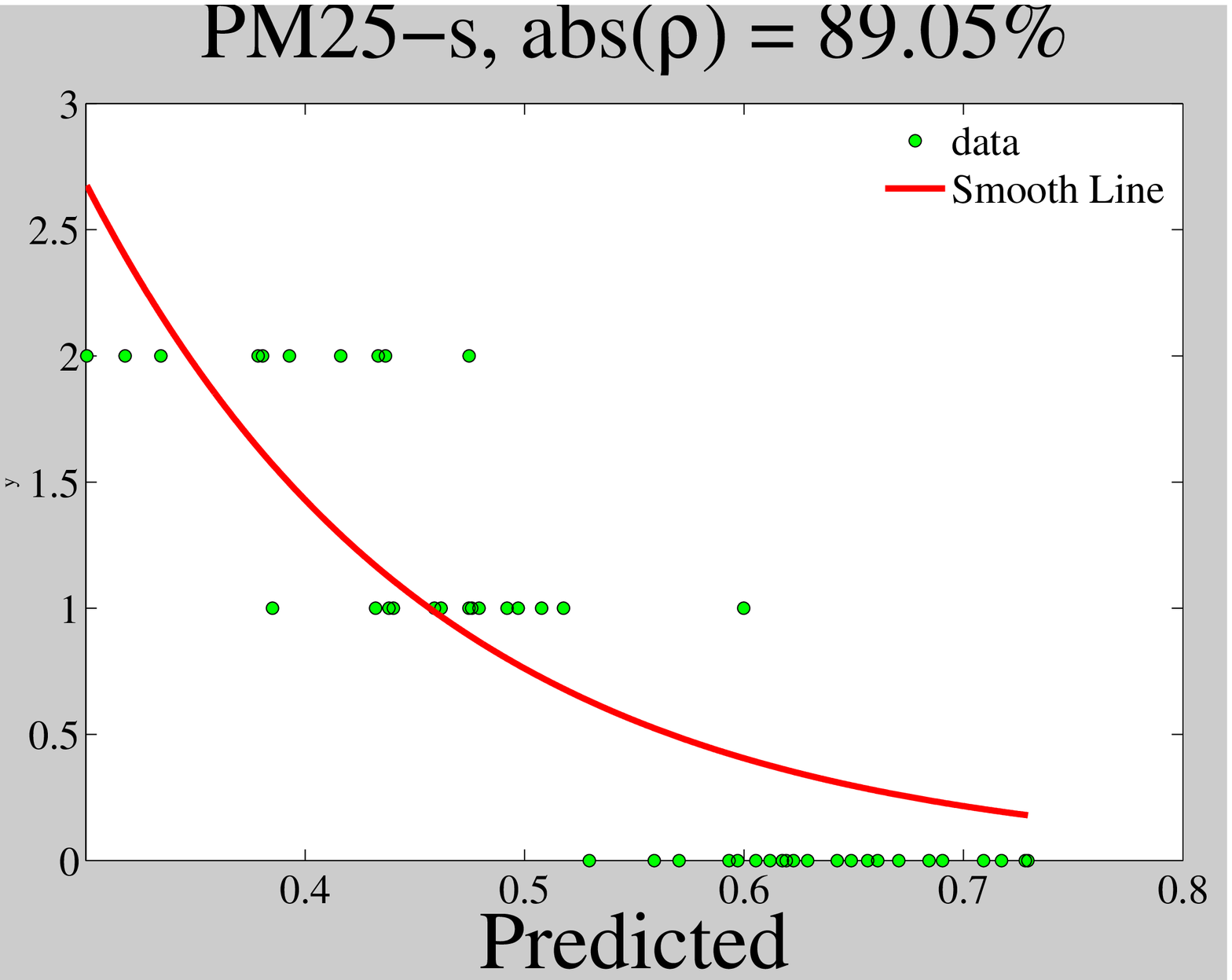}
    \end{subfigure}
    \caption{The ground truth and predicted haze level for the \textsf{depth$\otimes$trans} on each dataset. The points are the data instances and the line is a fitted trendline to illustrate the correlation.}
    \label{fig:corr}
\end{figure}

\newcommand{\egswidth}{.15\textwidth}
\newcommand{\egssubwidth}{1\textwidth}
\newcommand{\egssubheight}{1\textwidth}
\setlength\fboxsep{0pt}
\setlength\fboxrule{0pt}
\newcommand{\myfbox}[1]{\fcolorbox{green}{green}{#1}}
\newcommand{\mytbox}[1]{\fbox{#1}}

\begin{figure*}[t]
    \centering
    \caption{The examples of photos from the datasets at different haze level, alone with its the prediction results. The rows are from dataset \textbf{PM25}, \textbf{PM25}, \textbf{PM25}, \textbf{PM25-s}, \textbf{PM25-s}, \textbf{PM25-s}, \textbf{FRIDA1} and \textbf{FRIDA2}, respectively. The prediction errors are highlighted with thick green borders and they are analysed in the end of the Experiment section.}
    \label{fig:egs}
        \begin{subfigure}[b]{\egswidth}
    \mytbox{
        \begin{overpic}[width=\egssubwidth,height=\egssubheight]{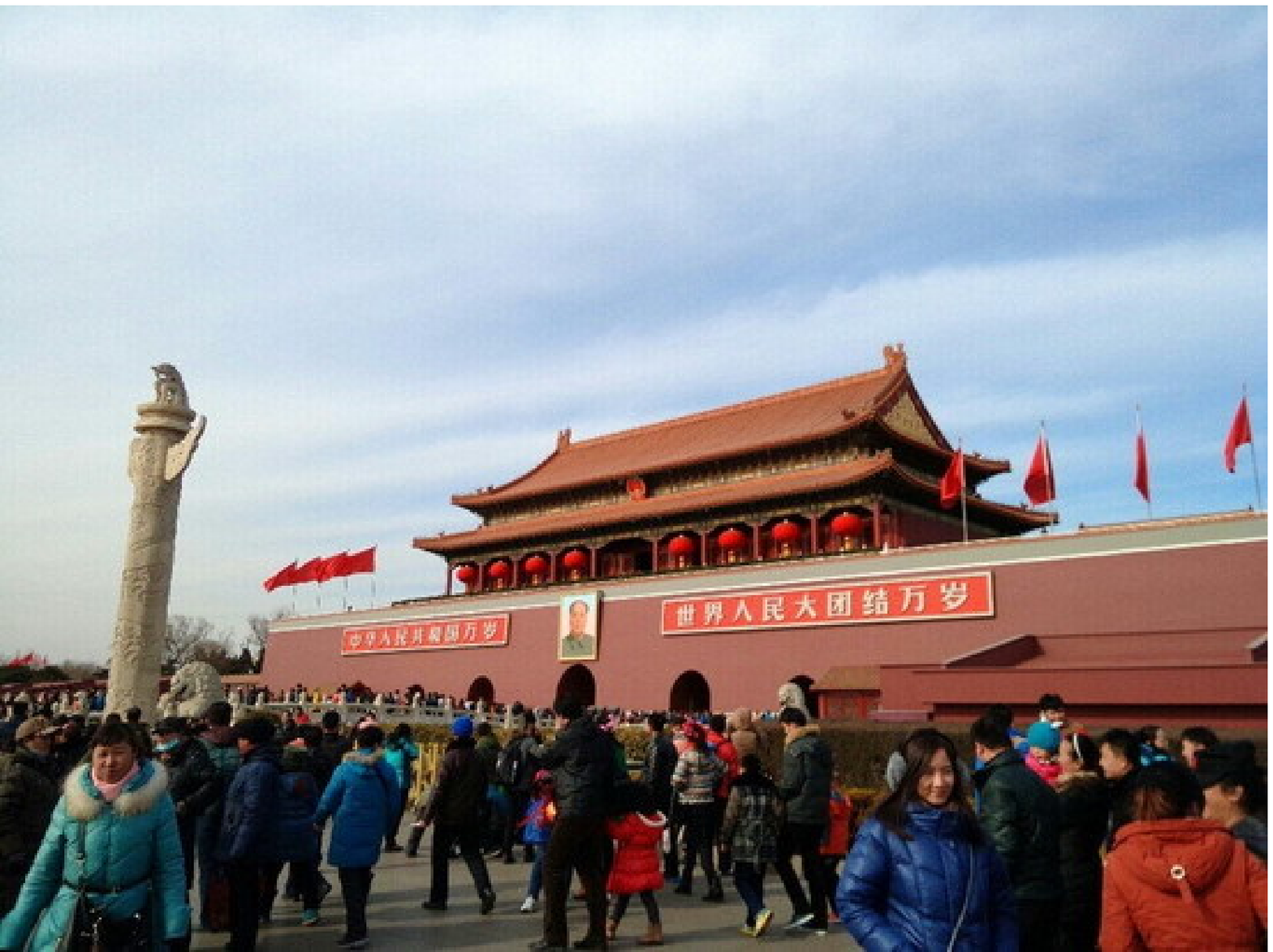}
         \put (20,90) {\red\textbf\small GT: Clear}
         \put (20,80) {\red\textbf\small Pred: Clear}
         \end{overpic}
    }
    \end{subfigure}
~    \begin{subfigure}[b]{\egswidth}
    \mytbox{
        \begin{overpic}[width=\egssubwidth,height=\egssubheight]{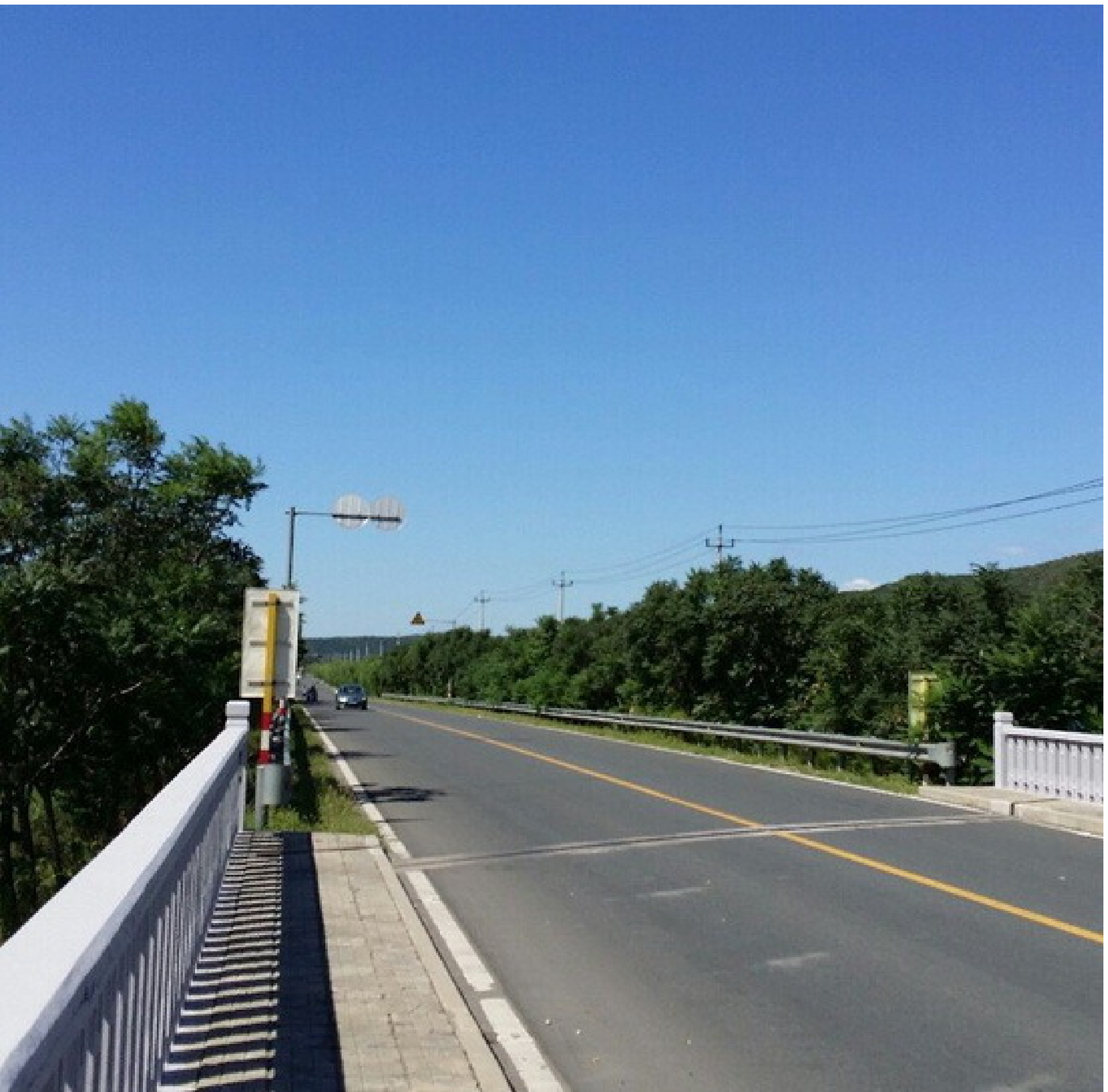}
         \put (20,90) {\red\textbf\small GT: Clear}
         \put (20,80) {\red\textbf\small Pred: Clear}
         \end{overpic}
    }
    \end{subfigure}
~    \begin{subfigure}[b]{\egswidth}
    \mytbox{
        \begin{overpic}[width=\egssubwidth,height=\egssubheight]{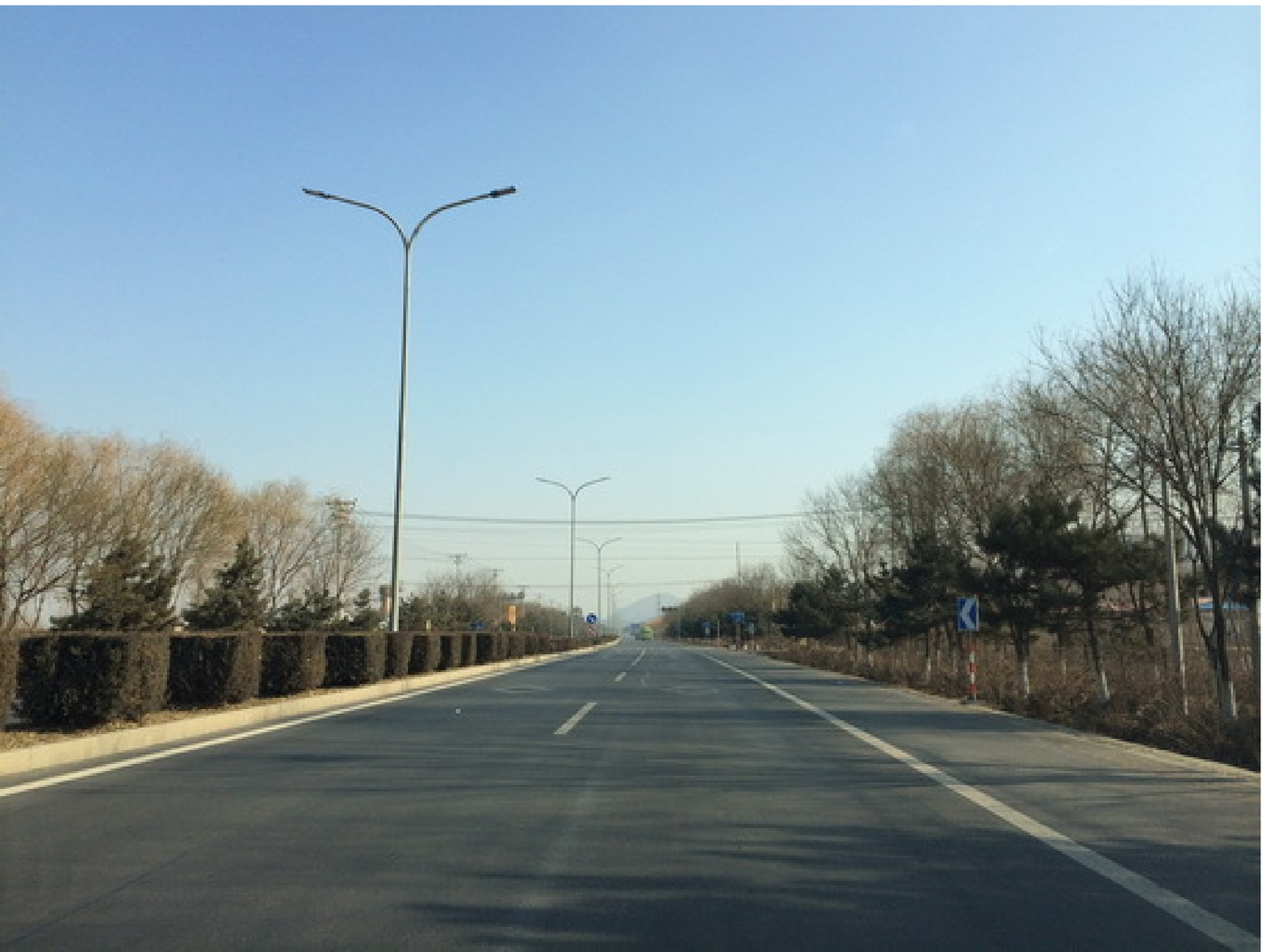}
         \put (20,90) {\red\textbf\small GT: Clear}
         \put (20,80) {\red\textbf\small Pred: Clear}
         \end{overpic}
    }
    \end{subfigure}
~    \begin{subfigure}[b]{\egswidth}
    \mytbox{
        \begin{overpic}[width=\egssubwidth,height=\egssubheight]{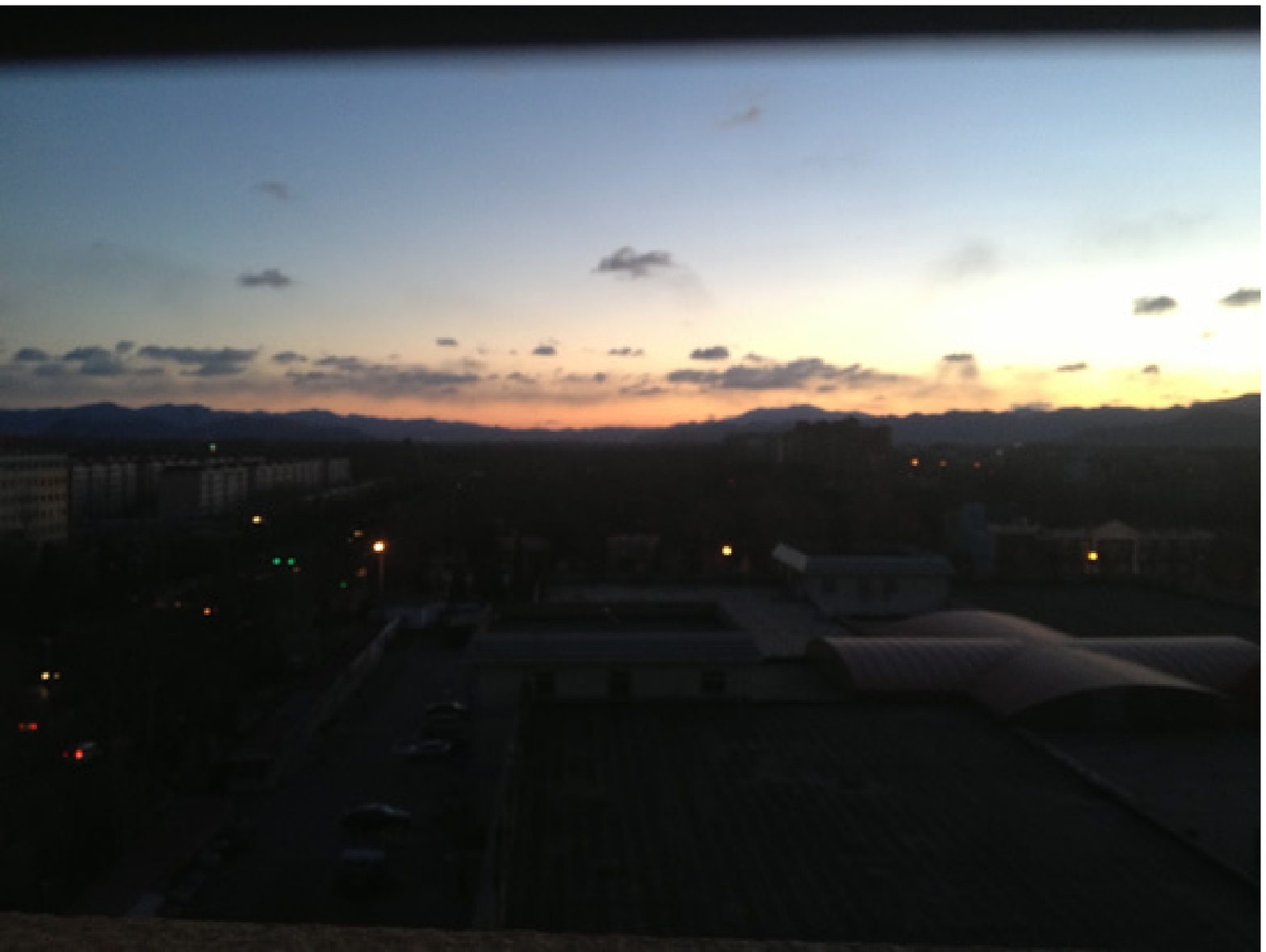}
         \put (20,90) {\red\textbf\small GT: Clear}
         \put (20,80) {\red\textbf\small Pred: Clear}
         \end{overpic}
    }
    \end{subfigure}
~    \begin{subfigure}[b]{\egswidth}
    \myfbox{
        \begin{overpic}[width=\egssubwidth,height=\egssubheight]{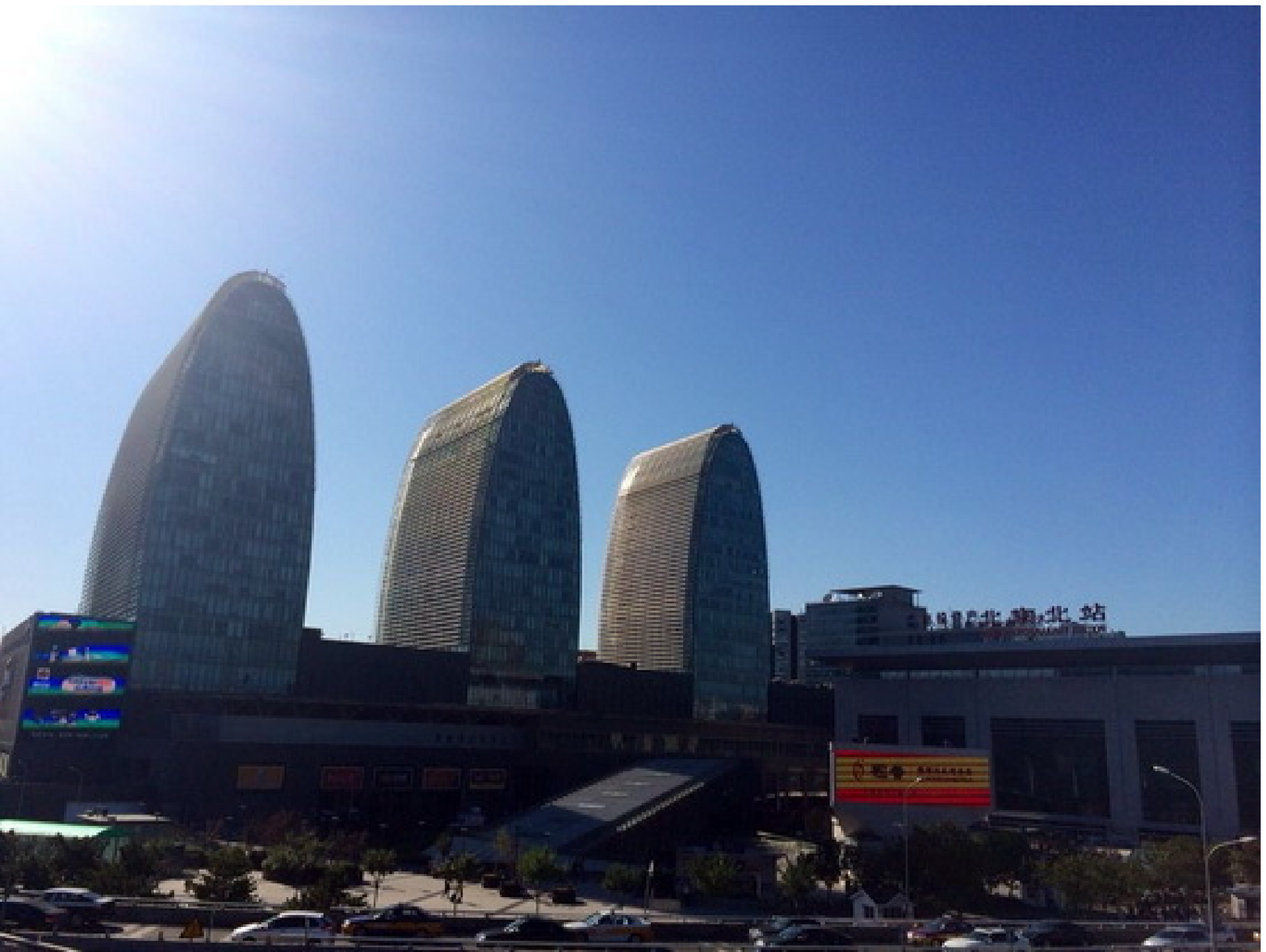}
         \put (20,90) {\red\textbf\small GT: Clear}
         \put (20,80) {\red\textbf\small Pred: Light}
         \end{overpic}
    }
    \end{subfigure}
~    \begin{subfigure}[b]{\egswidth}
    \mytbox{
        \begin{overpic}[width=\egssubwidth,height=\egssubheight]{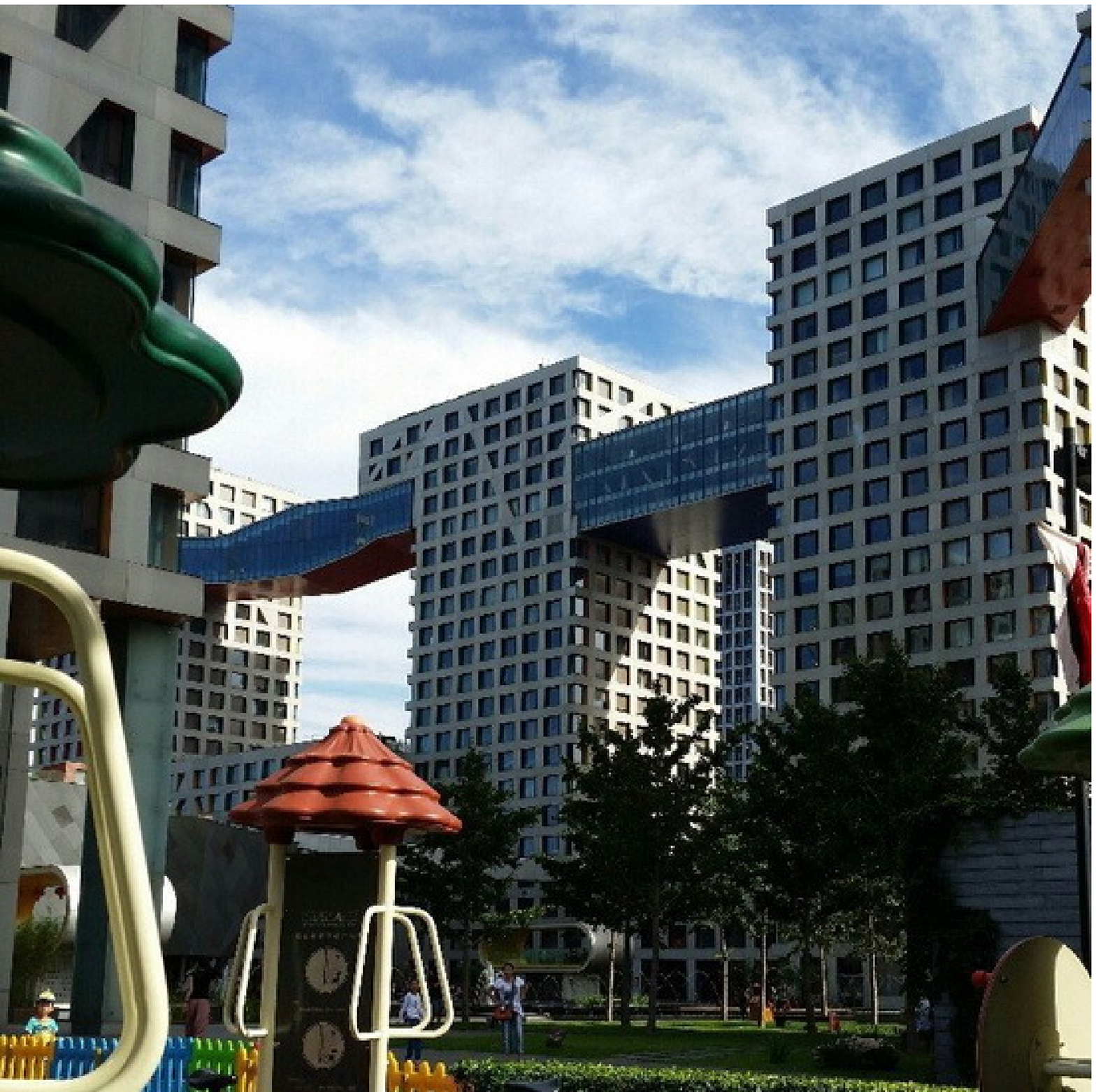}
         \put (20,90) {\red\textbf\small GT: Clear}
         \put (20,80) {\red\textbf\small Pred: Clear}
         \end{overpic}
    }
    \end{subfigure}
~    \begin{subfigure}[b]{\egswidth}
    \myfbox{
        \begin{overpic}[width=\egssubwidth,height=\egssubheight]{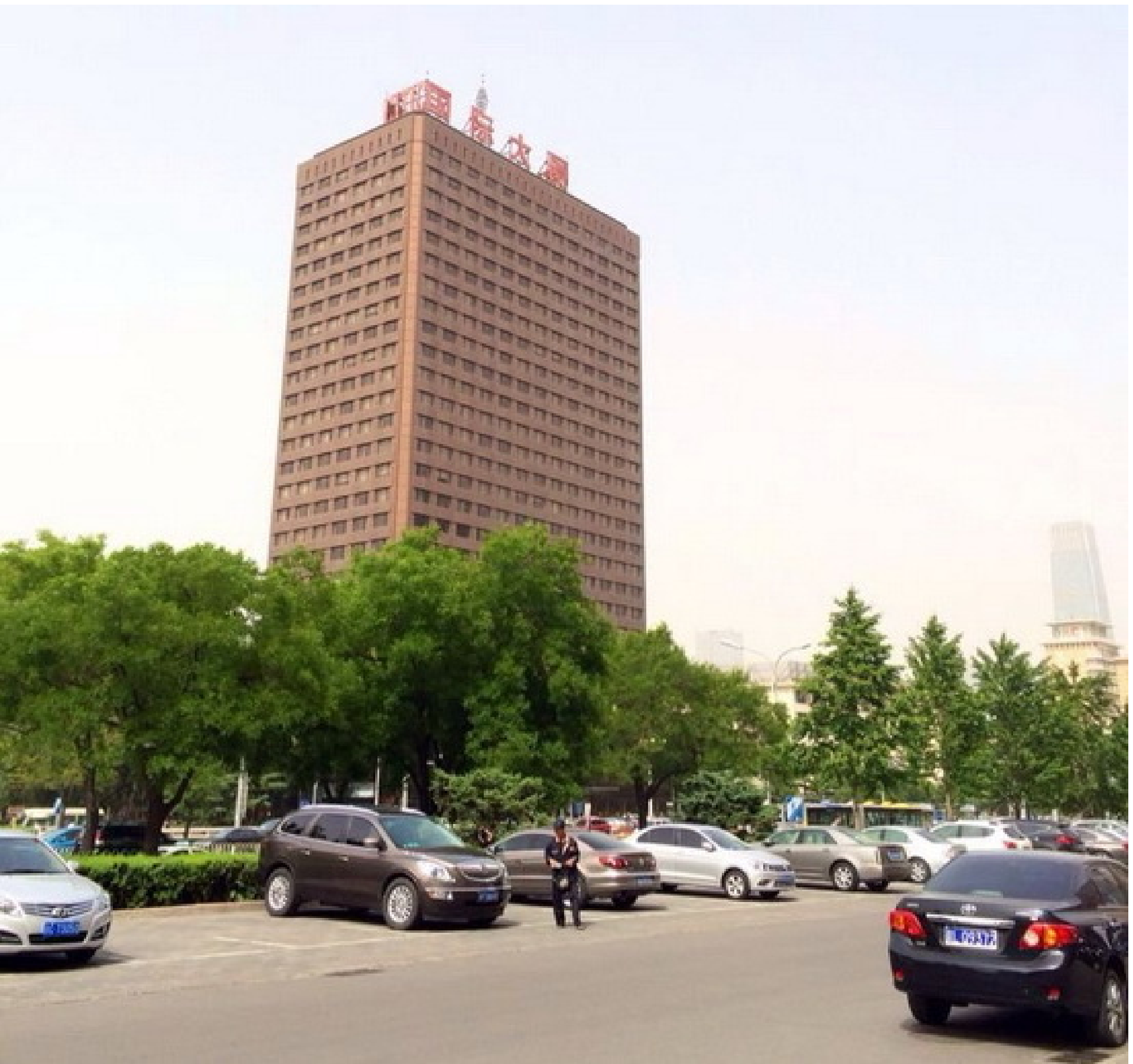}
         \put (20,90) {\red\textbf\small GT: Light}
         \put (20,80) {\red\textbf\small Pred: Heavy}
         \end{overpic}
    }
    \end{subfigure}
~    \begin{subfigure}[b]{\egswidth}
    \mytbox{
        \begin{overpic}[width=\egssubwidth,height=\egssubheight]{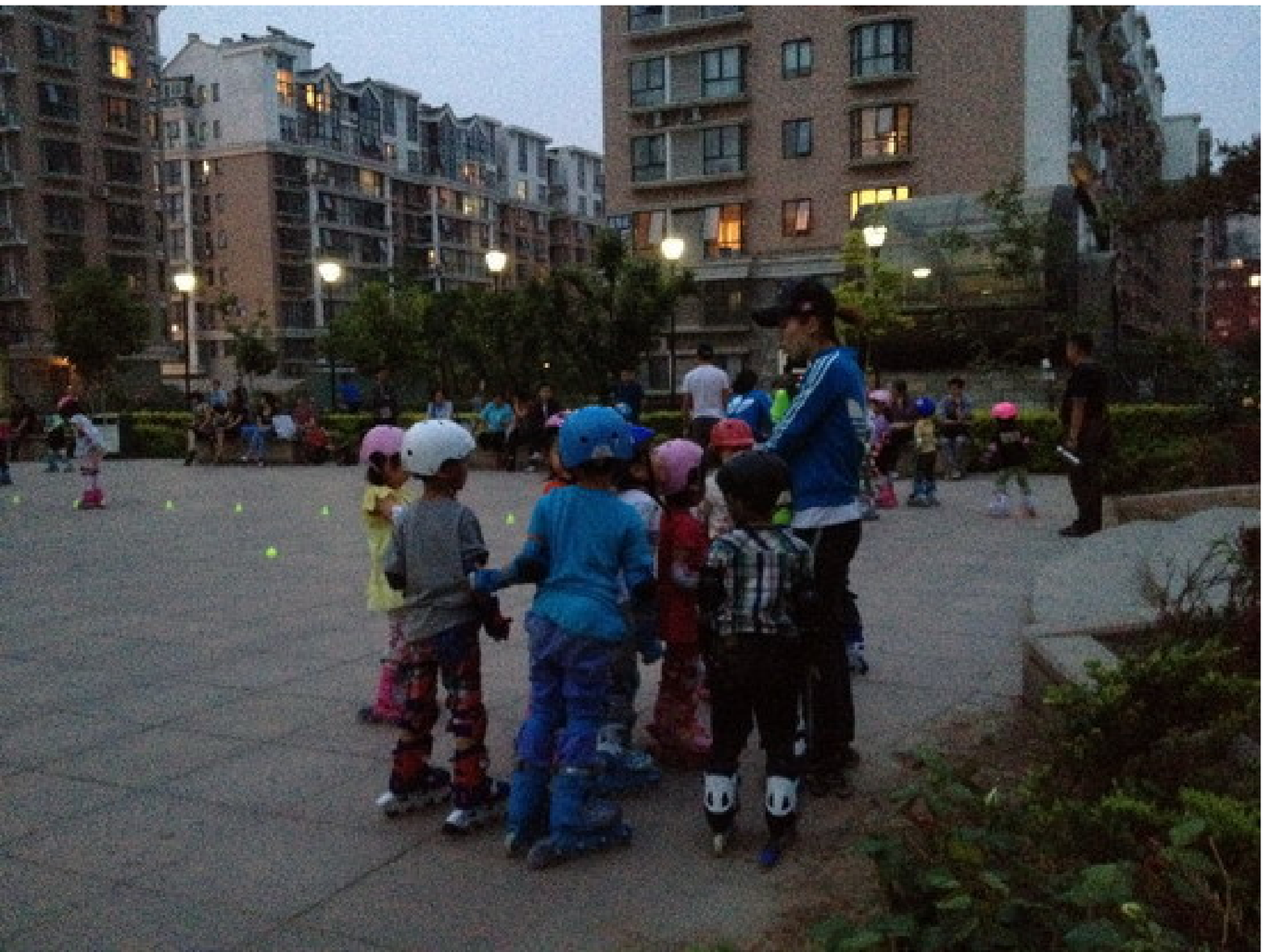}
         \put (20,90) {\red\textbf\small GT: Light}
         \put (20,80) {\red\textbf\small Pred: Light}
         \end{overpic}
    }
    \end{subfigure}
~    \begin{subfigure}[b]{\egswidth}
    \myfbox{
        \begin{overpic}[width=\egssubwidth,height=\egssubheight]{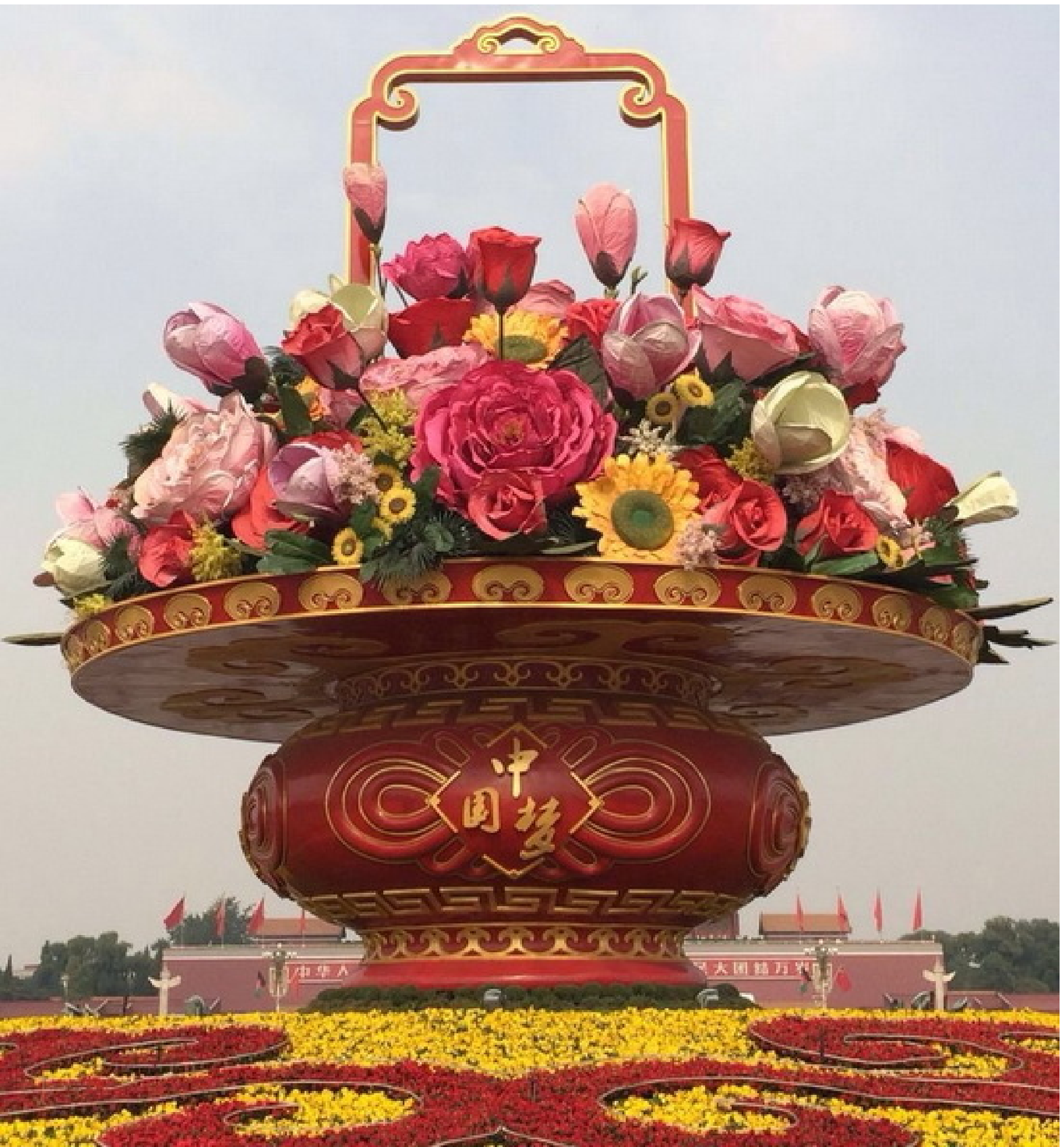}
         \put (20,90) {\red\textbf\small GT: Light}
         \put (20,80) {\red\textbf\small Pred: Clear}
         \end{overpic}
    }
    \end{subfigure}
~    \begin{subfigure}[b]{\egswidth}
    \mytbox{
        \begin{overpic}[width=\egssubwidth,height=\egssubheight]{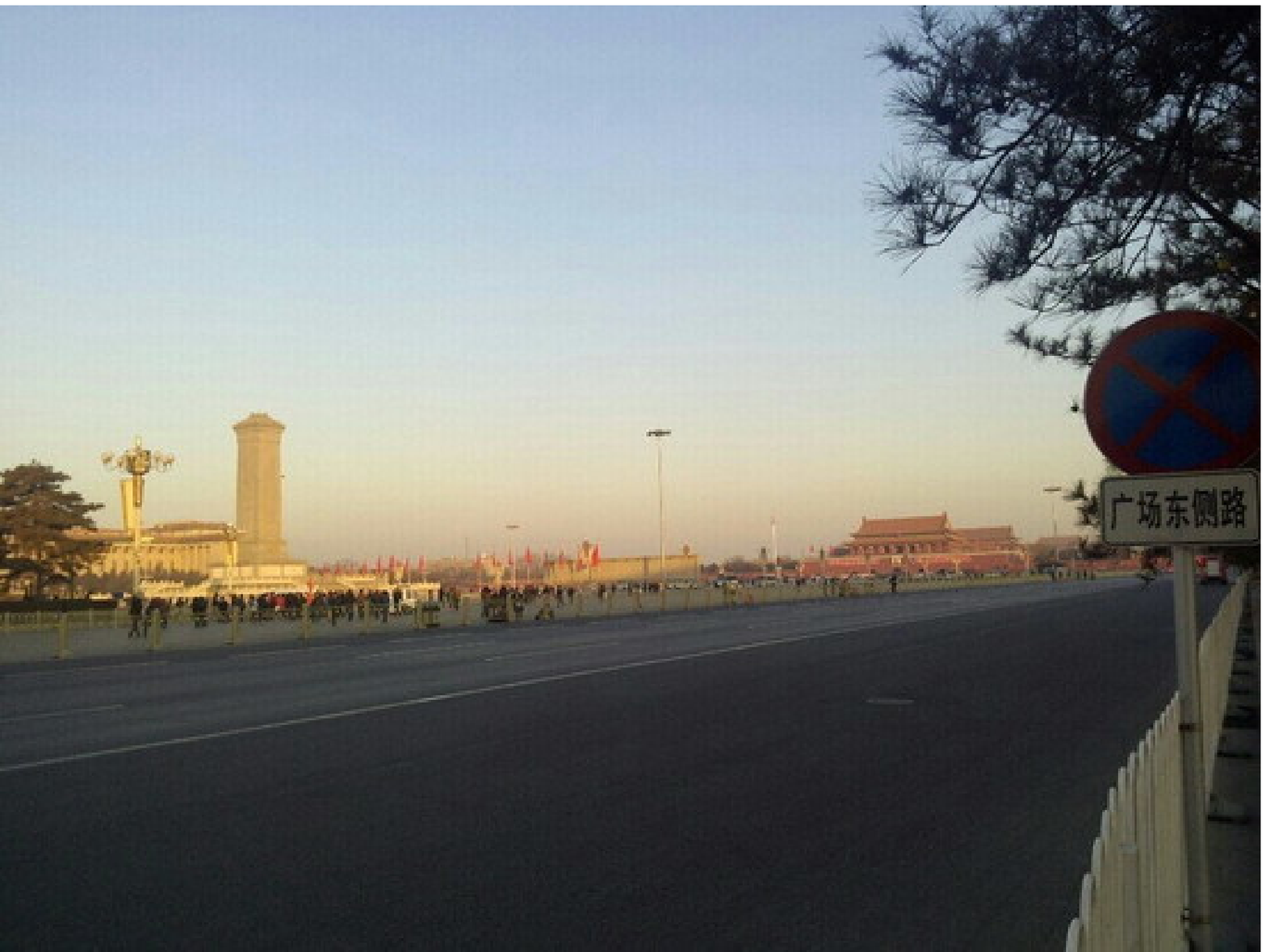}
         \put (20,90) {\red\textbf\small GT: Light}
         \put (20,80) {\red\textbf\small Pred: Light}
         \end{overpic}
    }
    \end{subfigure}
~    \begin{subfigure}[b]{\egswidth}
    \mytbox{
        \begin{overpic}[width=\egssubwidth,height=\egssubheight]{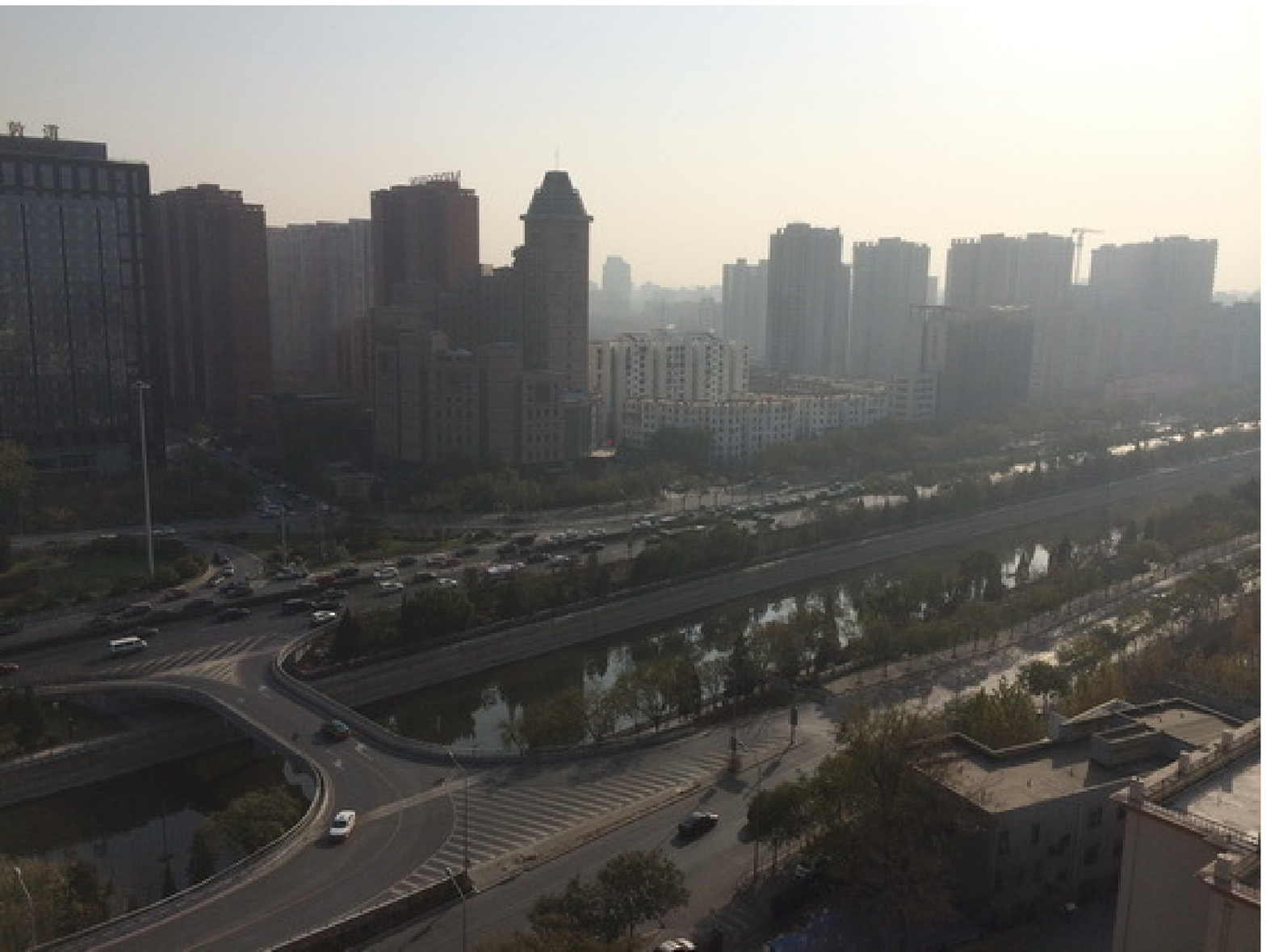}
         \put (20,90) {\red\textbf\small GT: Light}
         \put (20,80) {\red\textbf\small Pred: Light}
         \end{overpic}
    }
    \end{subfigure}
~    \begin{subfigure}[b]{\egswidth}
    \mytbox{
        \begin{overpic}[width=\egssubwidth,height=\egssubheight]{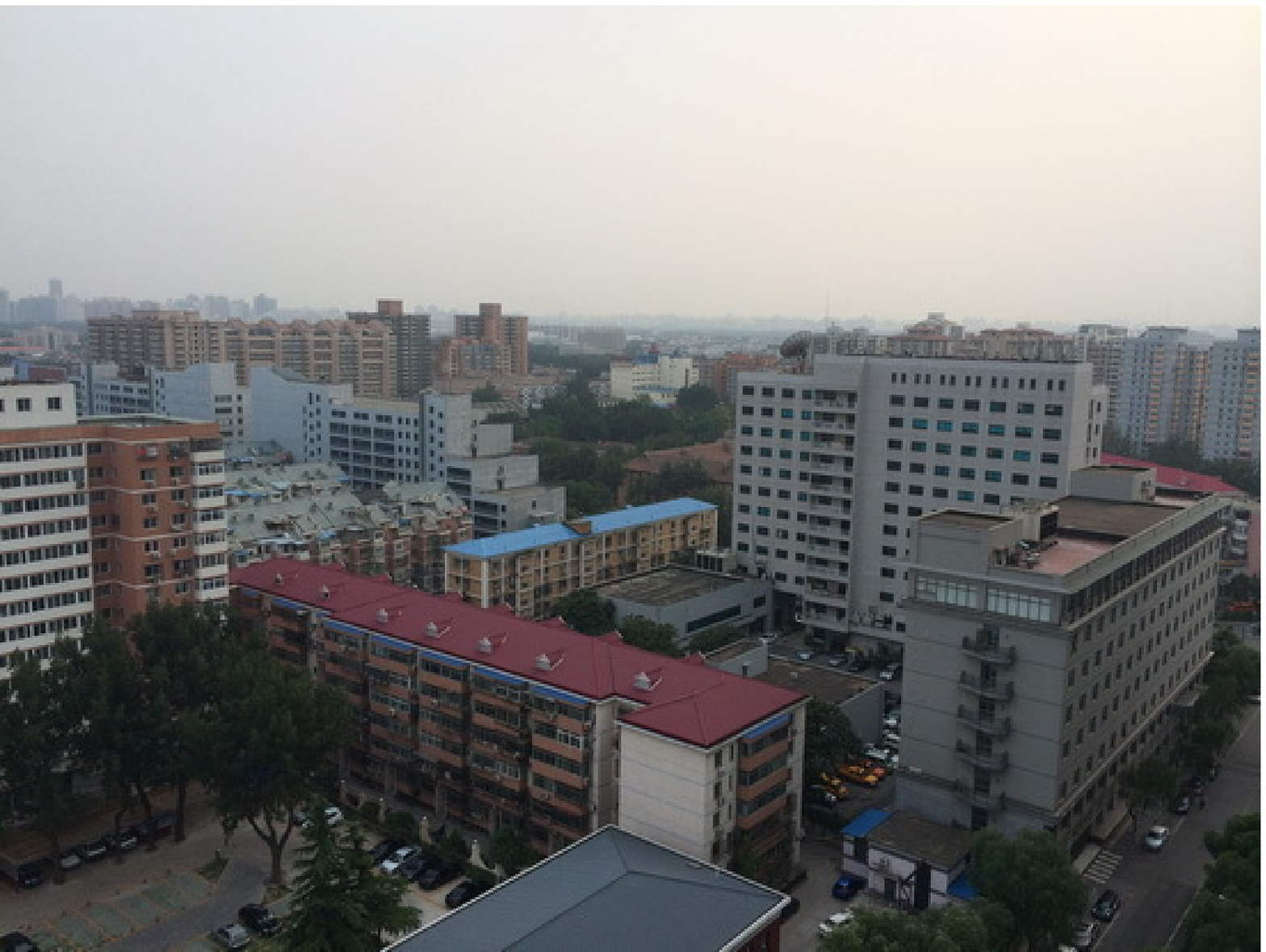}
         \put (20,90) {\red\textbf\small GT: Light}
         \put (20,80) {\red\textbf\small Pred: Light}
         \end{overpic}
    }
    \end{subfigure}
~    \begin{subfigure}[b]{\egswidth}
    \myfbox{
        \begin{overpic}[width=\egssubwidth,height=\egssubheight]{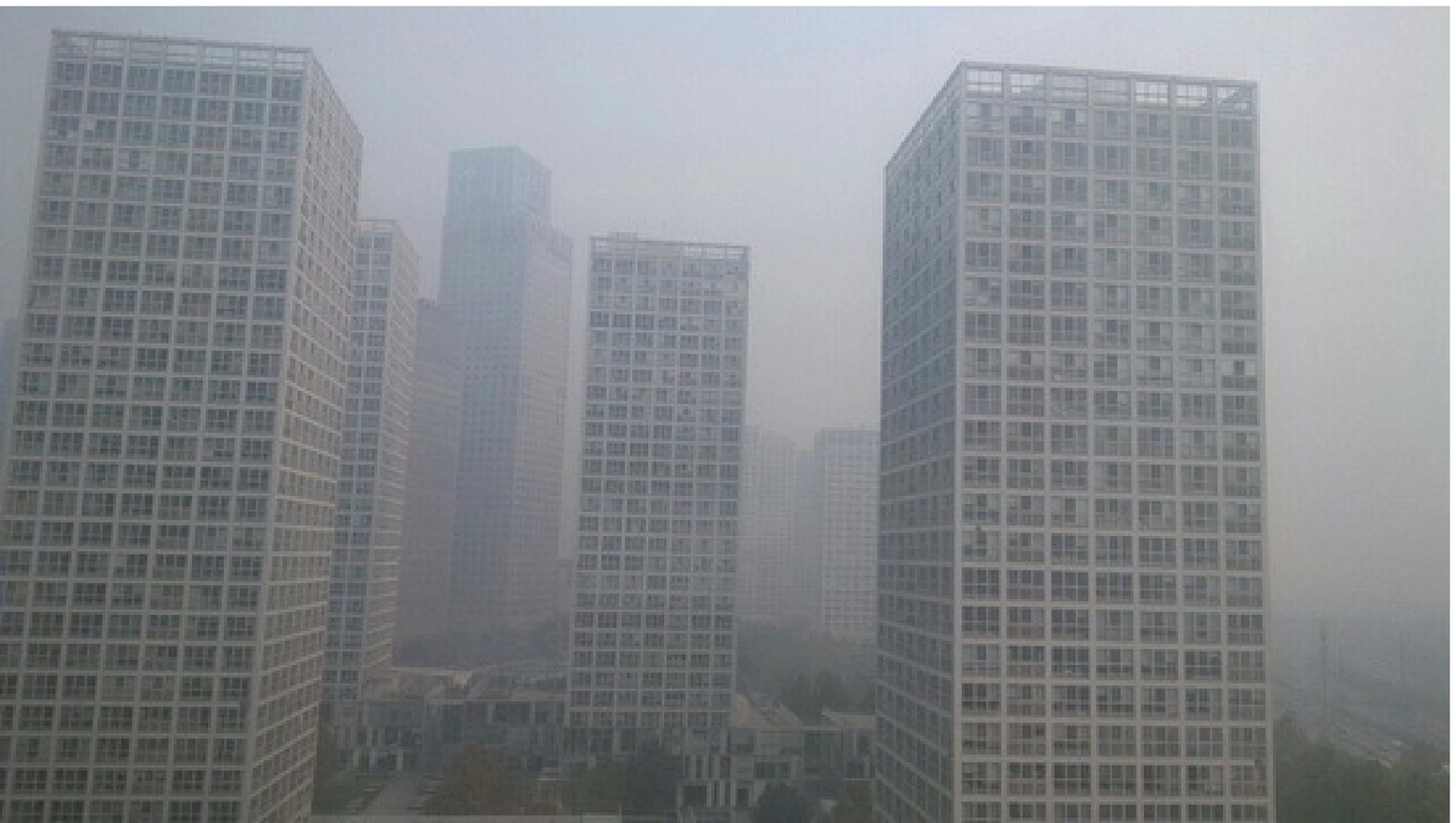}
         \put (20,90) {\red\textbf\small GT: Heavy}
         \put (20,80) {\red\textbf\small Pred: Light}
         \end{overpic}
    }
    \end{subfigure}
~    \begin{subfigure}[b]{\egswidth}
    \mytbox{
        \begin{overpic}[width=\egssubwidth,height=\egssubheight]{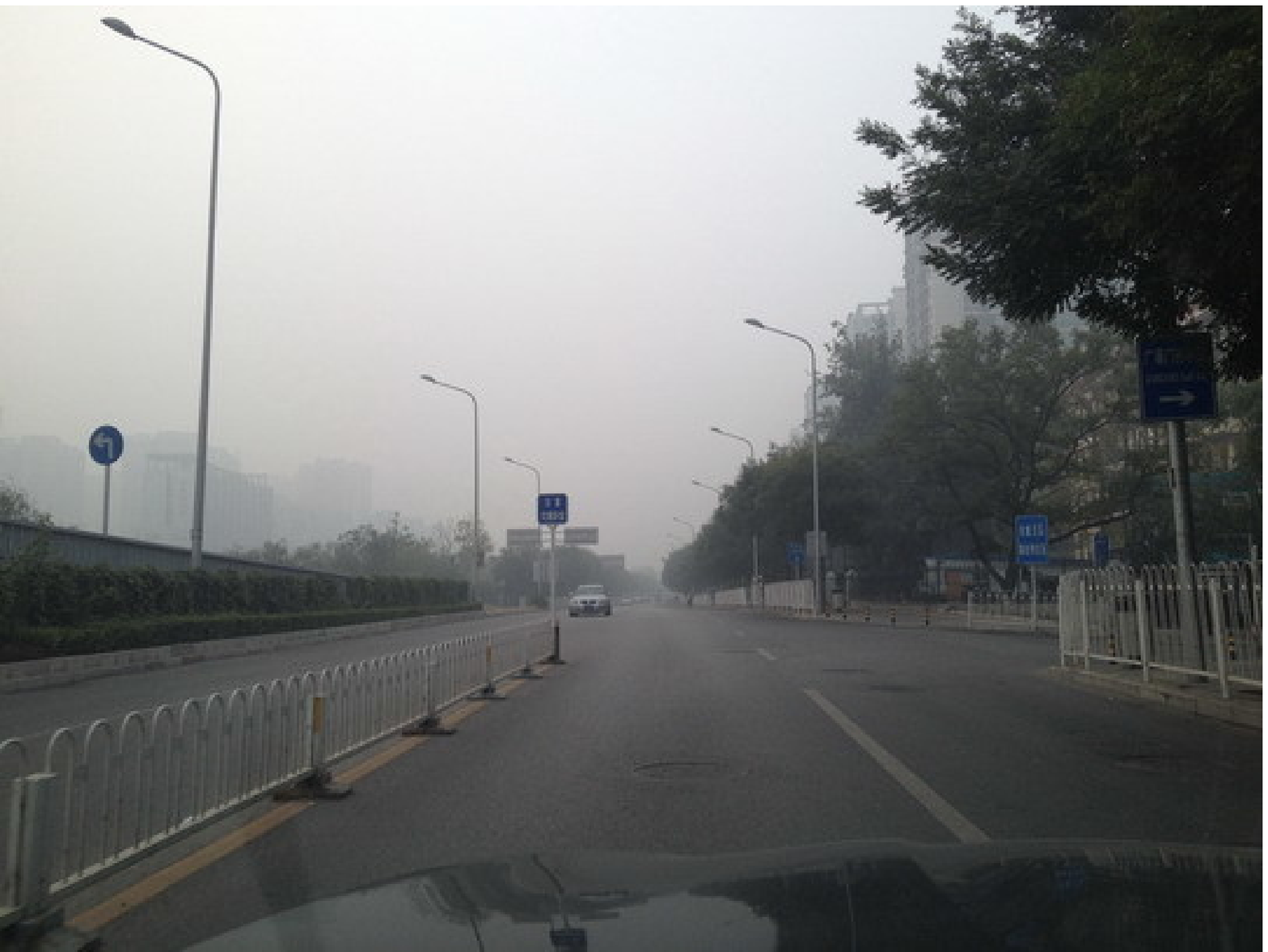}
         \put (20,90) {\red\textbf\small GT: Heavy}
         \put (20,80) {\red\textbf\small Pred: Heavy}
         \end{overpic}
    }
    \end{subfigure}
~    \begin{subfigure}[b]{\egswidth}
    \mytbox{
        \begin{overpic}[width=\egssubwidth,height=\egssubheight]{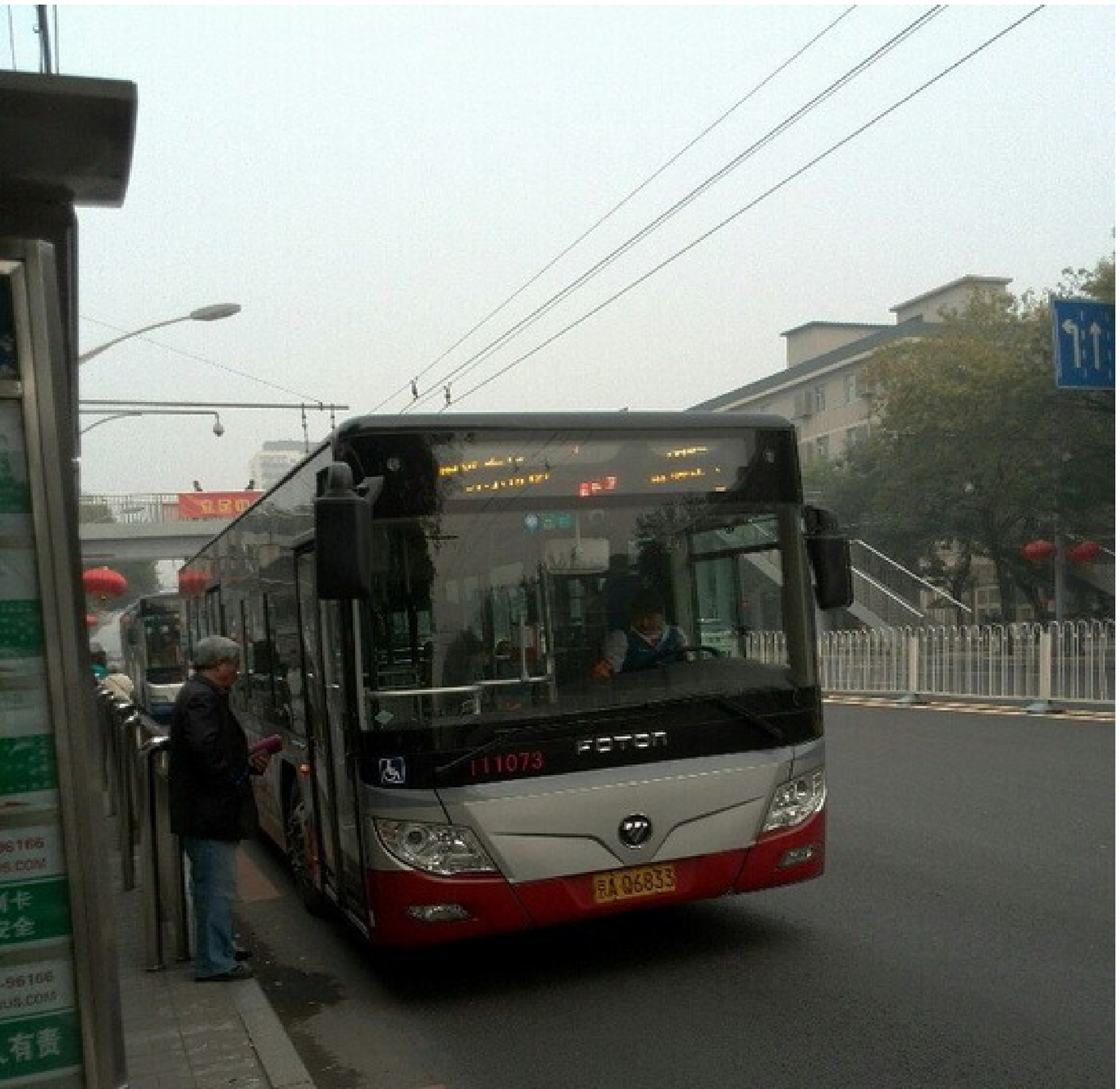}
         \put (20,90) {\red\textbf\small GT: Heavy}
         \put (20,80) {\red\textbf\small Pred: Heavy}
         \end{overpic}
    }
    \end{subfigure}
~    \begin{subfigure}[b]{\egswidth}
    \mytbox{
        \begin{overpic}[width=\egssubwidth,height=\egssubheight]{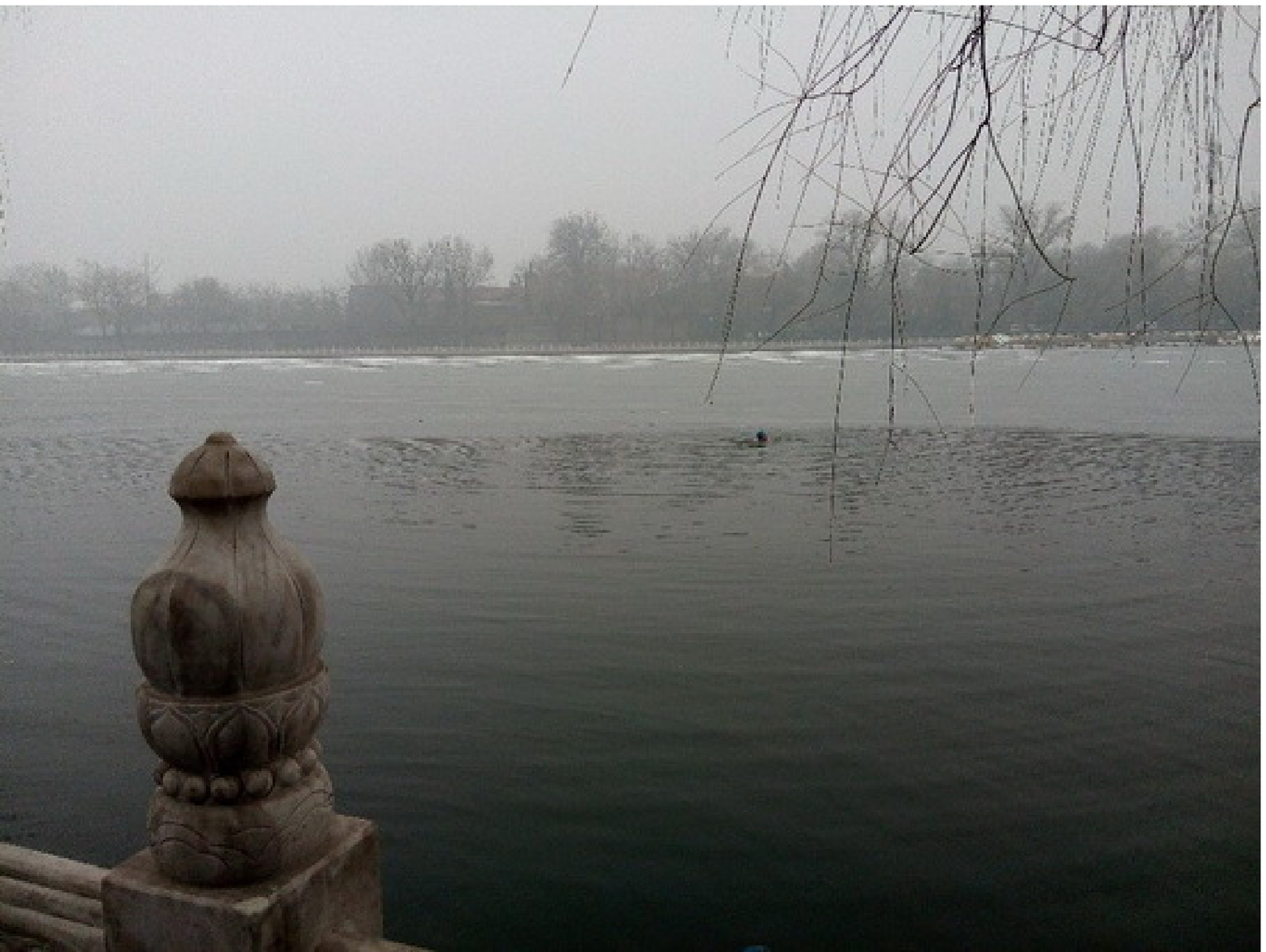}
         \put (20,90) {\red\textbf\small GT: Heavy}
         \put (20,80) {\red\textbf\small Pred: Heavy}
         \end{overpic}
    }
    \end{subfigure}
~    \begin{subfigure}[b]{\egswidth}
    \mytbox{
        \begin{overpic}[width=\egssubwidth,height=\egssubheight]{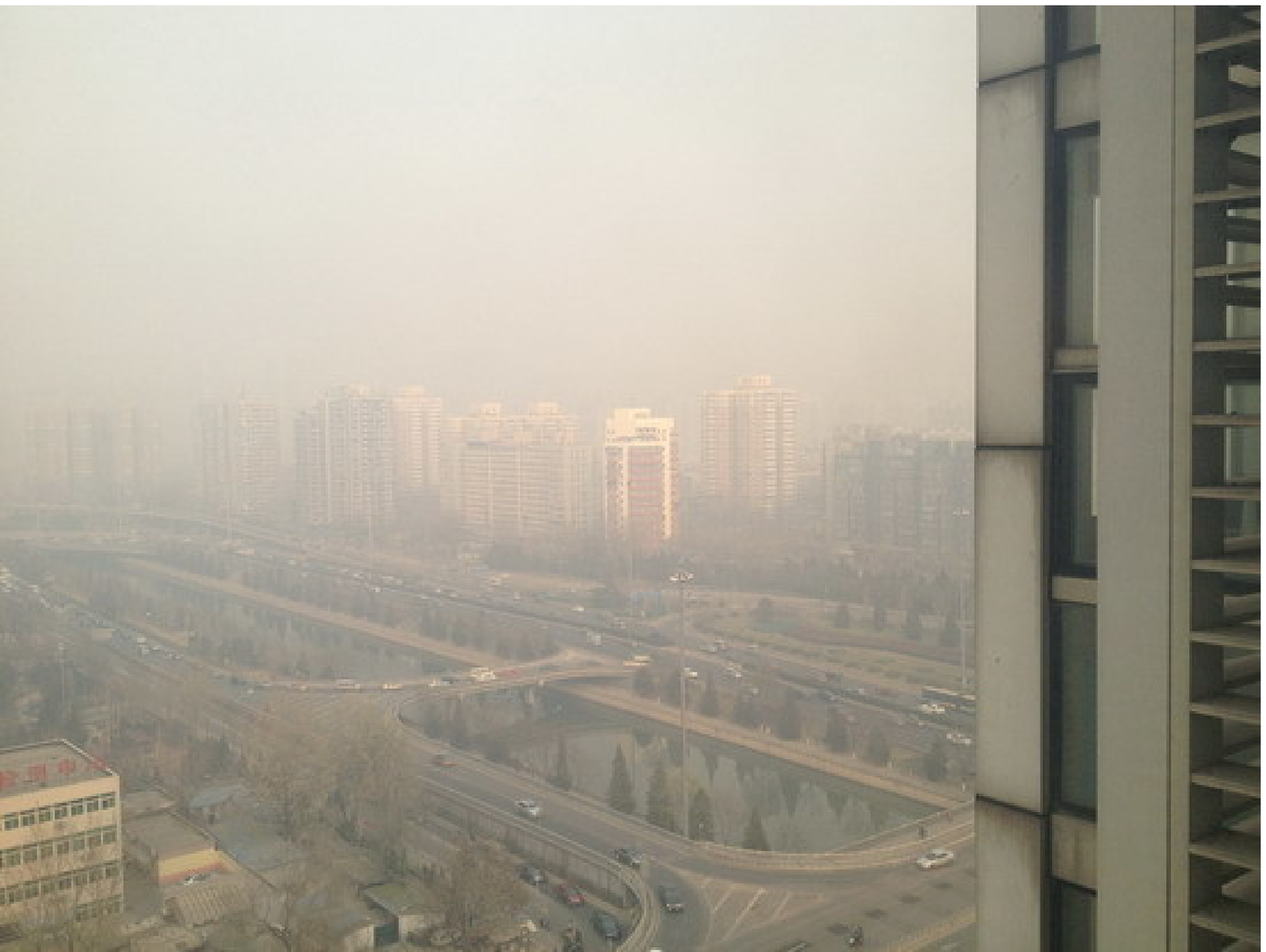}
         \put (20,90) {\red\textbf\small GT: Heavy}
         \put (20,80) {\red\textbf\small Pred: Heavy}
         \end{overpic}
    }
    \end{subfigure}
~    \begin{subfigure}[b]{\egswidth}
    \mytbox{
        \begin{overpic}[width=\egssubwidth,height=\egssubheight]{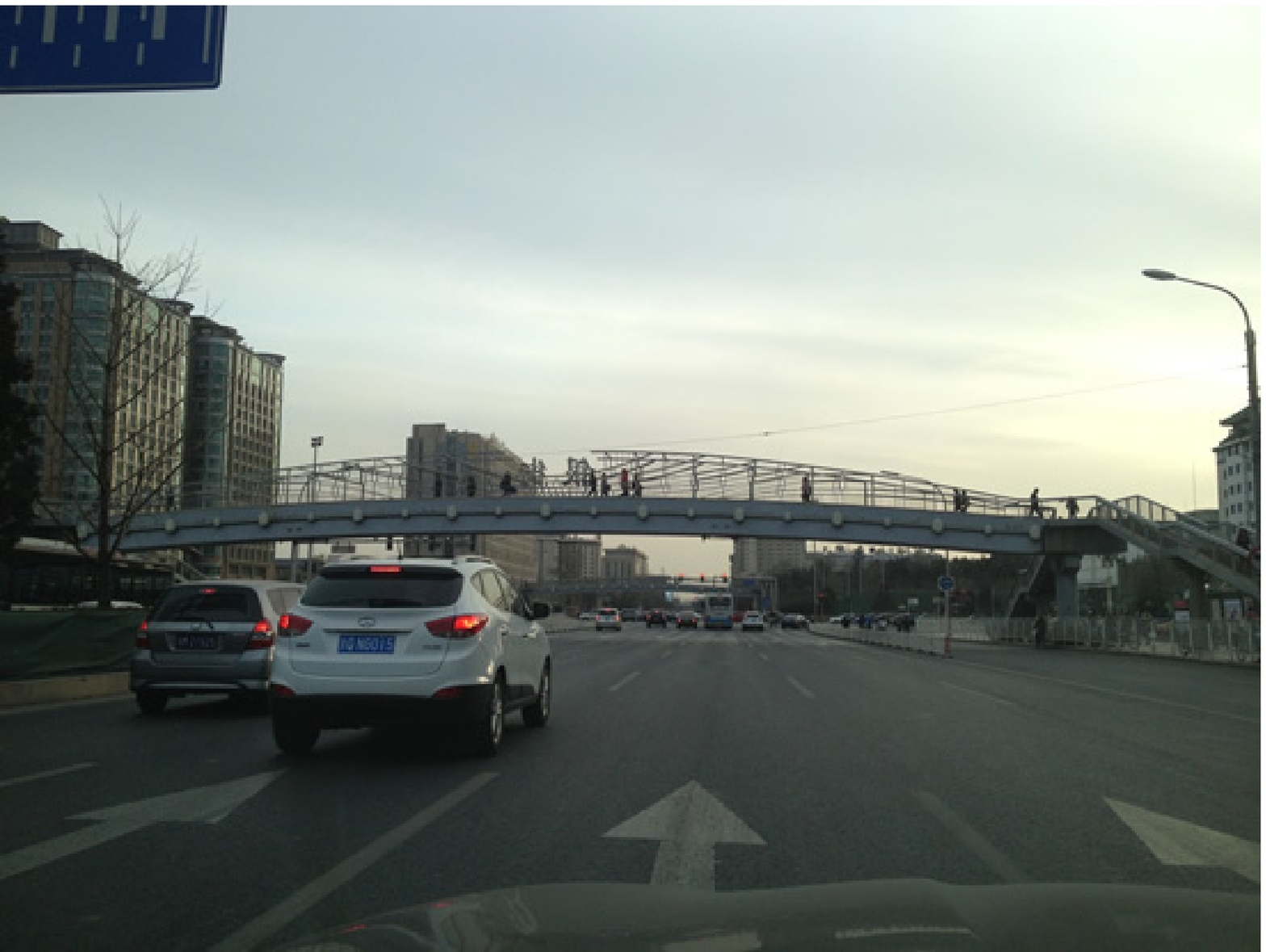}
         \put (20,90) {\red\textbf\small GT: Heavy}
         \put (20,80) {\red\textbf\small Pred: Heavy}
         \end{overpic}
    }
    \end{subfigure}
~    \begin{subfigure}[b]{\egswidth}
    \mytbox{
        \begin{overpic}[width=\egssubwidth,height=\egssubheight]{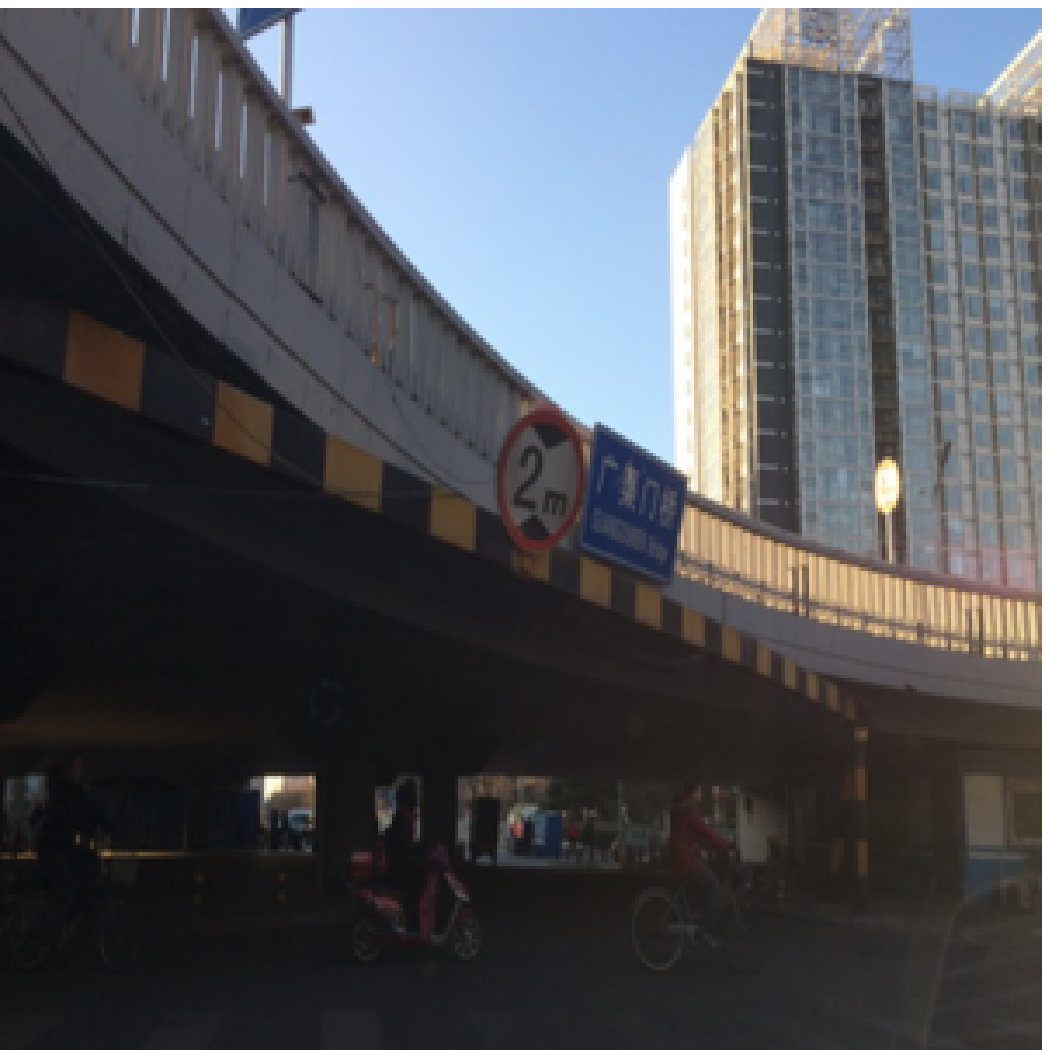}
         \put (20,90) {\red\textbf\small GT: Clear}
         \put (20,80) {\red\textbf\small Pred: Clear}
         \end{overpic}
    }
    \end{subfigure}
~    \begin{subfigure}[b]{\egswidth}
    \mytbox{
        \begin{overpic}[width=\egssubwidth,height=\egssubheight]{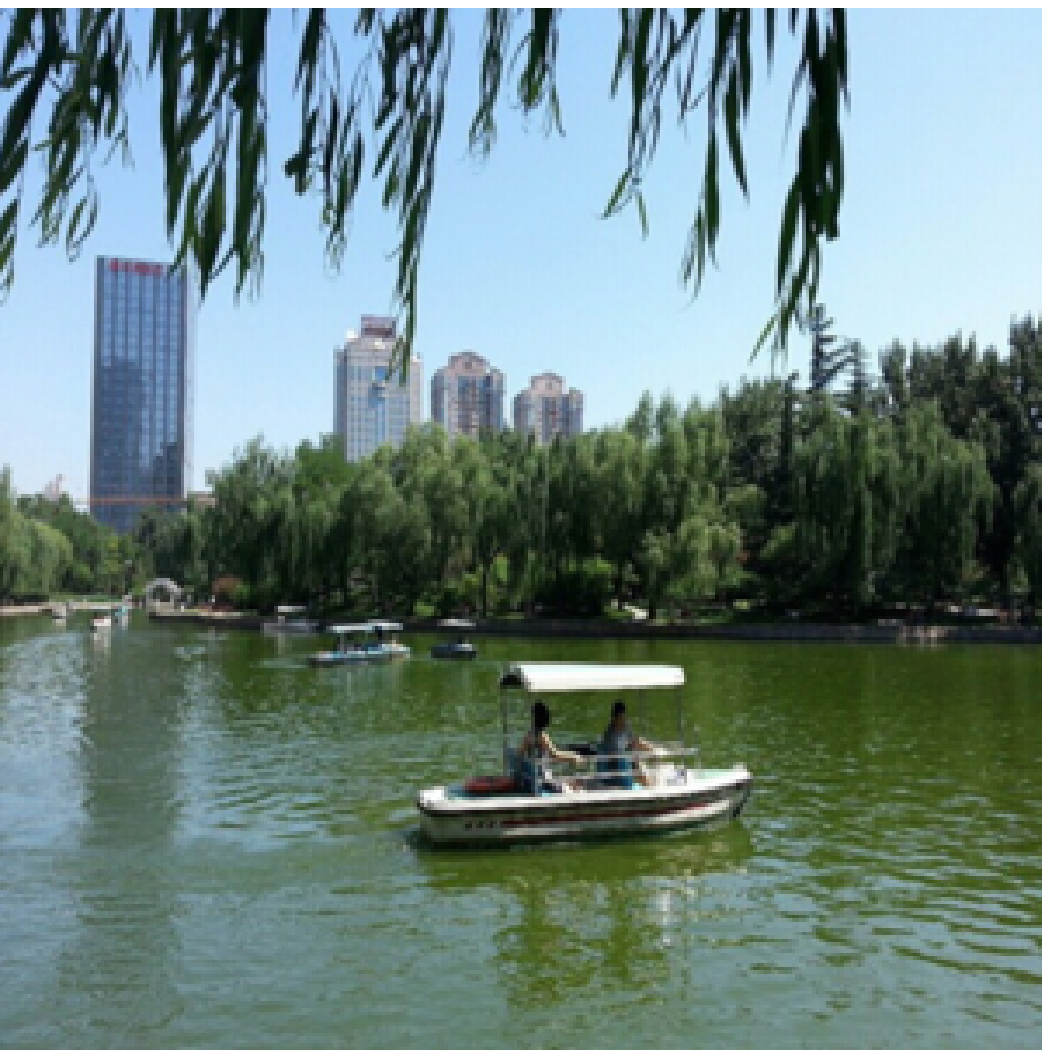}
         \put (20,90) {\red\textbf\small GT: Clear}
         \put (20,80) {\red\textbf\small Pred: Clear}
         \end{overpic}
    }
    \end{subfigure}
~    \begin{subfigure}[b]{\egswidth}
    \mytbox{
        \begin{overpic}[width=\egssubwidth,height=\egssubheight]{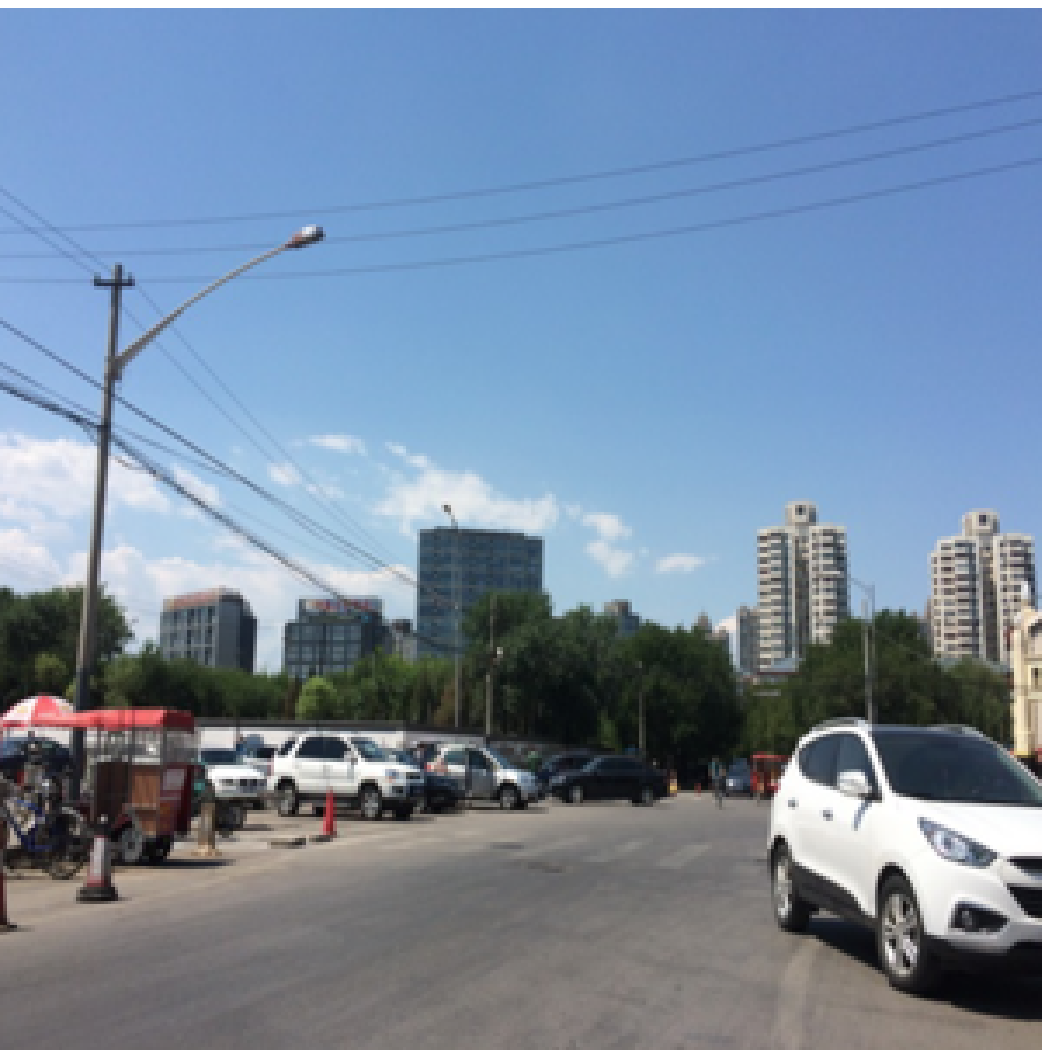}
         \put (20,90) {\red\textbf\small GT: Clear}
         \put (20,80) {\red\textbf\small Pred: Clear}
         \end{overpic}
    }
    \end{subfigure}
~    \begin{subfigure}[b]{\egswidth}
    \mytbox{
        \begin{overpic}[width=\egssubwidth,height=\egssubheight]{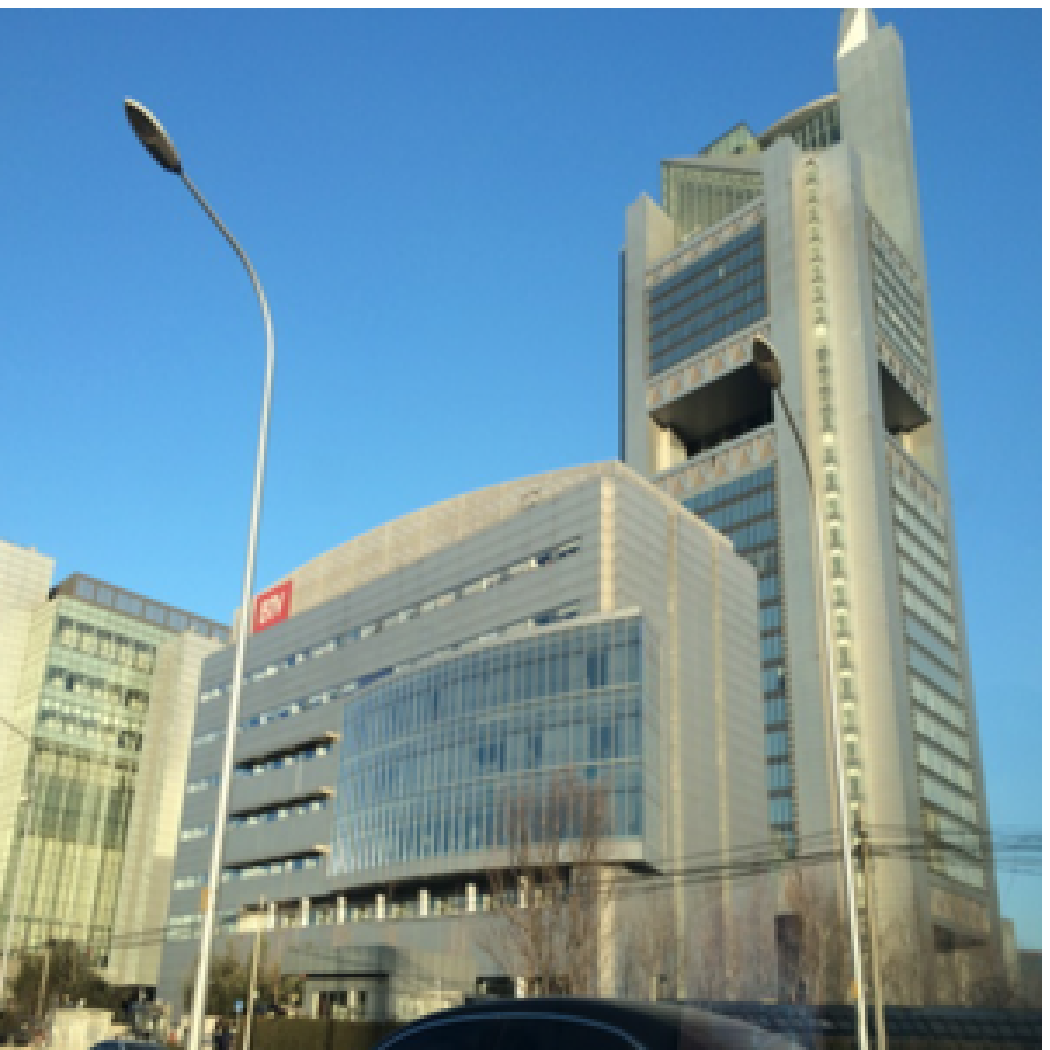}
         \put (20,90) {\red\textbf\small GT: Clear}
         \put (20,80) {\red\textbf\small Pred: Clear}
         \end{overpic}
    }
    \end{subfigure}
~    \begin{subfigure}[b]{\egswidth}
    \mytbox{
        \begin{overpic}[width=\egssubwidth,height=\egssubheight]{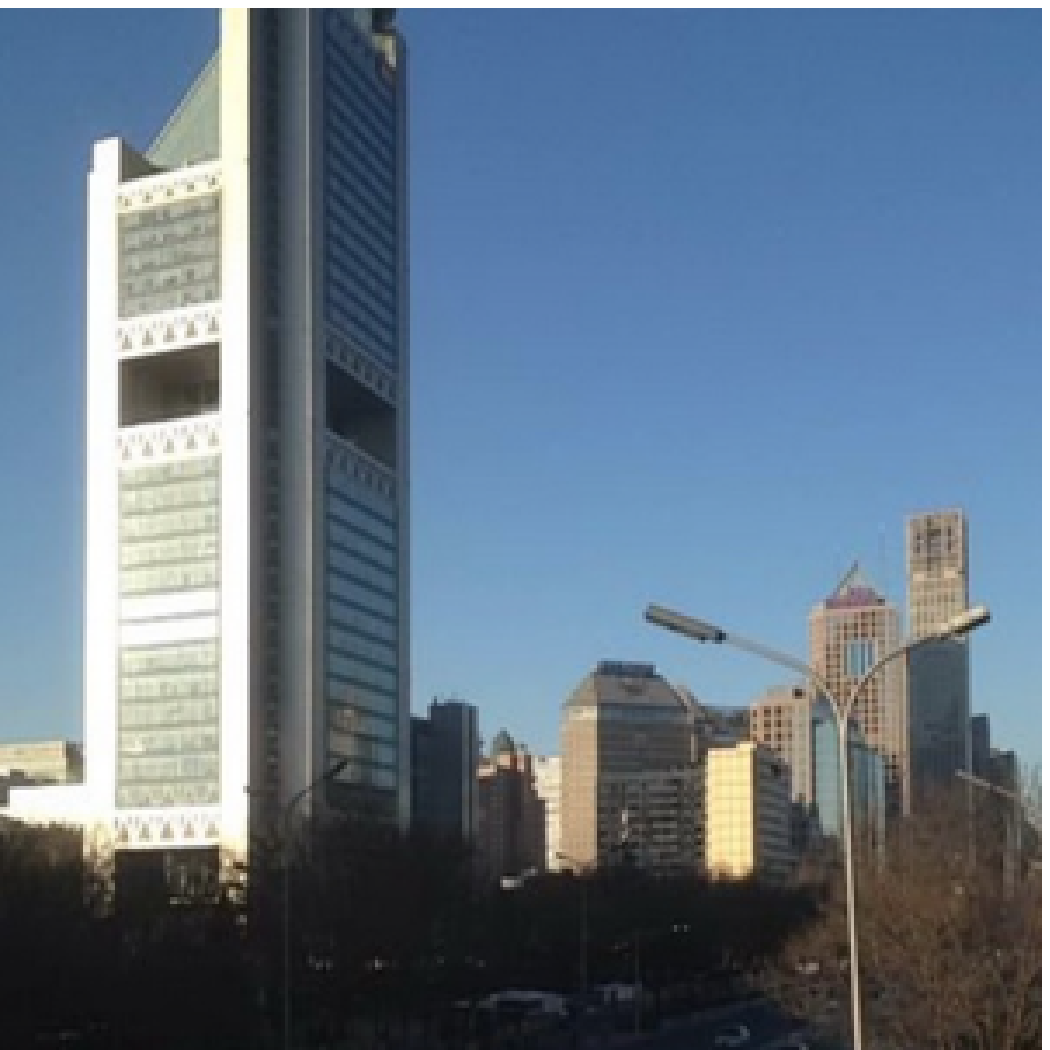}
         \put (20,90) {\red\textbf\small GT: Clear}
         \put (20,80) {\red\textbf\small Pred: Clear}
         \end{overpic}
    }
    \end{subfigure}
~    \begin{subfigure}[b]{\egswidth}
    \myfbox{
        \begin{overpic}[width=\egssubwidth,height=\egssubheight]{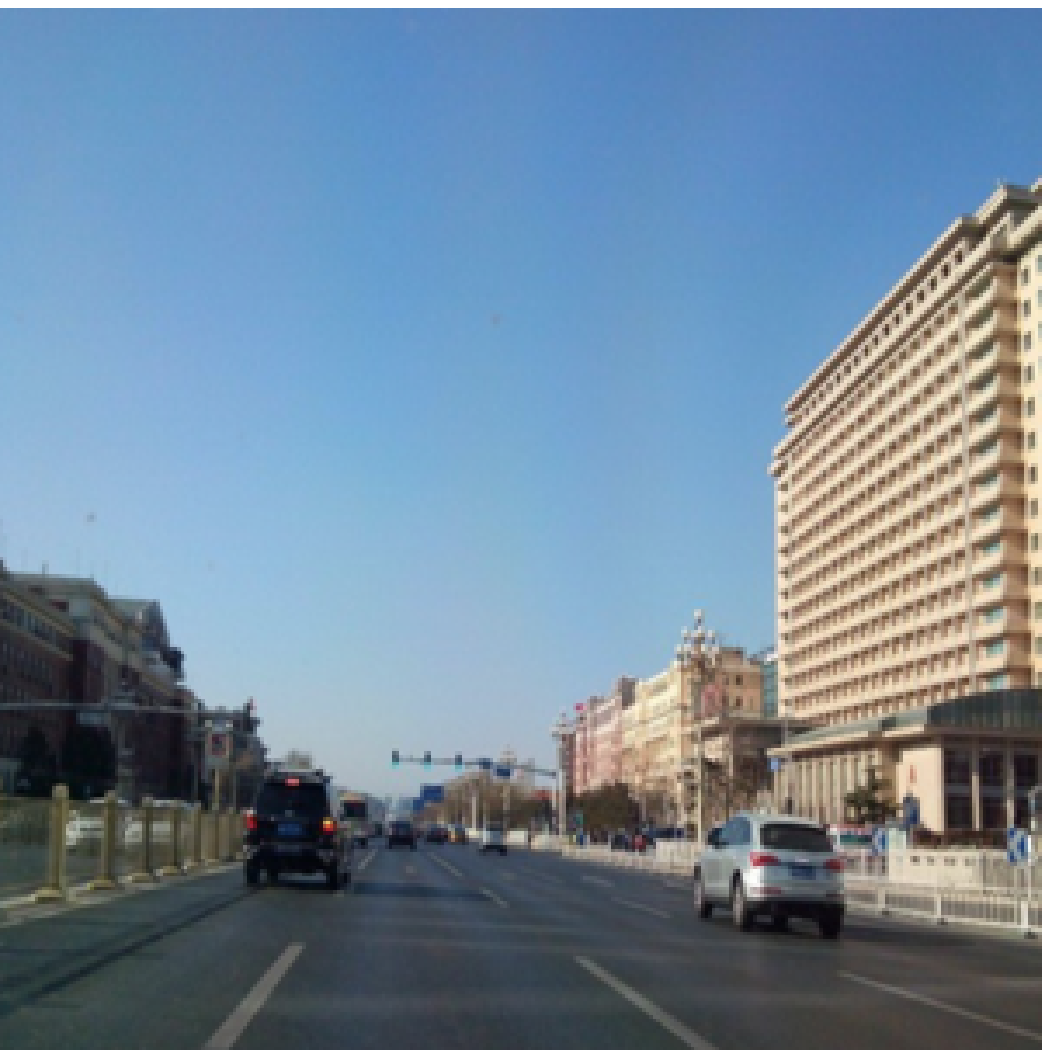}
         \put (20,90) {\red\textbf\small GT: Clear}
         \put (20,80) {\red\textbf\small Pred: Light}
         \end{overpic}
    }
    \end{subfigure}
~    \begin{subfigure}[b]{\egswidth}
    \myfbox{
        \begin{overpic}[width=\egssubwidth,height=\egssubheight]{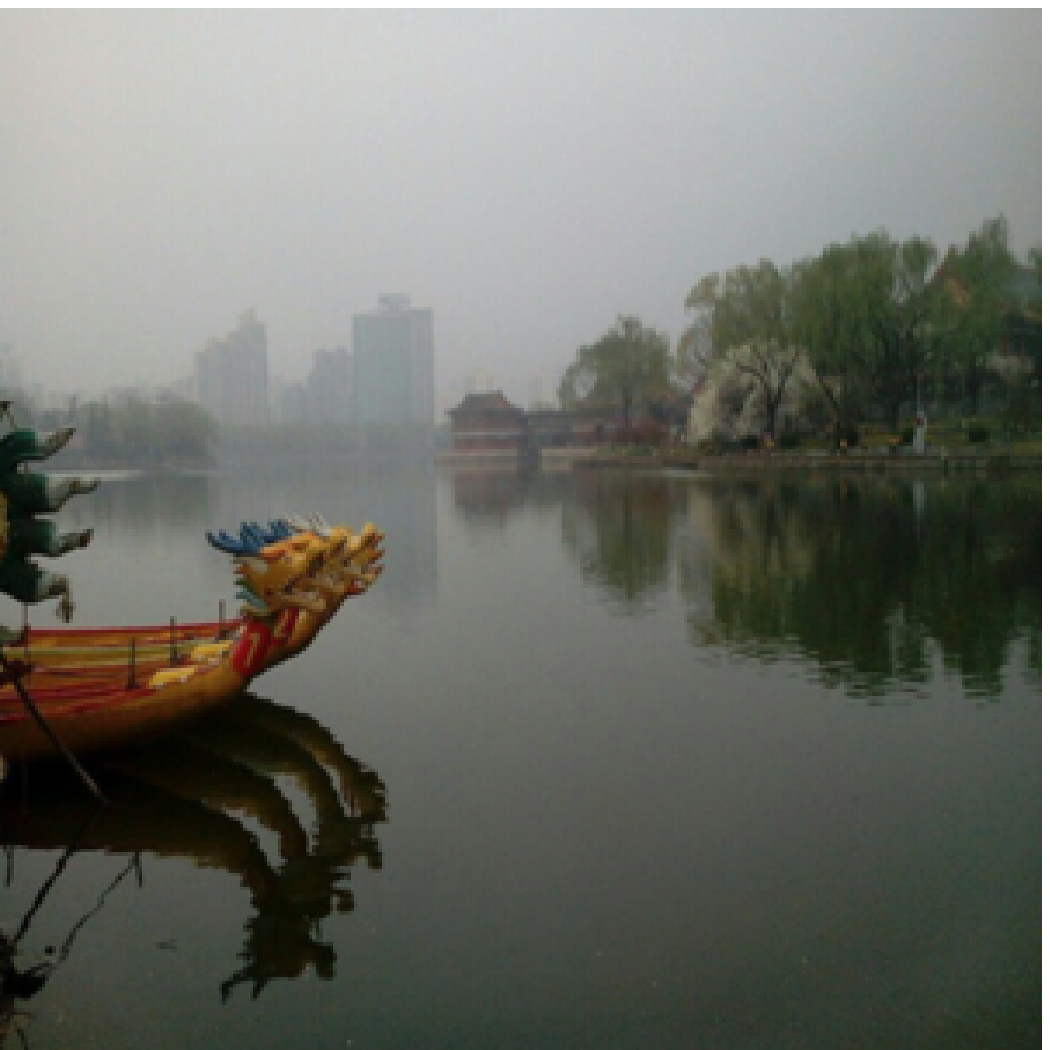}
         \put (20,90) {\red\textbf\small GT: Light}
         \put (20,80) {\red\textbf\small Pred: Clear}
         \end{overpic}
    }
    \end{subfigure}
~    \begin{subfigure}[b]{\egswidth}
    \mytbox{
        \begin{overpic}[width=\egssubwidth,height=\egssubheight]{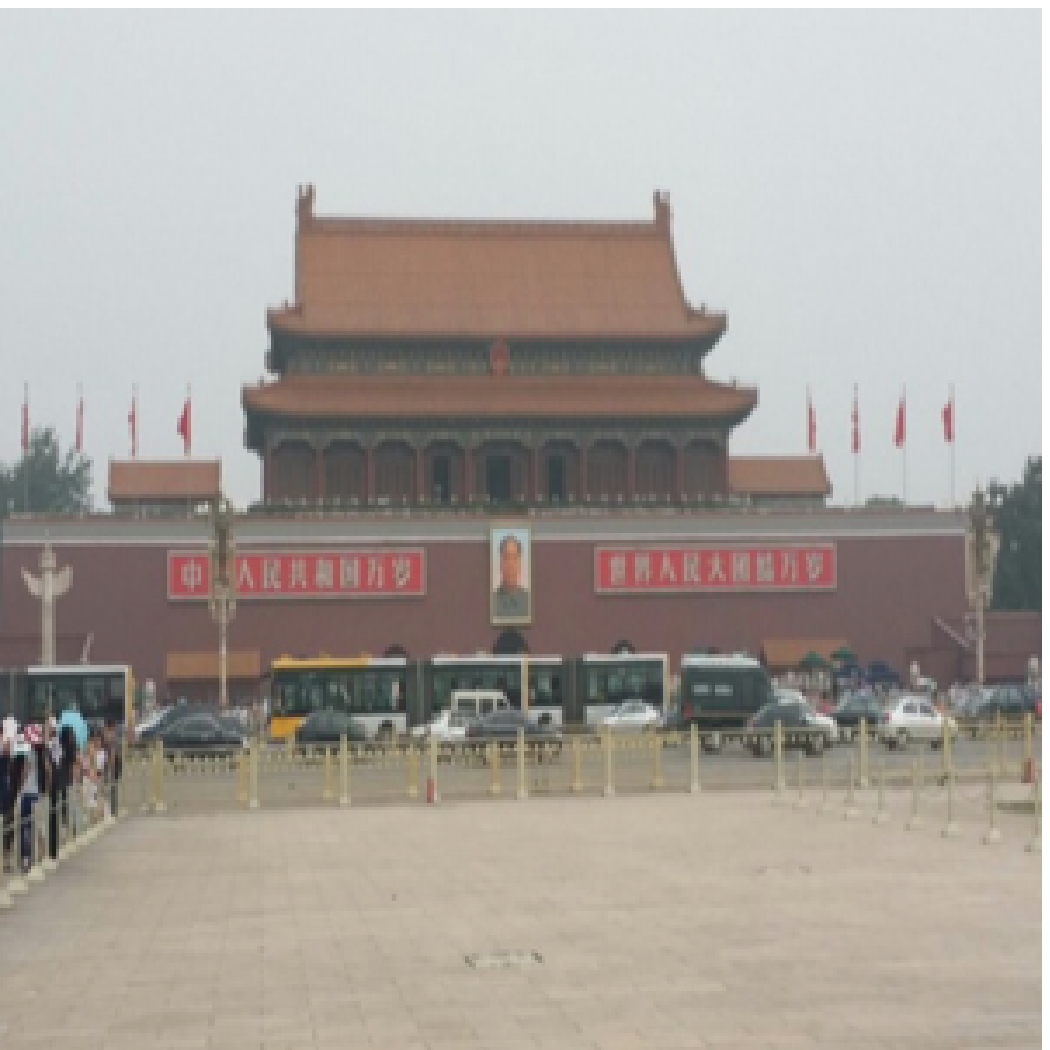}
         \put (20,90) {\red\textbf\small GT: Light}
         \put (20,80) {\red\textbf\small Pred: Light}
         \end{overpic}
    }
    \end{subfigure}
~    \begin{subfigure}[b]{\egswidth}
    \mytbox{
        \begin{overpic}[width=\egssubwidth,height=\egssubheight]{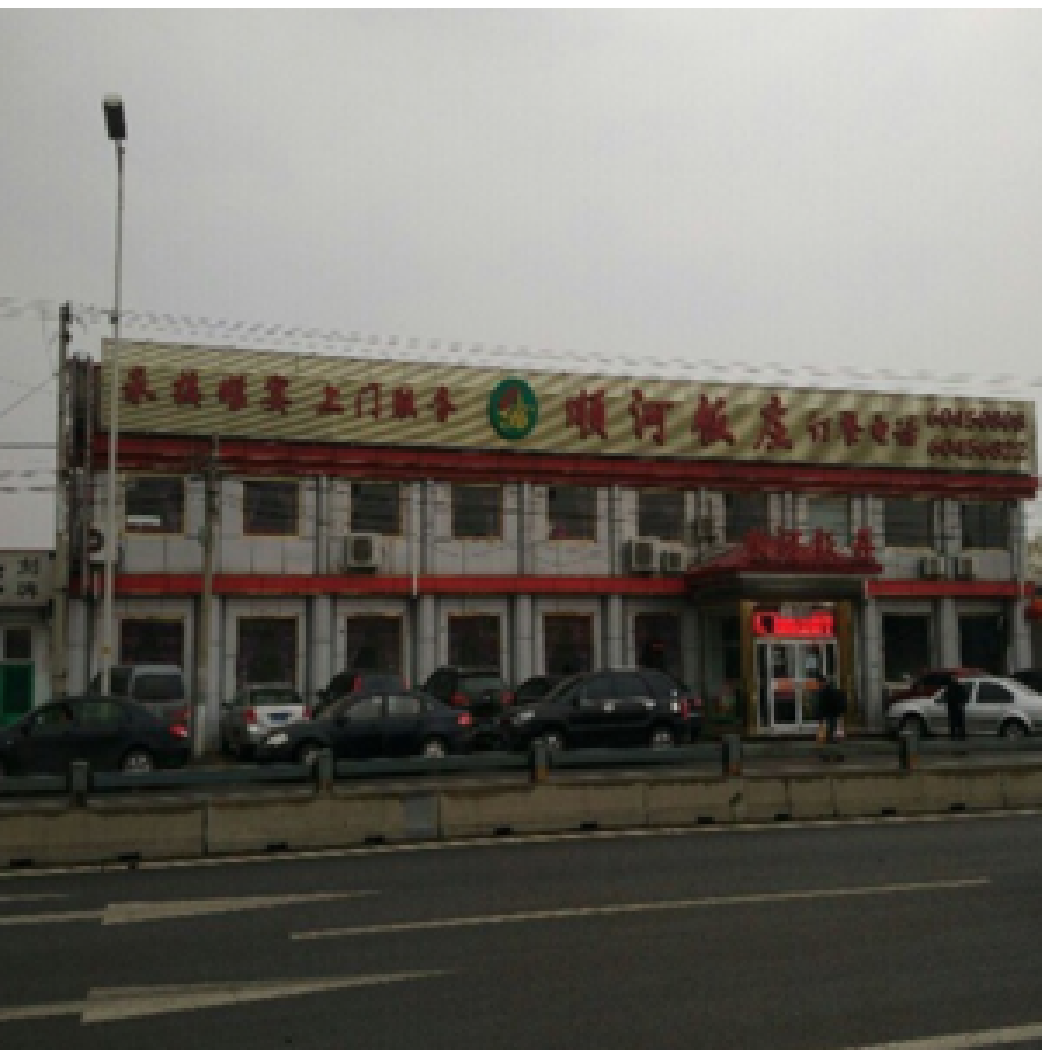}
         \put (20,90) {\red\textbf\small GT: Light}
         \put (20,80) {\red\textbf\small Pred: Light}
         \end{overpic}
    }
    \end{subfigure}
~    \begin{subfigure}[b]{\egswidth}
    \mytbox{
        \begin{overpic}[width=\egssubwidth,height=\egssubheight]{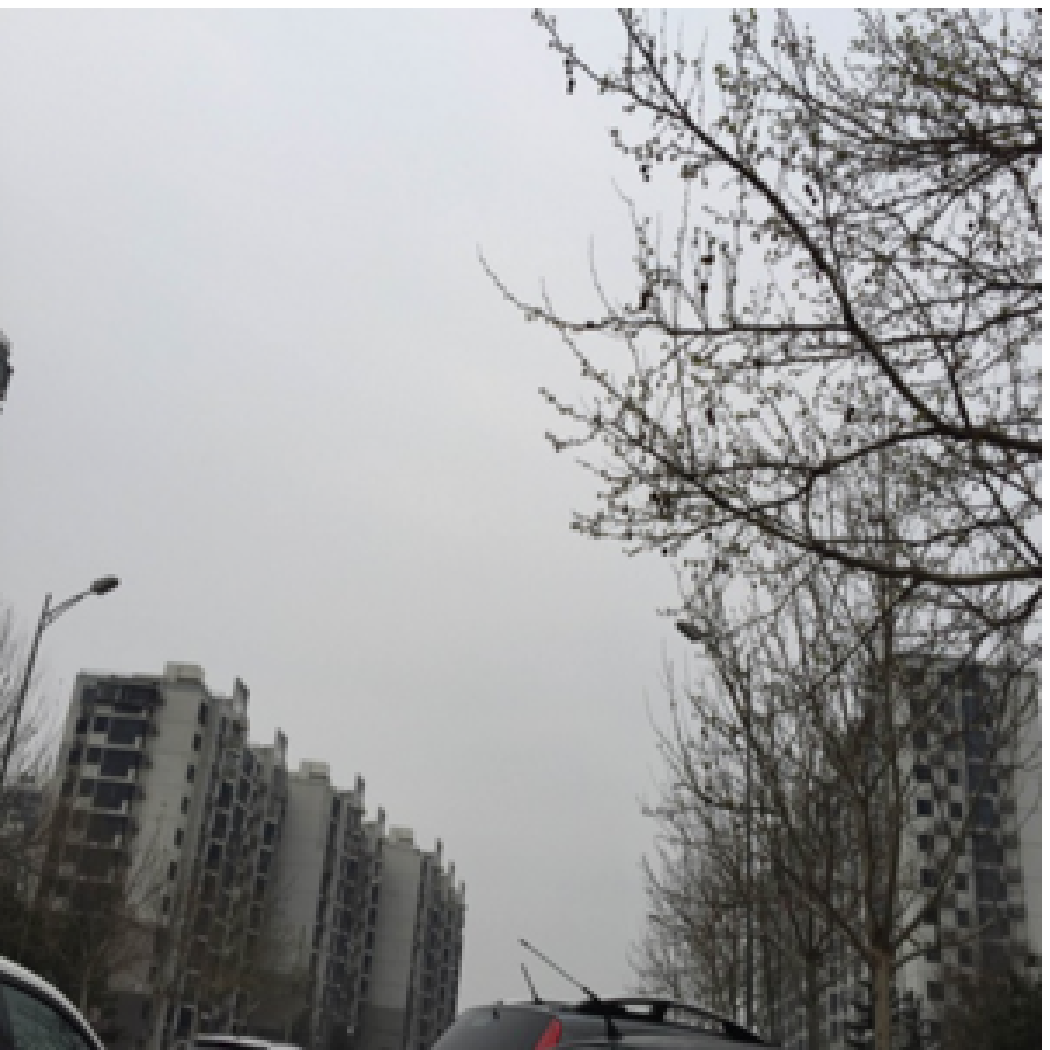}
         \put (20,90) {\red\textbf\small GT: Light}
         \put (20,80) {\red\textbf\small Pred: Light}
         \end{overpic}
    }
    \end{subfigure}
~    \begin{subfigure}[b]{\egswidth}
    \mytbox{
        \begin{overpic}[width=\egssubwidth,height=\egssubheight]{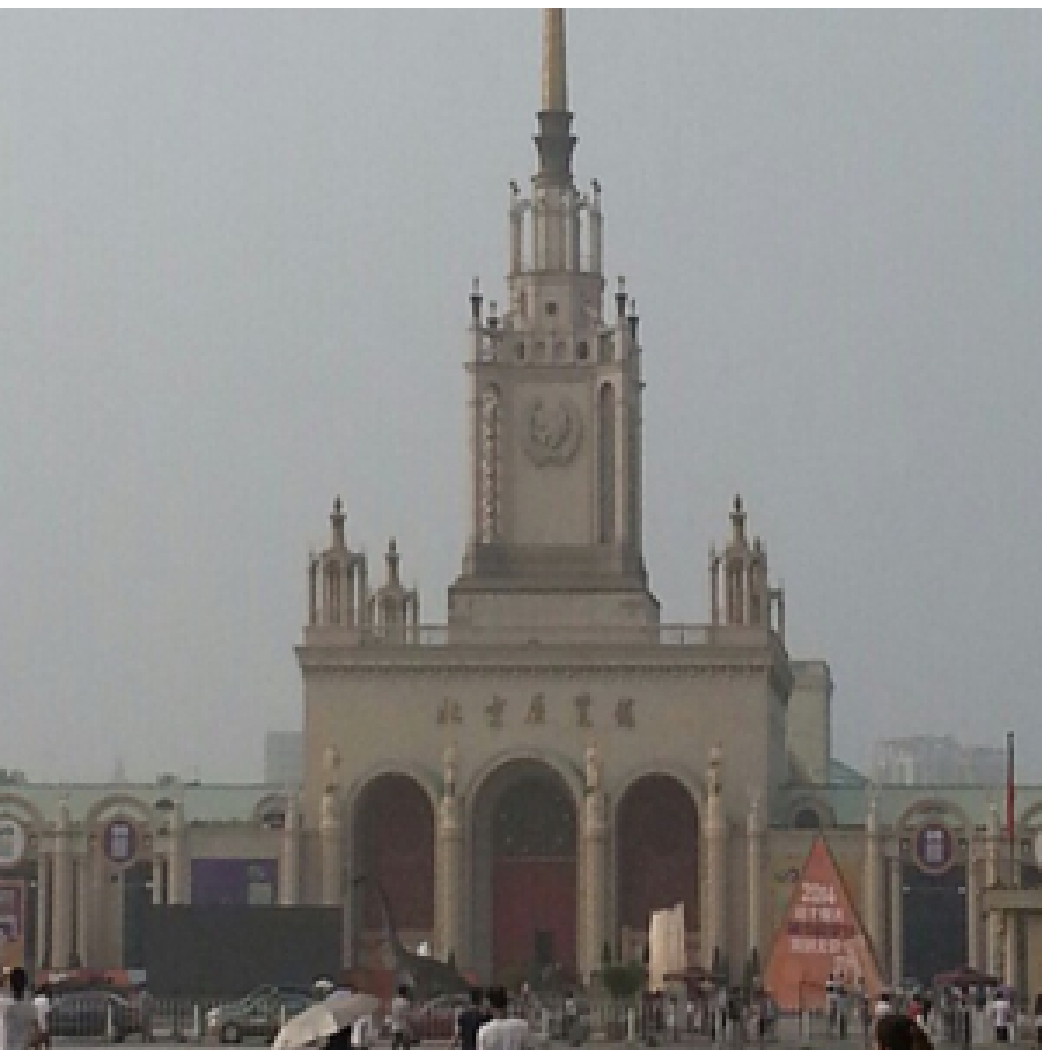}
         \put (20,90) {\red\textbf\small GT: Light}
         \put (20,80) {\red\textbf\small Pred: Light}
         \end{overpic}
    }
    \end{subfigure}
~    \begin{subfigure}[b]{\egswidth}
    \mytbox{
        \begin{overpic}[width=\egssubwidth,height=\egssubheight]{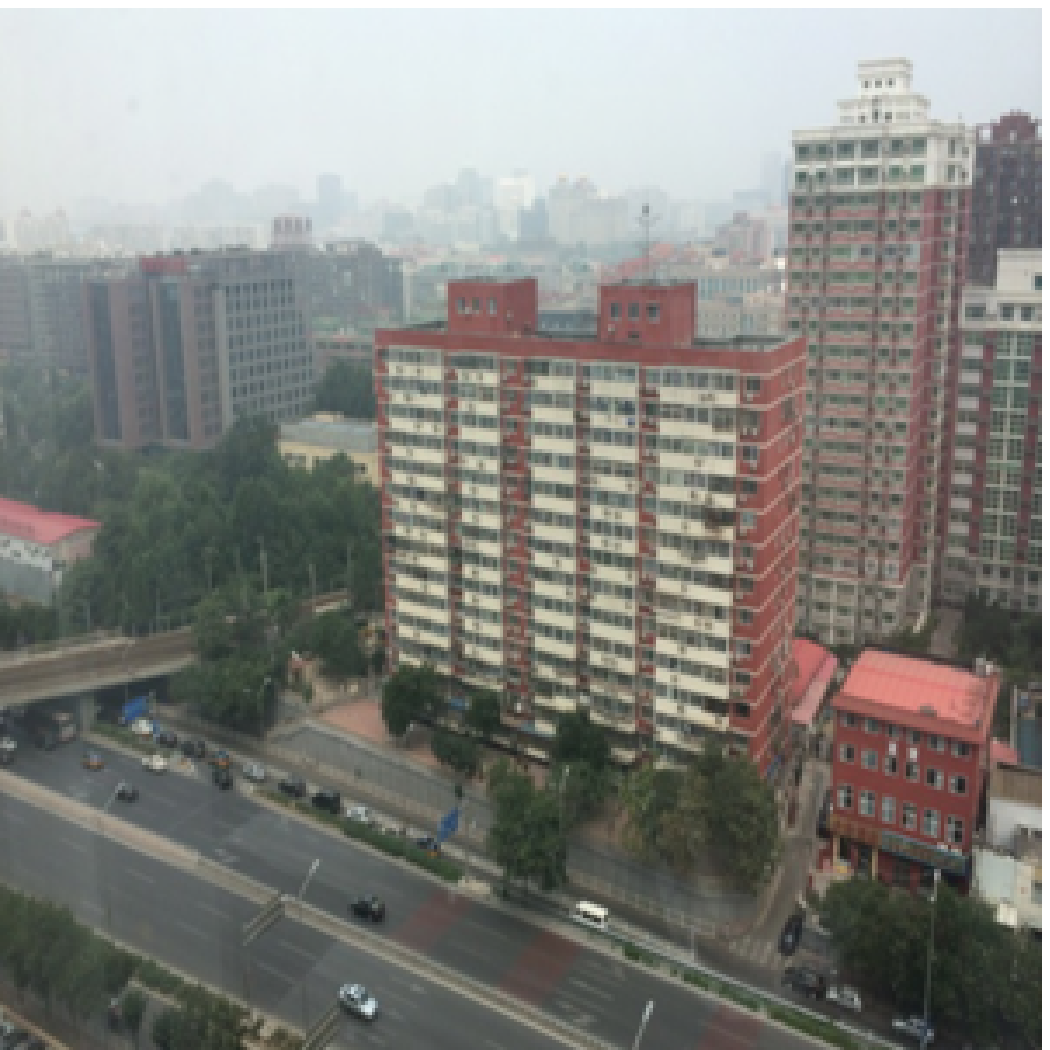}
         \put (20,90) {\red\textbf\small GT: Light}
         \put (20,80) {\red\textbf\small Pred: Light}
         \end{overpic}
    }
    \end{subfigure}
~    \begin{subfigure}[b]{\egswidth}
    \mytbox{
        \begin{overpic}[width=\egssubwidth,height=\egssubheight]{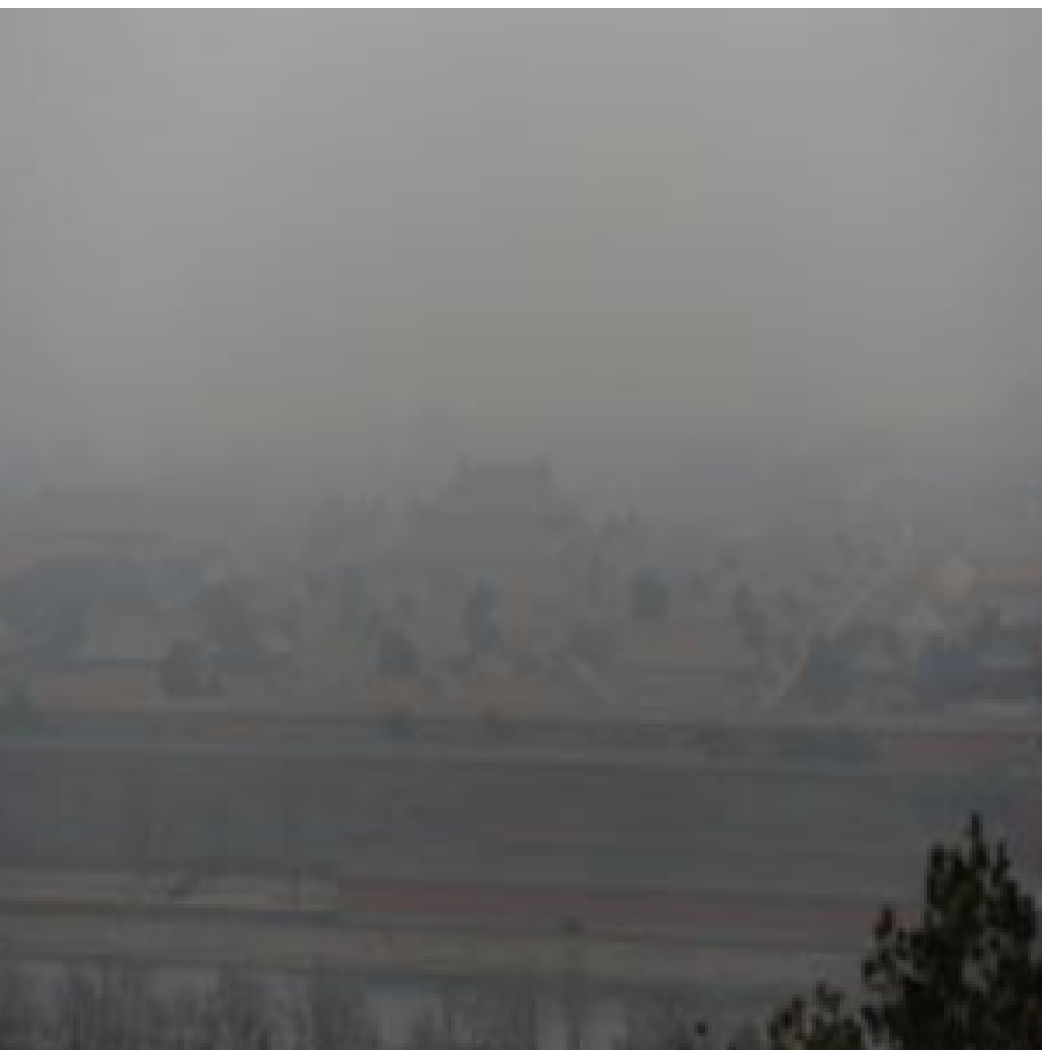}
         \put (20,90) {\red\textbf\small GT: Heavy}
         \put (20,80) {\red\textbf\small Pred: Heavy}
         \end{overpic}
    }
    \end{subfigure}
~    \begin{subfigure}[b]{\egswidth}
    \mytbox{
        \begin{overpic}[width=\egssubwidth,height=\egssubheight]{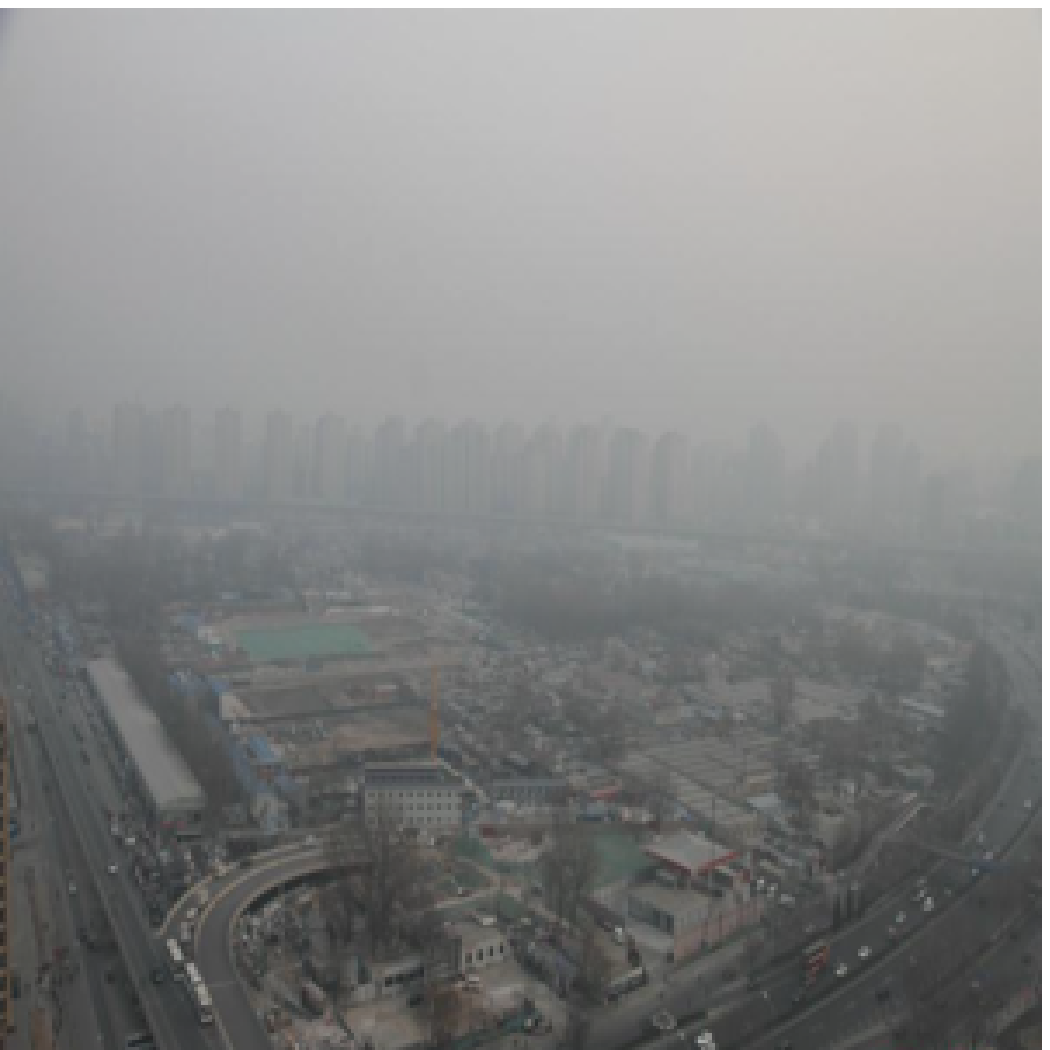}
         \put (20,90) {\red\textbf\small GT: Heavy}
         \put (20,80) {\red\textbf\small Pred: Heavy}
         \end{overpic}
    }
    \end{subfigure}
~    \begin{subfigure}[b]{\egswidth}
    \mytbox{
        \begin{overpic}[width=\egssubwidth,height=\egssubheight]{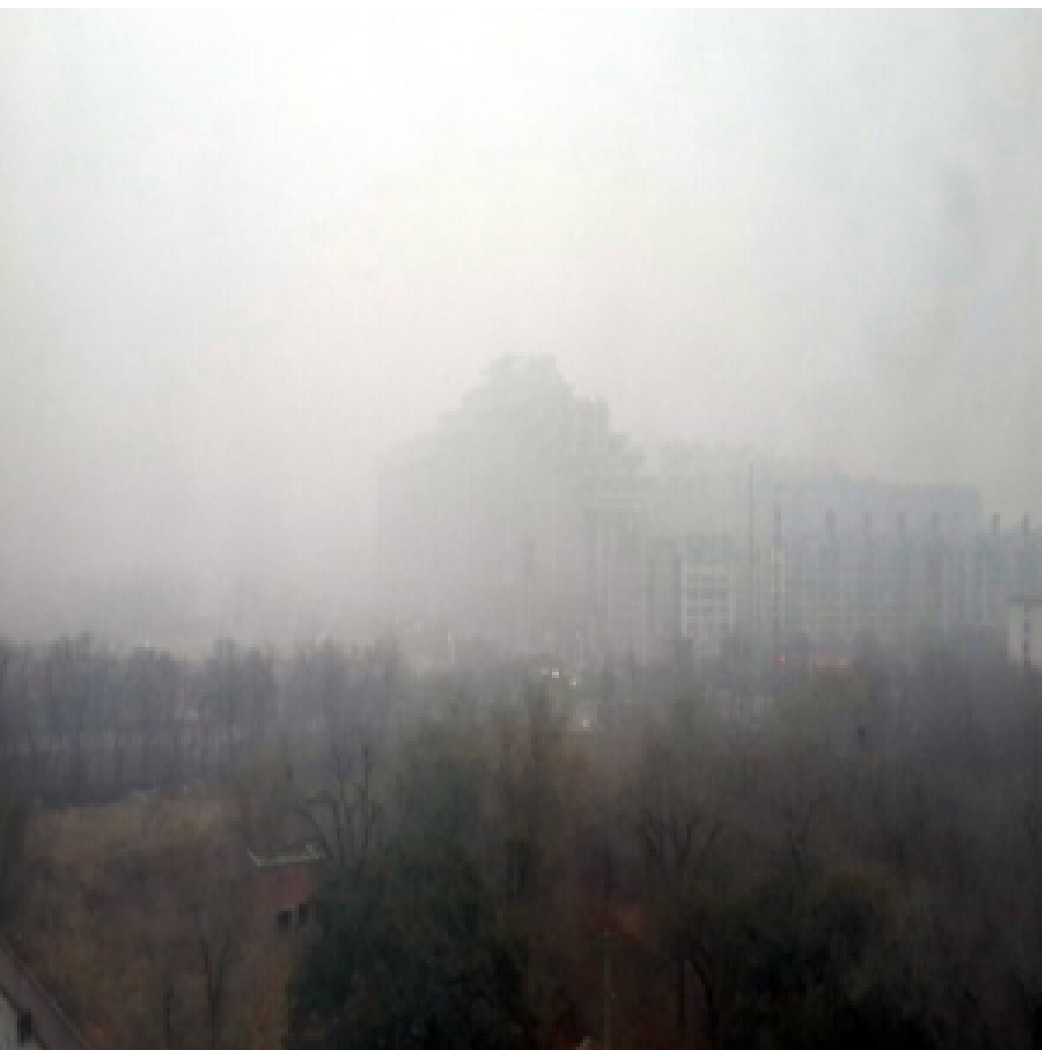}
         \put (20,90) {\red\textbf\small GT: Heavy}
         \put (20,80) {\red\textbf\small Pred: Heavy}
         \end{overpic}
    }
    \end{subfigure}
~    \begin{subfigure}[b]{\egswidth}
    \mytbox{
        \begin{overpic}[width=\egssubwidth,height=\egssubheight]{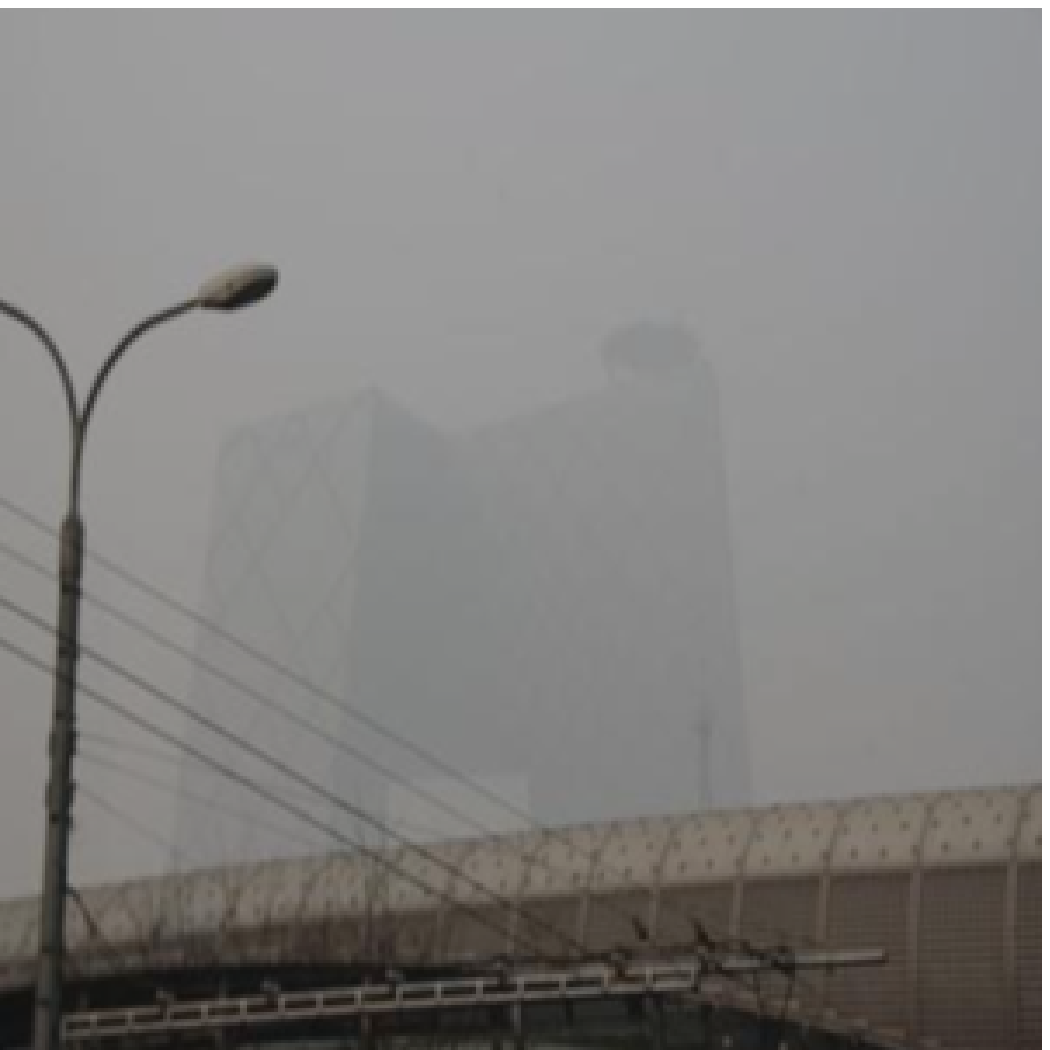}
         \put (20,90) {\red\textbf\small GT: Heavy}
         \put (20,80) {\red\textbf\small Pred: Heavy}
         \end{overpic}
    }
    \end{subfigure}
~    \begin{subfigure}[b]{\egswidth}
    \mytbox{
        \begin{overpic}[width=\egssubwidth,height=\egssubheight]{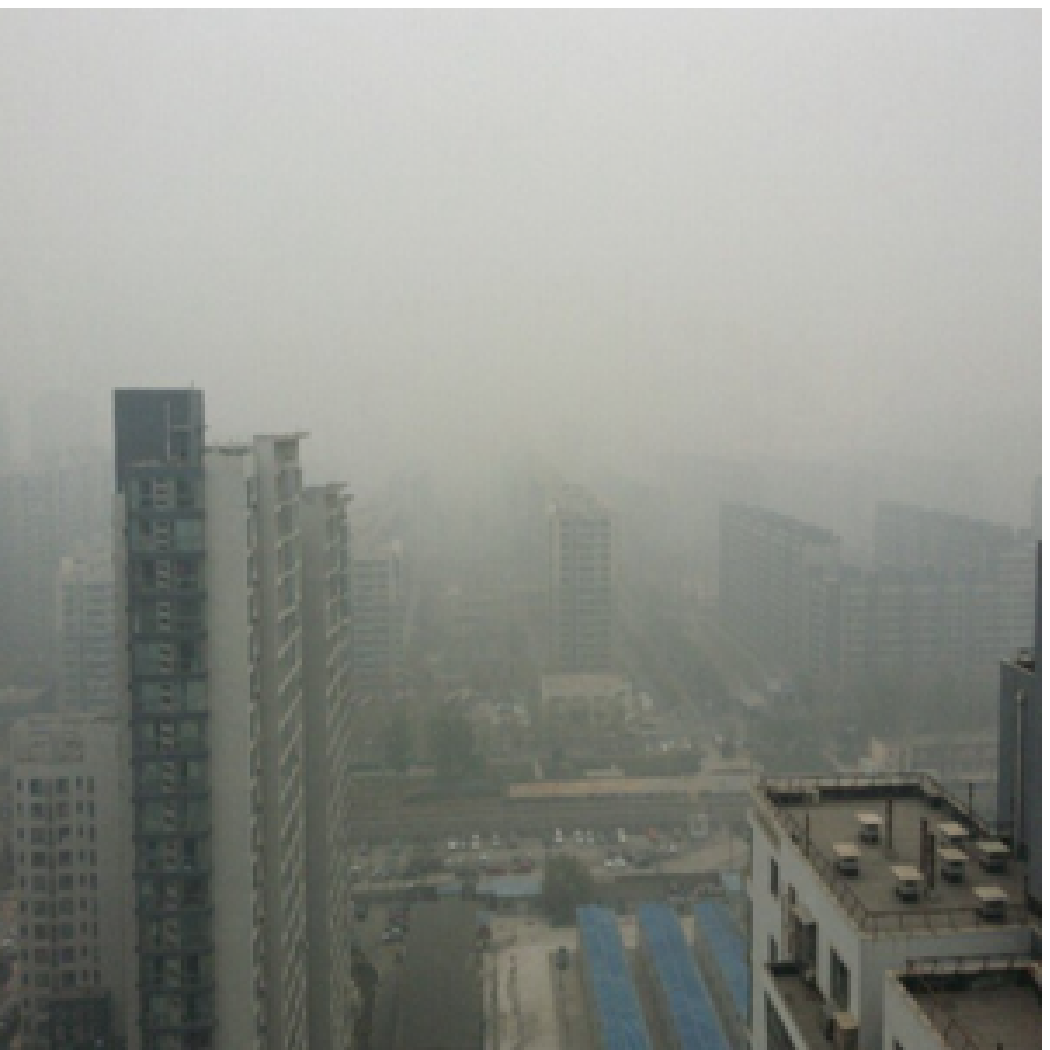}
         \put (20,90) {\red\textbf\small GT: Heavy}
         \put (20,80) {\red\textbf\small Pred: Heavy}
         \end{overpic}
    }
    \end{subfigure}
~    \begin{subfigure}[b]{\egswidth}
    \mytbox{
        \begin{overpic}[width=\egssubwidth,height=\egssubheight]{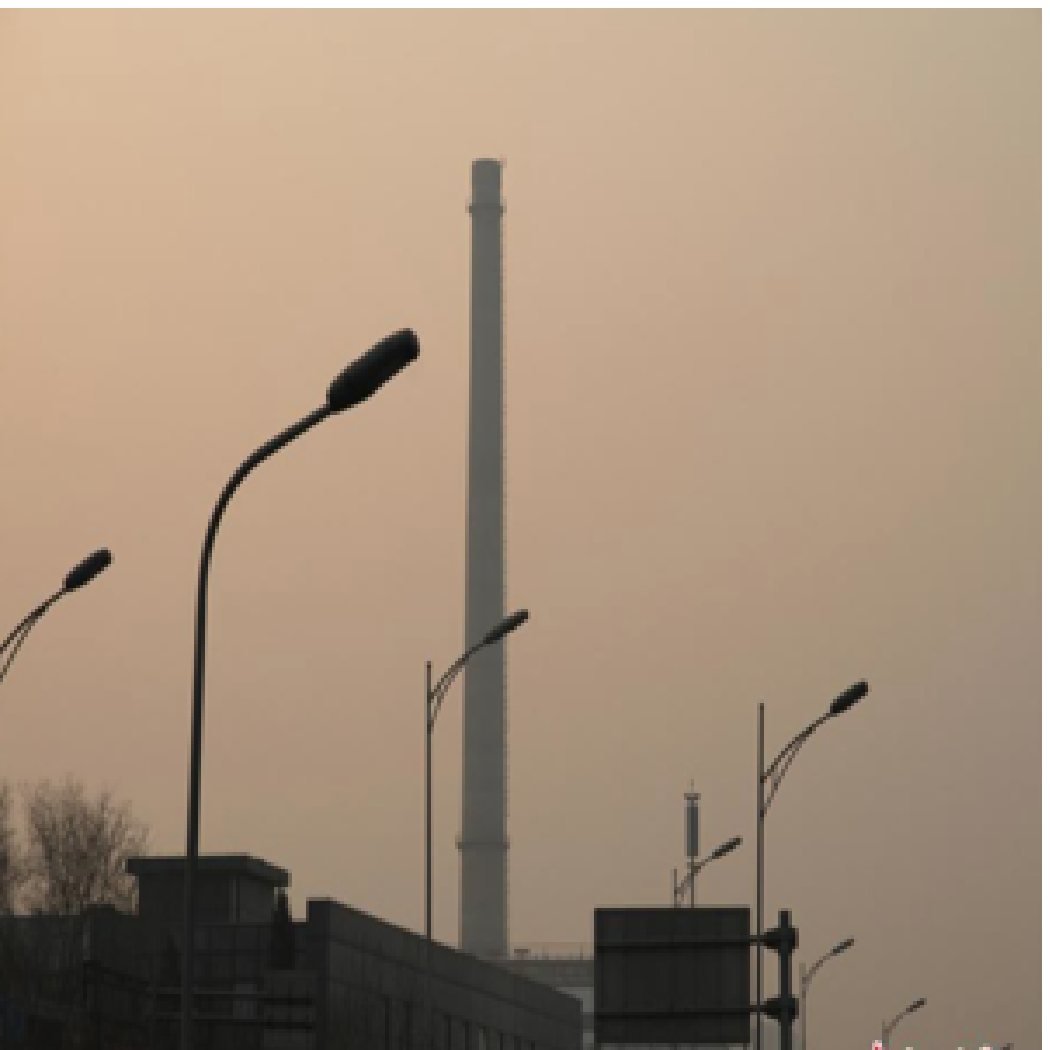}
         \put (20,90) {\red\textbf\small GT: Heavy}
         \put (20,80) {\red\textbf\small Pred: Heavy}
         \end{overpic}
    }
    \end{subfigure}
~    \begin{subfigure}[b]{\egswidth}
    \mytbox{
        \begin{overpic}[width=\egssubwidth,height=\egssubheight]{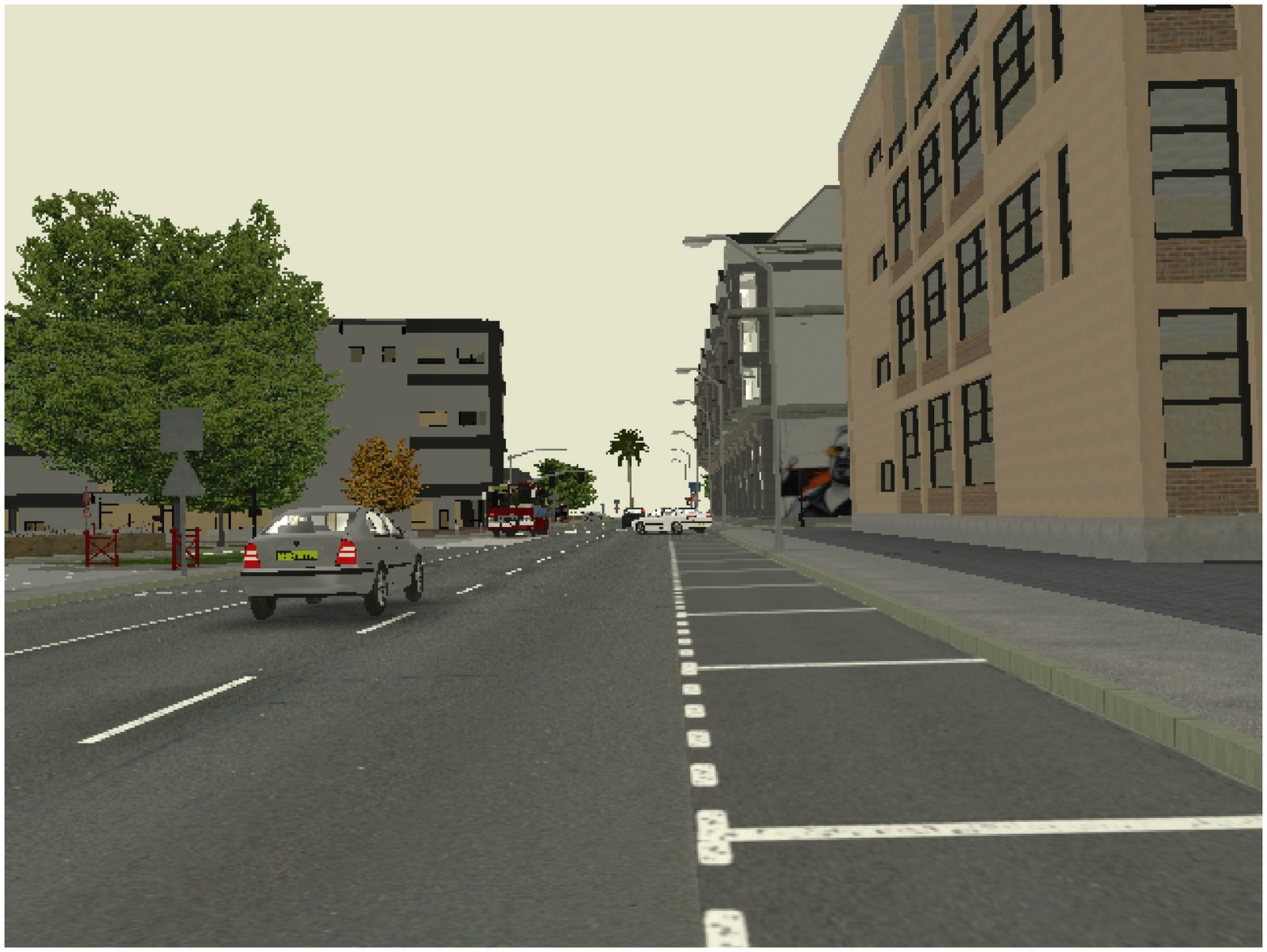}
         \put (20,90) {\red\textbf\small GT: Clear}
         \put (20,80) {\red\textbf\small Pred: Clear}
         \end{overpic}
    }
    \end{subfigure}
~    \begin{subfigure}[b]{\egswidth}
    \mytbox{
        \begin{overpic}[width=\egssubwidth,height=\egssubheight]{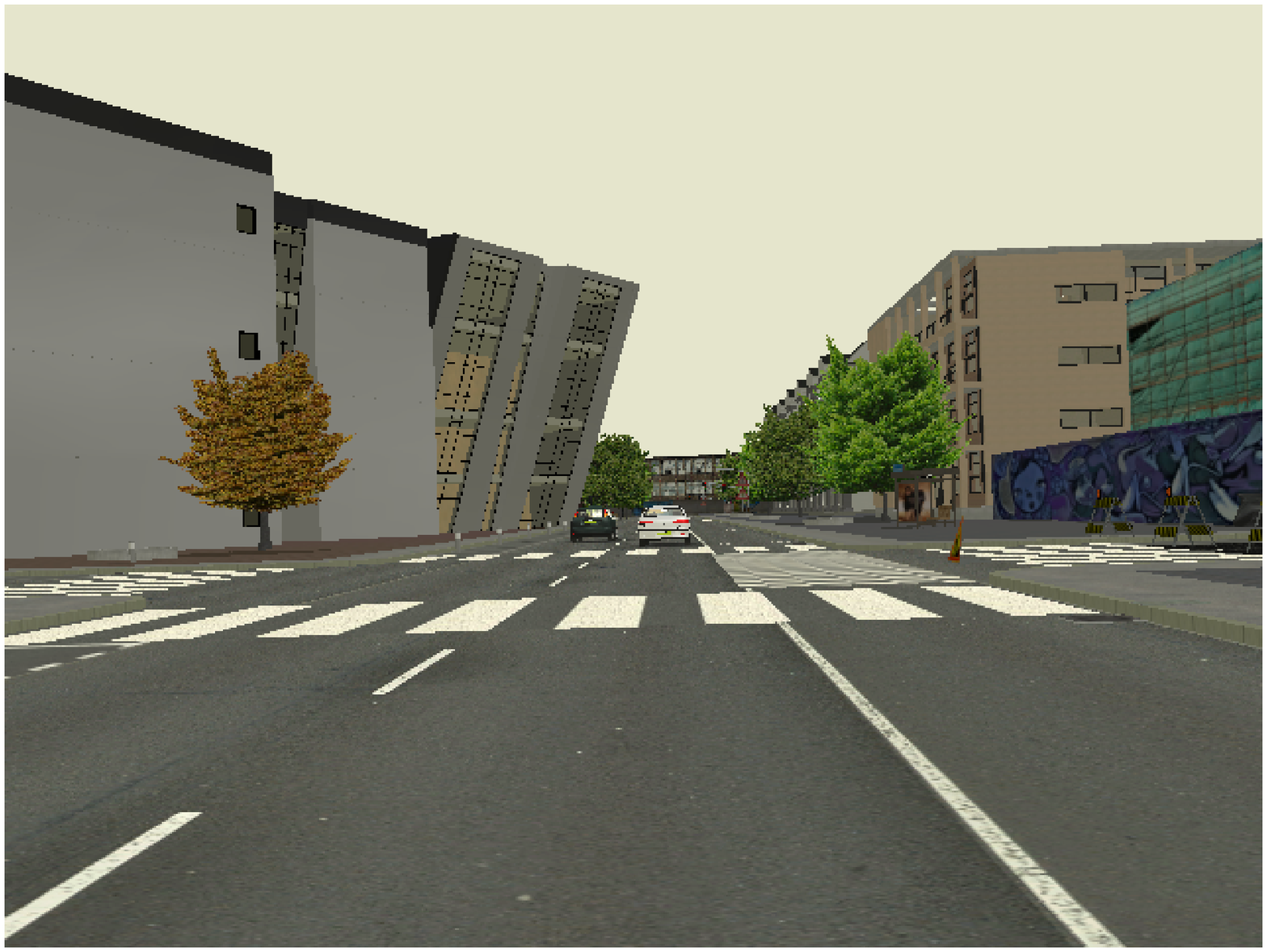}
         \put (20,90) {\red\textbf\small GT: Clear}
         \put (20,80) {\red\textbf\small Pred: Clear}
         \end{overpic}
    }
    \end{subfigure}
~    \begin{subfigure}[b]{\egswidth}
    \mytbox{
        \begin{overpic}[width=\egssubwidth,height=\egssubheight]{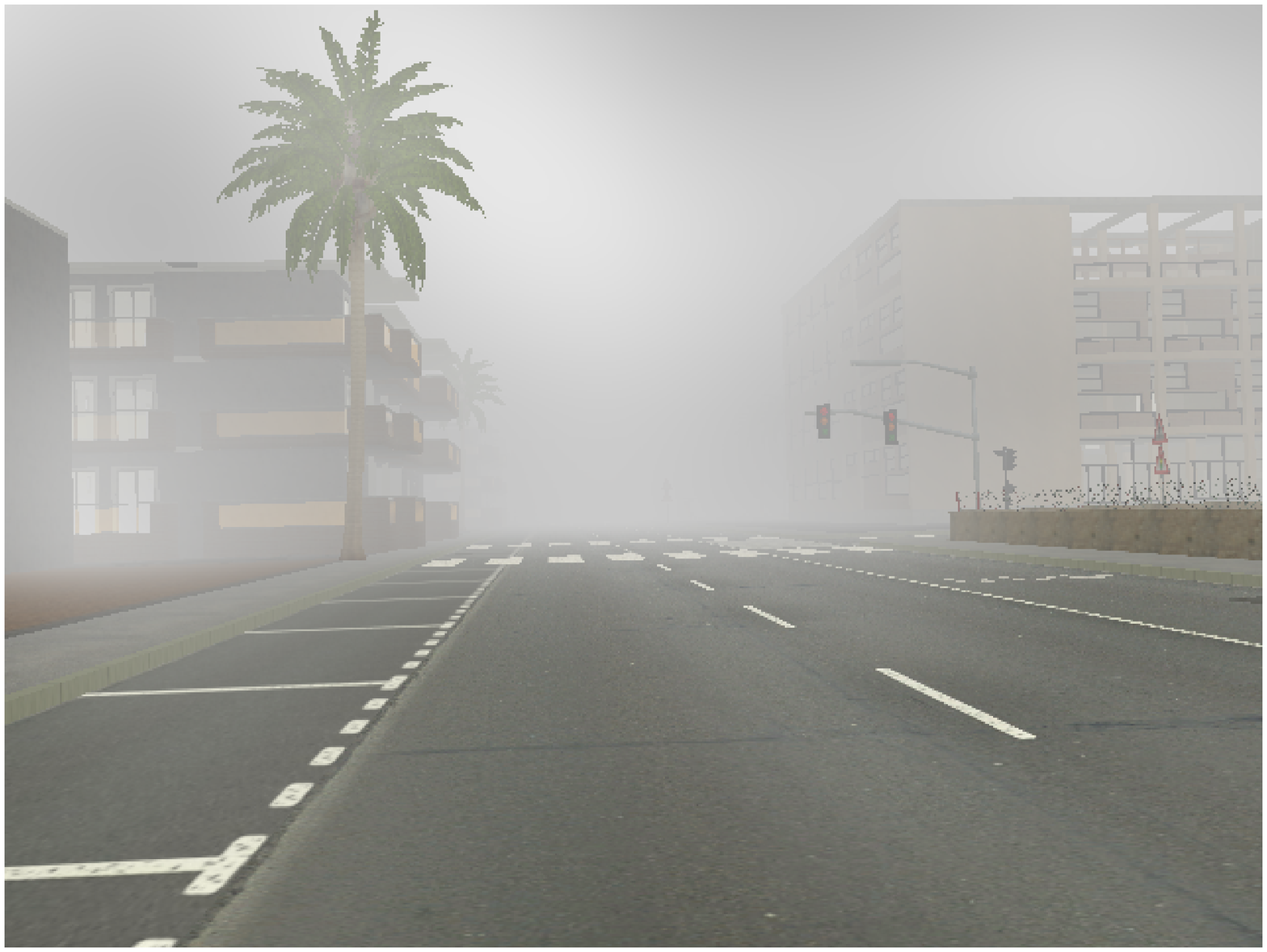}
         \put (20,90) {\red\textbf\small GT: Light}
         \put (20,80) {\red\textbf\small Pred: Light}
         \end{overpic}
    }
    \end{subfigure}
~    \begin{subfigure}[b]{\egswidth}
    \mytbox{
        \begin{overpic}[width=\egssubwidth,height=\egssubheight]{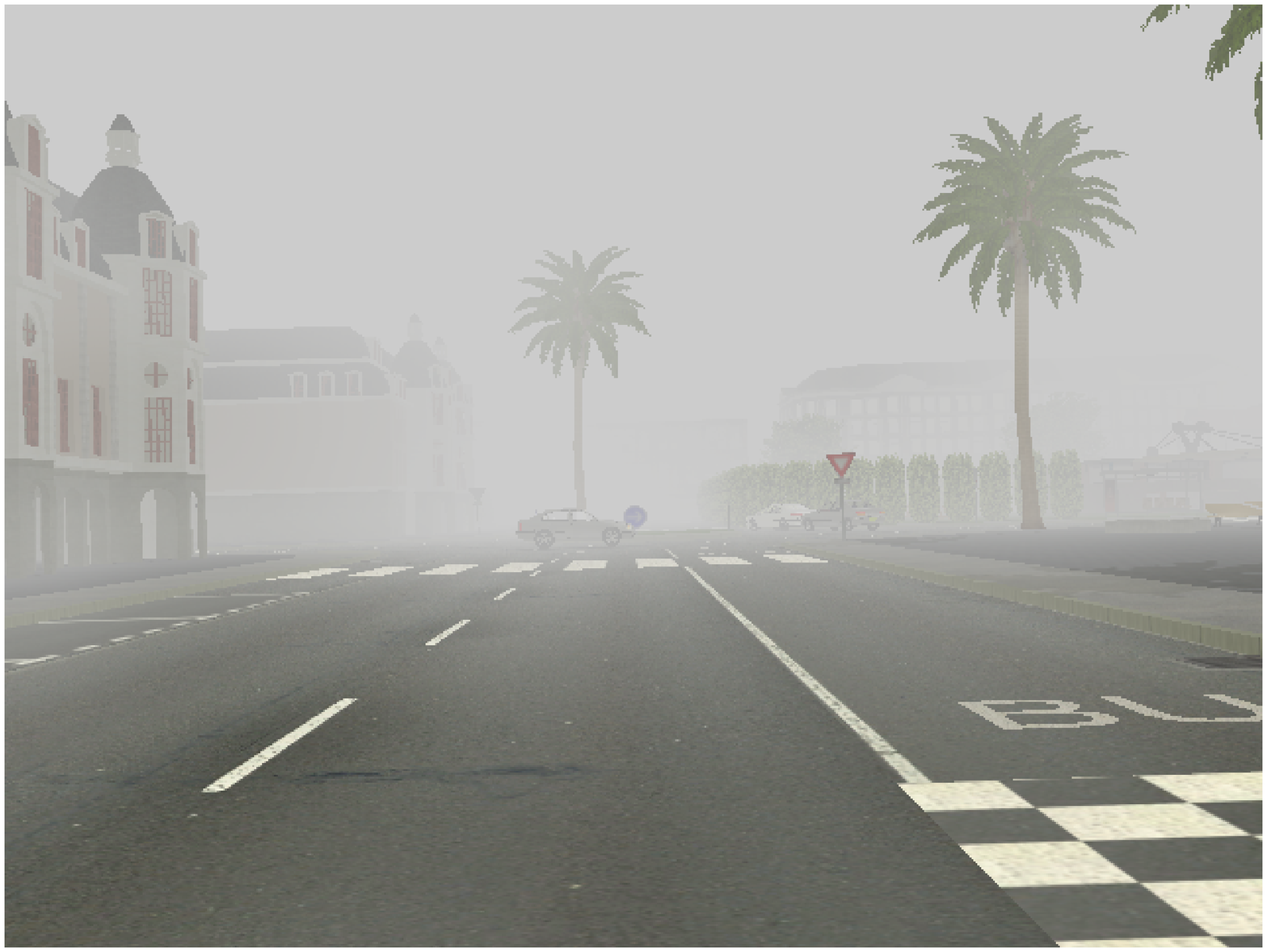}
         \put (20,90) {\red\textbf\small GT: Light}
         \put (20,80) {\red\textbf\small Pred: Light}
         \end{overpic}
    }
    \end{subfigure}
~    \begin{subfigure}[b]{\egswidth}
    \mytbox{
        \begin{overpic}[width=\egssubwidth,height=\egssubheight]{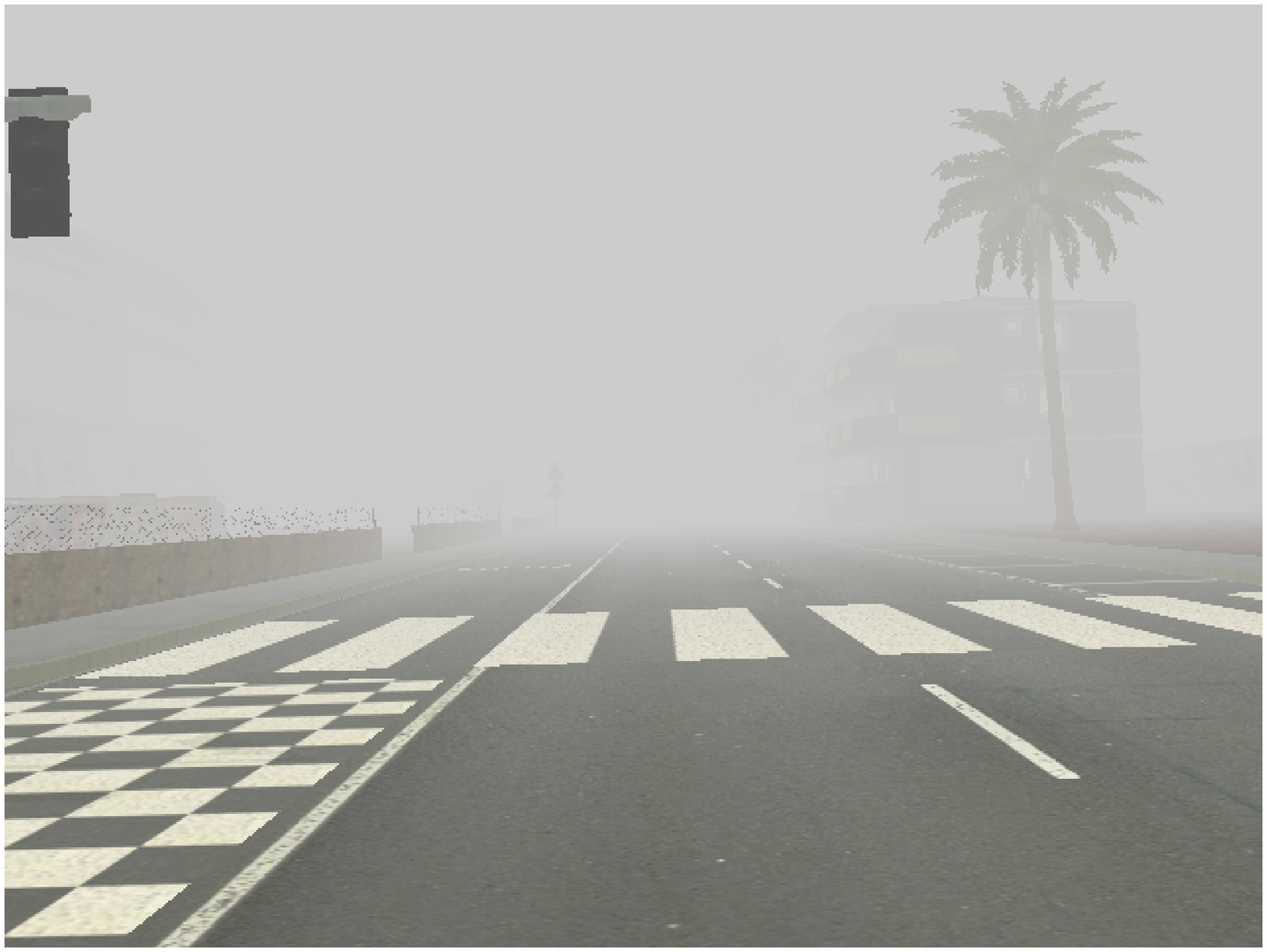}
         \put (20,90) {\red\textbf\small GT: Heavy}
         \put (20,80) {\red\textbf\small Pred: Heavy}
         \end{overpic}
    }
    \end{subfigure}
~    \begin{subfigure}[b]{\egswidth}
    \mytbox{
        \begin{overpic}[width=\egssubwidth,height=\egssubheight]{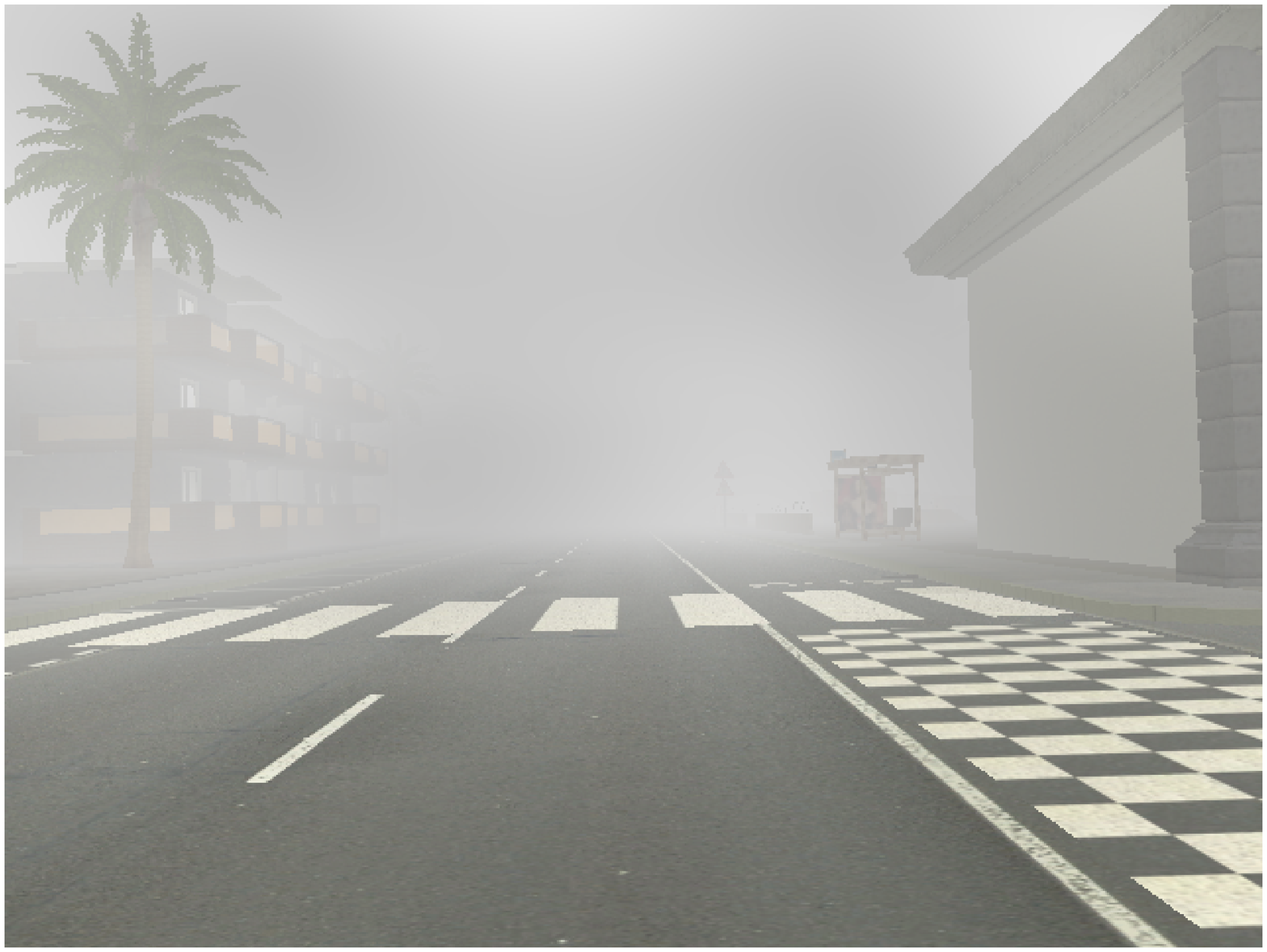}
         \put (20,90) {\red\textbf\small GT: Heavy}
         \put (20,80) {\red\textbf\small Pred: Heavy}
         \end{overpic}
    }
    \end{subfigure}
~    \begin{subfigure}[b]{\egswidth}
    \mytbox{
        \begin{overpic}[width=\egssubwidth,height=\egssubheight]{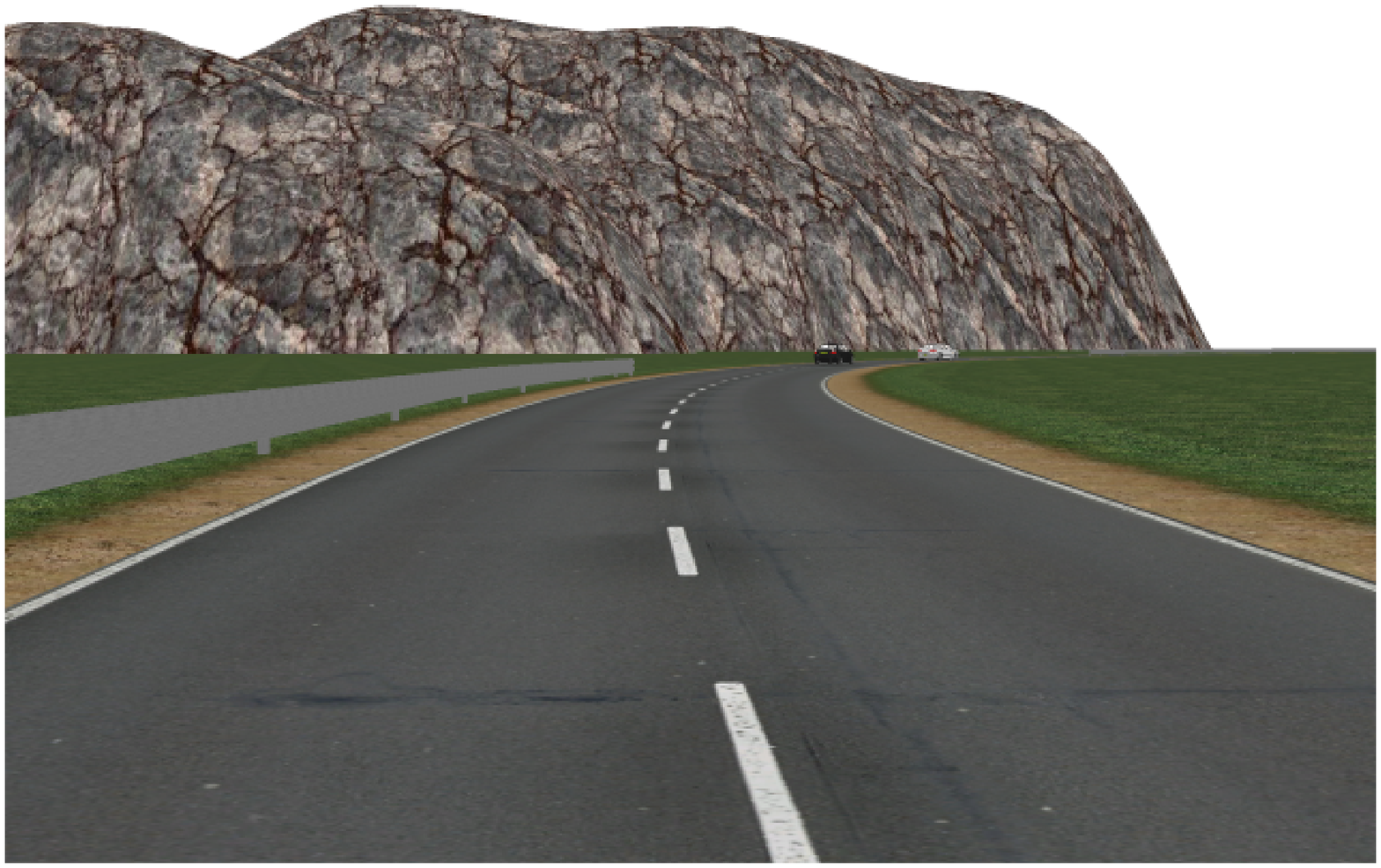}
         \put (20,90) {\red\textbf\small GT: Clear}
         \put (20,80) {\red\textbf\small Pred: Clear}
         \end{overpic}
    }
    \end{subfigure}
~    \begin{subfigure}[b]{\egswidth}
    \mytbox{
        \begin{overpic}[width=\egssubwidth,height=\egssubheight]{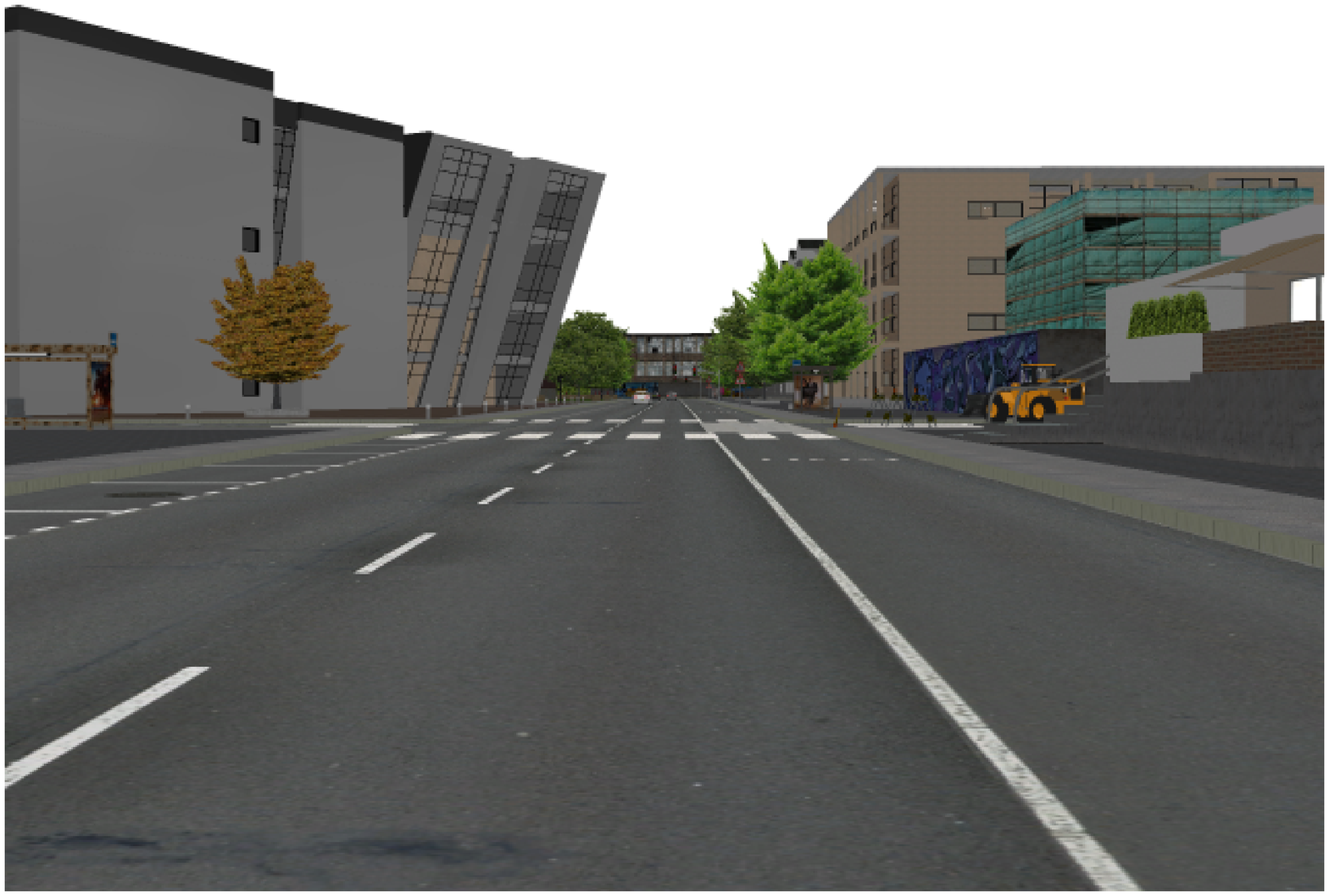}
         \put (20,90) {\red\textbf\small GT: Clear}
         \put (20,80) {\red\textbf\small Pred: Clear}
         \end{overpic}
    }
    \end{subfigure}
~    \begin{subfigure}[b]{\egswidth}
    \mytbox{
        \begin{overpic}[width=\egssubwidth,height=\egssubheight]{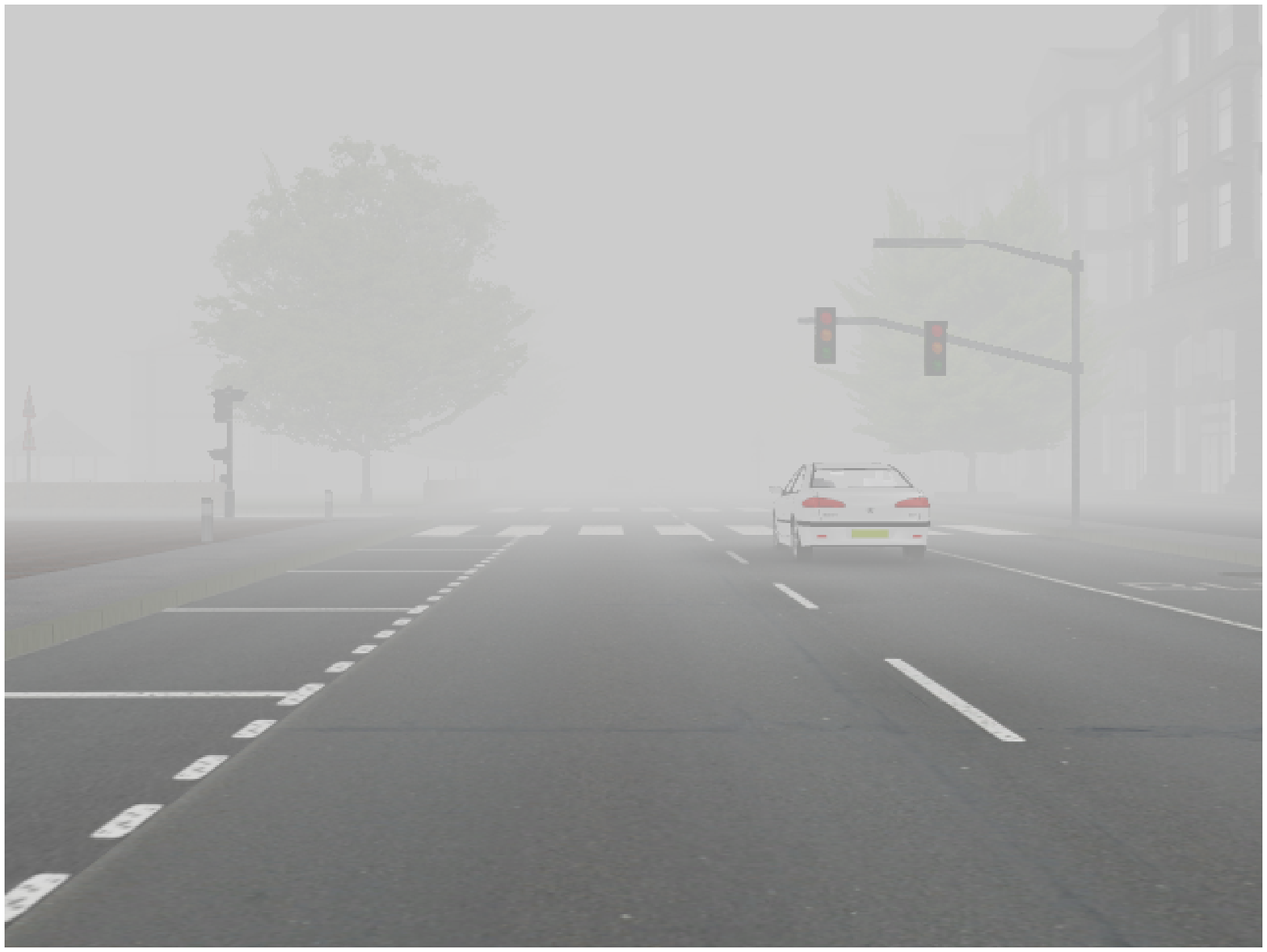}
         \put (20,90) {\red\textbf\small GT: Light}
         \put (20,80) {\red\textbf\small Pred: Light}
         \end{overpic}
    }
    \end{subfigure}
~    \begin{subfigure}[b]{\egswidth}
    \mytbox{
        \begin{overpic}[width=\egssubwidth,height=\egssubheight]{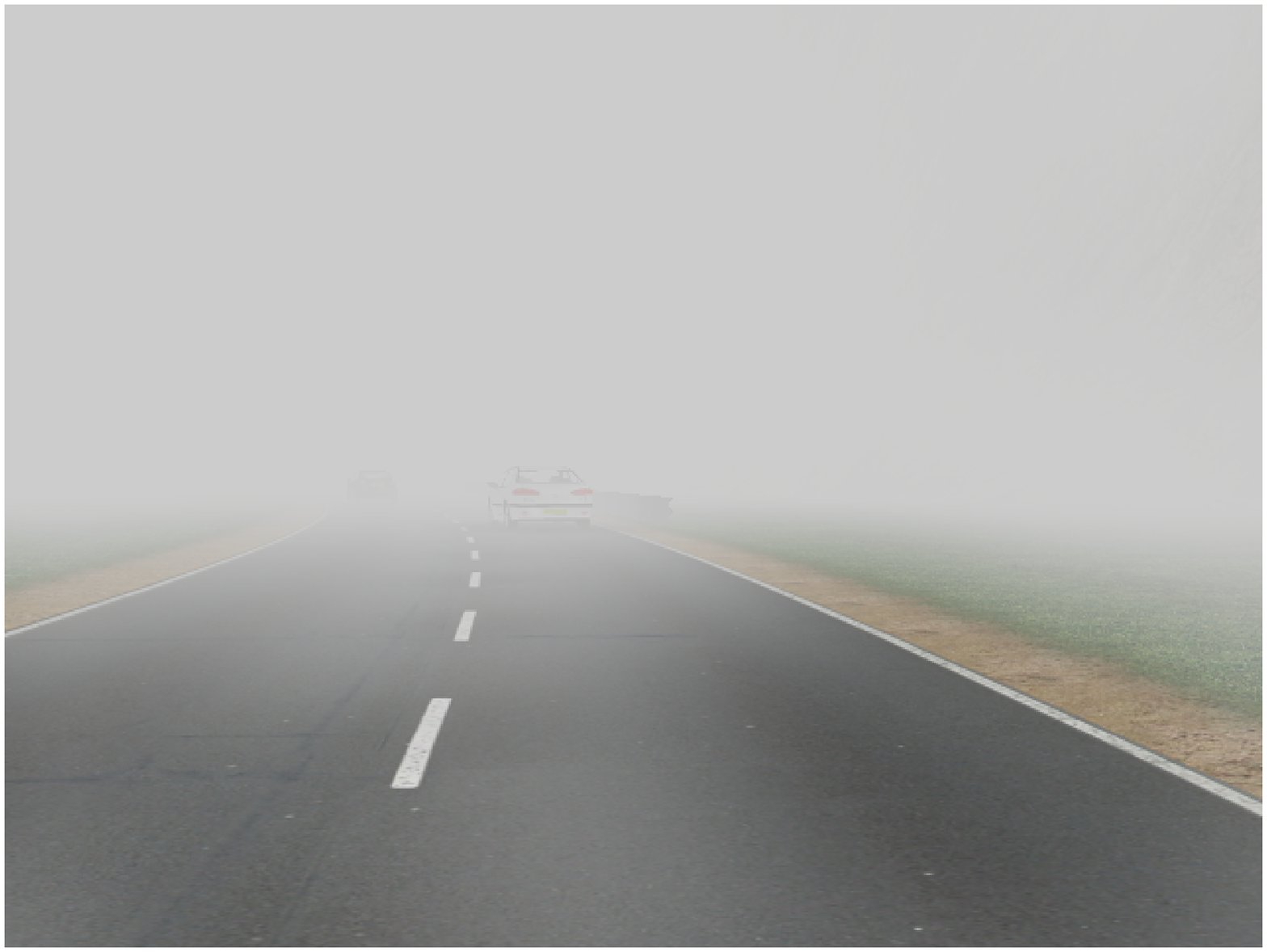}
         \put (20,90) {\red\textbf\small GT: Light}
         \put (20,80) {\red\textbf\small Pred: Light}
         \end{overpic}
    }
    \end{subfigure}
~    \begin{subfigure}[b]{\egswidth}
    \mytbox{
        \begin{overpic}[width=\egssubwidth,height=\egssubheight]{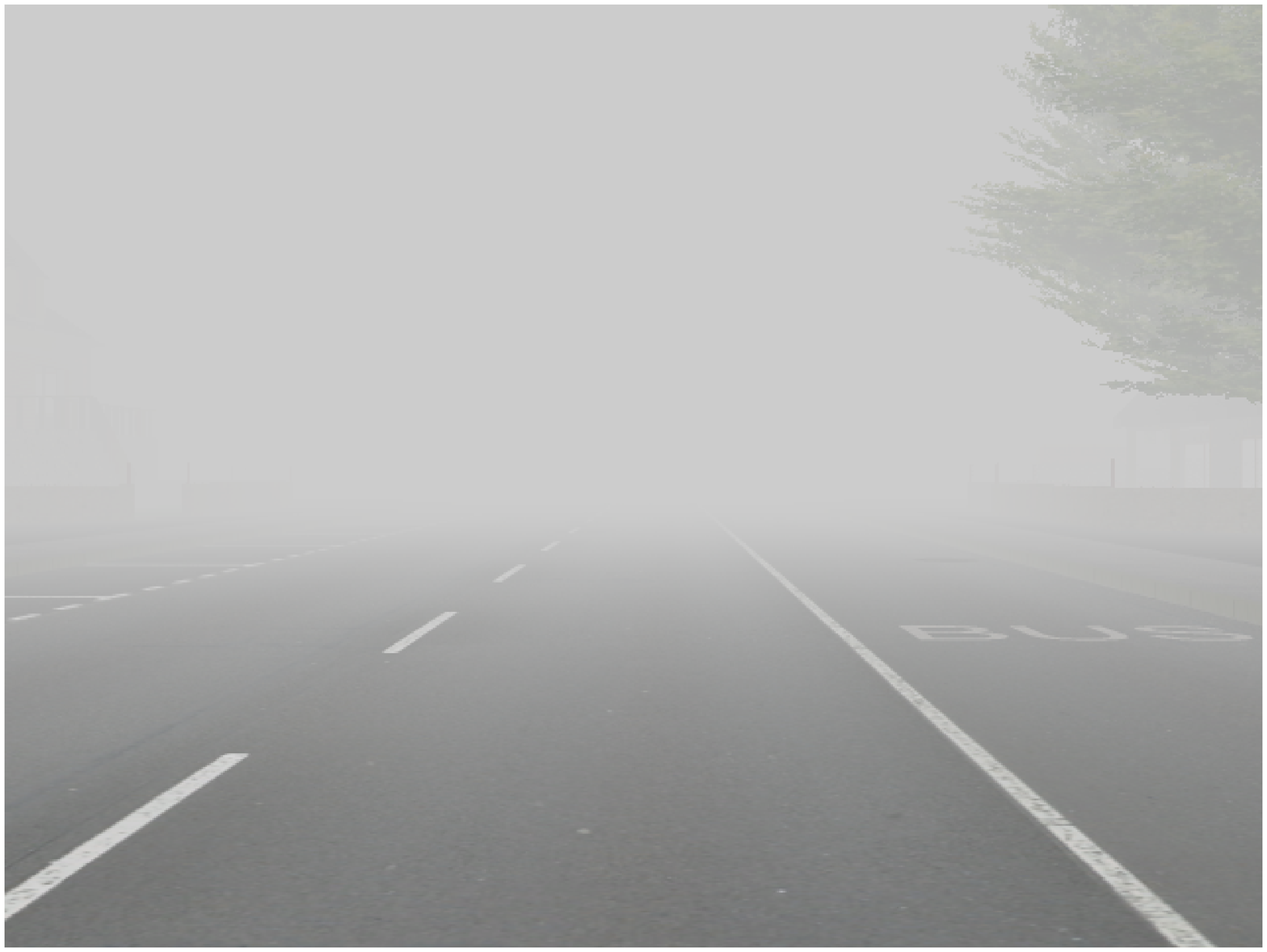}
         \put (20,90) {\red\textbf\small GT: Heavy}
         \put (20,80) {\red\textbf\small Pred: Heavy}
         \end{overpic}
    }
    \end{subfigure}
~    \begin{subfigure}[b]{\egswidth}
    \mytbox{
        \begin{overpic}[width=\egssubwidth,height=\egssubheight]{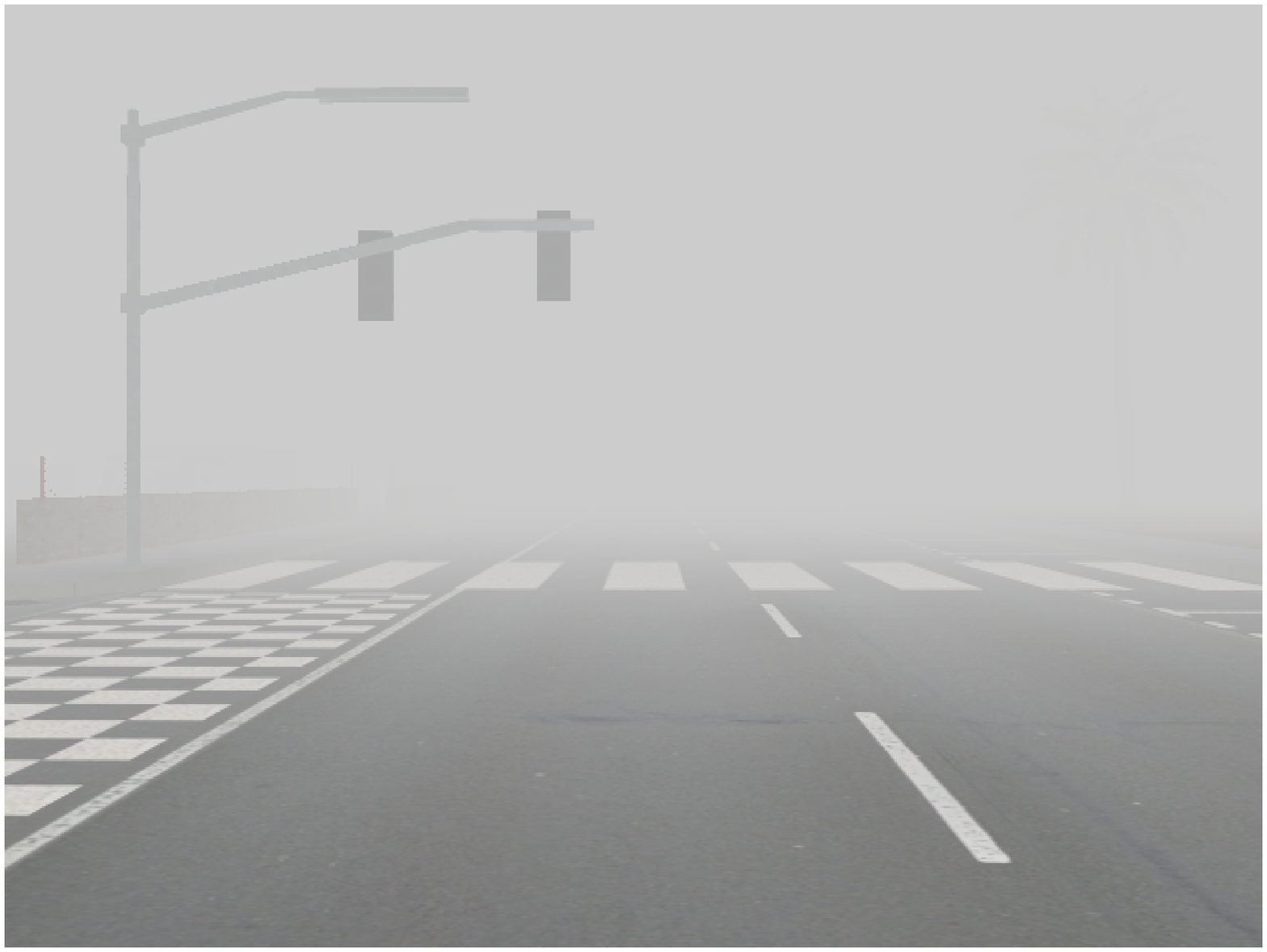}
         \put (20,90) {\red\textbf\small GT: Heavy}
         \put (20,80) {\red\textbf\small Pred: Heavy}
         \end{overpic}
    }
    \end{subfigure}

\end{figure*}

\section*{Acknowledgement}
The authors would like to thank the generous support of Google, Xerox,  New York State  CEIS, and NYS IDS.

%
{\small
    \bibliographystyle{abbrv}
    \bibliography{refers}  

\begin{thebibliography}{10}

\bibitem{Aoki:2009:VRU:1518701.1518762}
P.~M. Aoki, R.~J. Honicky, A.~Mainwaring, C.~Myers, E.~Paulos, S.~Subramanian,
  and A.~Woodruff.
\newblock A vehicle for research: Using street sweepers to explore the
  landscape of environmental community action.
\newblock SIGCHI '09. ACM.

\bibitem{1975}
D.~J. Best and D.~E. Roberts.
\newblock Algorithm as 89: The upper tail probabilities of spearman's rho.
\newblock {\em Journal of the Royal Statistical Society. Series C (Applied
  Statistics)}, 24(3):pp. 377--379, 1975.

\bibitem{Chen:2014:BSM:2567948.2576941}
J.~Chen, H.~Chen, G.~Zheng, J.~Z. Pan, H.~Wu, and N.~Zhang.
\newblock Big smog meets web science: Smog disaster analysis based on social
  media and device data on the web.
\newblock WWW Companion '14, 2014.

\bibitem{5567108}
K.~He, J.~Sun, and X.~Tang.
\newblock Single image haze removal using dark channel prior.
\newblock {\em PAMI}, 33(12), Dec 2011.

\bibitem{6319316}
K.~He, J.~Sun, and X.~Tang.
\newblock Guided image filtering.
\newblock {\em PAMI}, 35(6), June 2013.

\bibitem{jcsb.1000161}
S.~W. Jun~Mao, Uthai~Phommasak and H.~Shioya.
\newblock Detecting foggy images and estimating the haze degree factor.
\newblock {\em Journal of Computer Science \& Systems Biology}, 7(6):226--228,
  2014.

\bibitem{Depth2015Liu}
F.~Liu, C.~Shen, G.~Lin, and I.~Reid.
\newblock Learning depth from single monocular images using deep convolutional
  neural fields.
\newblock {\em Technical report, University of Adelaide}, 2015.

\bibitem{6921638}
S.~Mei, H.~Li, J.~Fan, X.~Zhu, and C.~Dyer.
\newblock Inferring air pollution by sniffing social media.
\newblock In {\em Advances in Social Networks Analysis and Mining (ASONAM),
  2014 IEEE/ACM International Conference on}, Aug 2014.

\bibitem{poduri2010visibility}
S.~Poduri, A.~Nimkar, and G.~S. Sukhatme.
\newblock Visibility monitoring using mobile phones.
\newblock {\em Annual Report: Center for Embedded Networked Sensing}, pages
  125--127, 2010.

\bibitem{raaschou2013air}
O.~Raaschou-Nielsen, Z.~J. Andersen, R.~Beelen, E.~Samoli, M.~Stafoggia,
  G.~Weinmayr, B.~Hoffmann, P.~Fischer, M.~J. Nieuwenhuijsen, B.~Brunekreef,
  et~al.
\newblock Air pollution and lung cancer incidence in 17 european cohorts:
  prospective analyses from the european study of cohorts for air pollution
  effects (escape).
\newblock {\em The lancet oncology}, 14(9):813--822, 2013.

\bibitem{4531745}
A.~Saxena, M.~Sun, and A.~Ng.
\newblock Make3d: Learning 3d scene structure from a single still image.
\newblock {\em PAMI}, 31(5):824--840, May 2009.

\bibitem{6190796}
J.-P. Tarel, N.~Hautiere, L.~Caraffa, A.~Cord, H.~Halmaoui, and D.~Gruyer.
\newblock Vision enhancement in homogeneous and heterogeneous fog.
\newblock {\em Intelligent Transportation Systems Magazine, IEEE}, 4(2):6--20,
  Summer 2012.

\bibitem{5548128}
J.-P. Tarel, N.~Hautière, A.~Cord, D.~Gruyer, and H.~Halmaoui.
\newblock Improved visibility of road scene images under heterogeneous fog.
\newblock In {\em Intelligent Vehicles Symposium (IV), 2010 IEEE}, pages
  478--485, June 2010.

\end{thebibliography}
}
\balancecolumns 
\end{document}